\documentclass[10pt,twocolumn,letterpaper]{article}

\usepackage{times}
\usepackage{epsfig}
\usepackage{graphicx}
\usepackage{amsmath}
\usepackage{amssymb}
\usepackage{sidecap}
\usepackage{epstopdf}
\usepackage{float}

\usepackage{caption}
\usepackage{subcaption}

\newsavebox\CBox
\def\textBF#1{\sbox\CBox{#1}\resizebox{\wd\CBox}{\ht\CBox}{\textbf{#1}}}

\usepackage[pagebackref=true,breaklinks=true,letterpaper=true,colorlinks,bookmarks=false]{hyperref}
\usepackage[top = 2.00cm, bottom = 2.00cm, left = 1.5cm, right = 1.75cm]{geometry}

\begin{document}

\title{\resizebox{2\columnwidth}{!}
{
DSLR-Quality Photos on Mobile Devices with Deep Convolutional Networks}\vspace{2.6mm}
}

\author{Andrey Ignatov, \hspace{2mm} Nikolay Kobyshev,  \hspace{2mm} Kenneth Vanhoey,  \hspace{2mm} Radu Timofte,  \hspace{2mm} Luc Van Gool \vspace{1.8mm} \\
\\
\textsf{ETH Zurich}\\
\\
{\tt\small andrey.ignatoff@gmail.com, \{nk, vanhoey, timofter, vangool\}@vision.ee.ethz.ch}\vspace{-9.4mm}\\
}

\date{}

\makeatletter
\g@addto@macro\@maketitle{
  \begin{figure}[H]
  \setlength{\linewidth}{\textwidth}
  \setlength{\hsize}{\textwidth}
  \centering
  \includegraphics[width=0.79\linewidth]{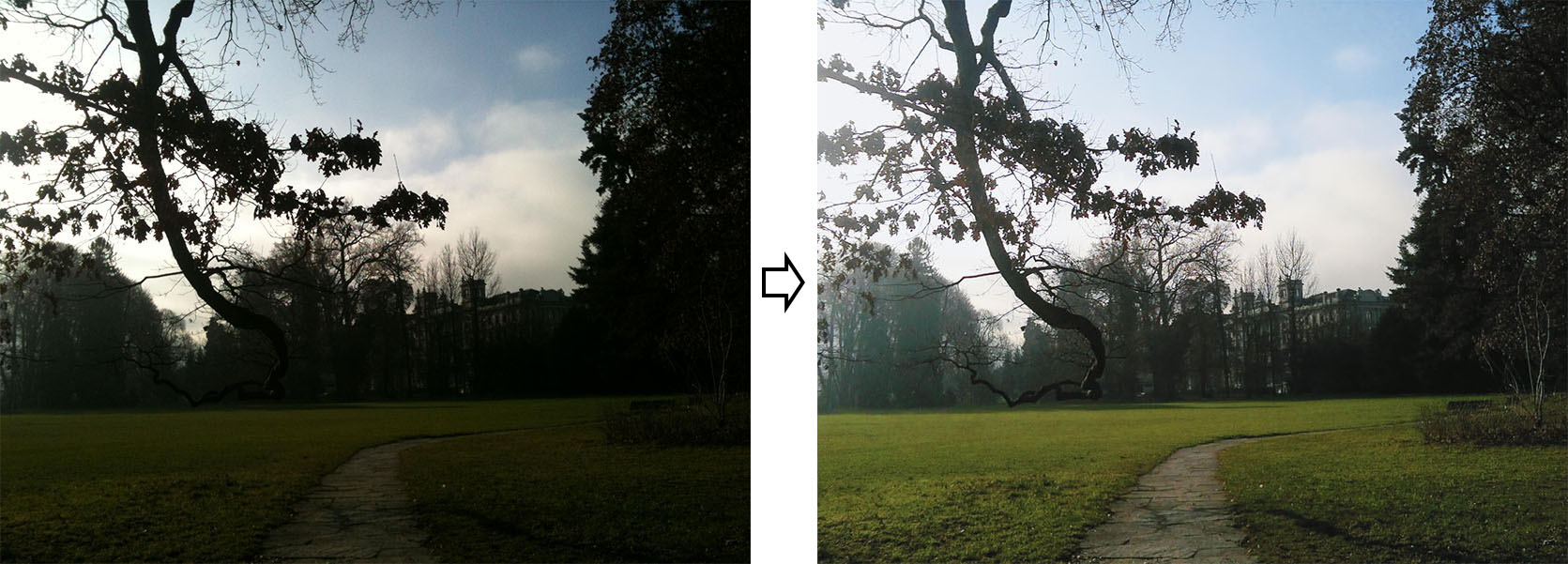}
  \vspace{0.1cm}
  \caption{iPhone 3GS photo enhanced to DSLR-quality by our method. Best zoomed on screen.}
  \vspace{0.2cm}
  \end{figure}
}

\makeatother

\maketitle

\begin{abstract}
Despite a rapid rise in the quality of built-in smartphone cameras, their physical limitations~--- small sensor size, compact lenses and the lack of specific hardware,~--- impede them to achieve the quality results of DSLR cameras. In this work we present an end-to-end deep learning approach that bridges this gap by translating ordinary photos into DSLR-quality images. We propose learning the translation function using a residual convolutional neural network that improves both color rendition and image sharpness. Since the standard mean squared loss is not well suited for measuring perceptual image quality, we introduce a composite perceptual error function that combines content, color and texture losses. The first two losses are defined analytically, while the texture loss is learned in an adversarial fashion. We also present DPED, a large-scale dataset that consists of real photos captured from three different phones and one high-end reflex camera. Our quantitative and qualitative assessments reveal that the enhanced image quality is comparable to that of DSLR-taken photos, while the methodology is generalized to  any type of digital camera.
\end{abstract}

\section{Introduction}

During the last several years there has been a significant improvement in compact camera sensors quality, which has brought mobile photography to a substantially new level. Even low-end devices are now able to take reasonably good photos in appropriate lighting conditions, thanks to their advanced software and hardware tools for post-processing.
However, when it comes to artistic quality, mobile devices still fall behind their DSLR counterparts. Larger sensors and high-aperture optics yield better photo resolution, color rendition and less noise, whereas their additional sensors help to fine-tune shooting parameters. These physical differences result in strong obstacles, making DSLR camera quality unattainable for compact mobile devices.


While a number of photographer tools for automatic image enhancement exist, they are usually focused on adjusting only global parameters such as contrast or brightness, without improving texture quality or taking image semantics into account. Besides that, they are usually based on a pre-defined set of rules that do not always consider the specifics of a particular device. Therefore, the dominant approach to photo post-processing is still based on manual image correction using specialized retouching software.

\subsection{Related work}
The problem of automatic image quality enhancement has not been addressed in its entirety in the area of computer vision, though a number of sub-tasks and related problems have been already successfully solved using deep learning techniques. Such tasks are usually dealing with image-to-image translation problems, and their common property is that they are targeted at removing artificially added artifacts to the original images. Among the related problems are the following:

\textbf{Image super-resolution} aims at restoring the original image from its downscaled version.
In \cite{Dong2014} a CNN architecture and MSE loss are used for directly learning low to high resolution mapping.
It is the first CNN-based solution to achieve top performance in single image super-resolution, comparable with non-CNN methods~\cite{Timofte2015}.
The subsequent works developed deeper and more complex CNN architectures (e.g.,~\cite{Kim2016,Shi2016,Mao2016}).
Currently, the best photo-realistic results on this task are achieved using a VGG-based loss function~\cite{Johnson2016} and adversarial networks~\cite{Ledig2016} that turned out to be efficient at recovering plausible high-frequency components.

\textbf{Image deblurring/dehazing} tries to remove artificially added haze or blur from the images.
Usually, MSE is used as a target loss function and the proposed CNN architectures consist of 3 to 15 convolutional layers~\cite{Ling2016,Cai2016,Hradis2015} or are bi-channel CNNs~\cite{Ren2016}.

\textbf{Image denoising/sparse inpainting} similarly targets removal of noise and artifacts from the pictures. In~\cite{Zhang2016} the authors proposed weighted MSE together with a 3-layer CNN, while in~\cite{Svoboda2016} it was shown that an 8-layer residual CNN performs better when using a standard mean square error. Among other solutions are a bi-channel CNN~\cite{Zhou2015}, a 17-layer CNN~\cite{Kai2016} and a recurrent CNN~\cite{Yang2016} that was reapplied several times to the produced results.

\textbf{Image colorization.} Here the goal is to recover colors that were removed from the original image. The baseline approach for this problem is to predict new values for each pixel based on its local description that consists of various hand-crafted features~\cite{Cheng2015}. Considerably better performance on this task was obtained using generative adversarial networks~\cite{Isola2016} or a 16-layer CNN with a multinomial cross-entropy loss function~\cite{Richard2016}.

\textbf{Image adjustment.} A few works considered the problem of image color/contrast/exposure adjustment. In~\cite{Yuan2012} the authors proposed an algorithm for automatic exposure correction using hand-designed features and predefined rules. In~\cite{Yan2016}, a more general algorithm was proposed that --~similarly to~\cite{Cheng2015}~-- uses local description of image pixels for reproducing various photographic styles. A different approach was considered in~\cite{Lee2016}, where images with similar content are retrieved from a database and their styles are applied to the target picture. All of these adjustments are implicitly included in our end-to-end transformation learning approach by design.

\subsection{Contributions}
The key challenge we face is dealing with all the aforementioned enhancements at once.
Even advanced tools cannot notably improve image sharpness, texture details or small color variations that were lost by the camera sensor, thus we can not generate target enhanced photos from the existing ones.
Corrupting DSLR photos and training an algorithm on the corrupted images does not work either: the solution would not generalize to real-world and very complex artifacts unless they are modeled and applied as corruptions, which is infeasible.
To tackle this problem, we present a different approach:
we propose to learn the transformation that modifies photos taken by a given camera to DSLR-quality ones.
Thus, the goal is to learn a cross-distribution translation function, where the input distribution is defined by a given mobile camera sensor, and the target distribution by a DSLR sensor.
To supervise the learning process, we create and leverage a dataset of images capturing the same scene with different cameras.
Once the function is learned, it can be further applied to unseen photos at will.

\smallskip

\smallskip

Our main contributions are:
\begin{itemize}
\setlength\itemsep{-0.3em}
\item A novel approach\footnote{\url{https://github.com/aiff22/DPED}} for the photo enhancement task based on learning a mapping function between photos from mobile devices and a DSLR camera. The target model is trained in an end-to-end fashion without using any additional supervision or handcrafted features.
\item A new large-scale dataset of over 6K photos taken synchronously by a DSLR camera and 3 low-end cameras of smartphones in a wide variety of conditions.
\item A multi-term loss function composed of color, texture and content
terms, allowing an efficient image quality estimation.
\item Experiments measuring objective and subjective quality demonstrating the advantage of the enhanced photos over the originals and, at the same time, their comparable quality with the DSLR counterparts.
\end{itemize}

The remainder of the paper is structured as follows.
In \mbox{Section~\ref{sec:dataset}} we describe the new DPED dataset.
\mbox{Section~\ref{sec:methods}} presents our architecture and the chosen loss functions.
\mbox{Section~\ref{sec:experiments}} shows and analyzes the experimental results. Finally, Section~\ref{sec:conclusions} concludes the paper.

\begin{table}[t!]
\centering
\caption{DPED camera characteristics.}
\vspace{-0.1cm}
\resizebox{\columnwidth}{!}
{
\begin{tabular}{lccl}
Camera & Sensor & Image size & Photo quality\\[0.1cm]
\hline
\textit{iPhone 3GS}& 3 MP & $2048\times1536$ & Poor\\
\textit{BlackBerry Passport}& 13 MP & $4160\times3120$ & Mediocre\\
\textit{Sony Xperia Z}&13 MP & $2592\times 1944$ & Average\\
\textit{Canon 70D DSLR}&20 MP & $3648\times 2432$ & Excellent
\end{tabular}
}
\label{tab:DPED_cameras}
\end{table}

\begin{figure}[t!]
\vspace{0.07cm}
\centering
    \includegraphics[width=0.65\linewidth]{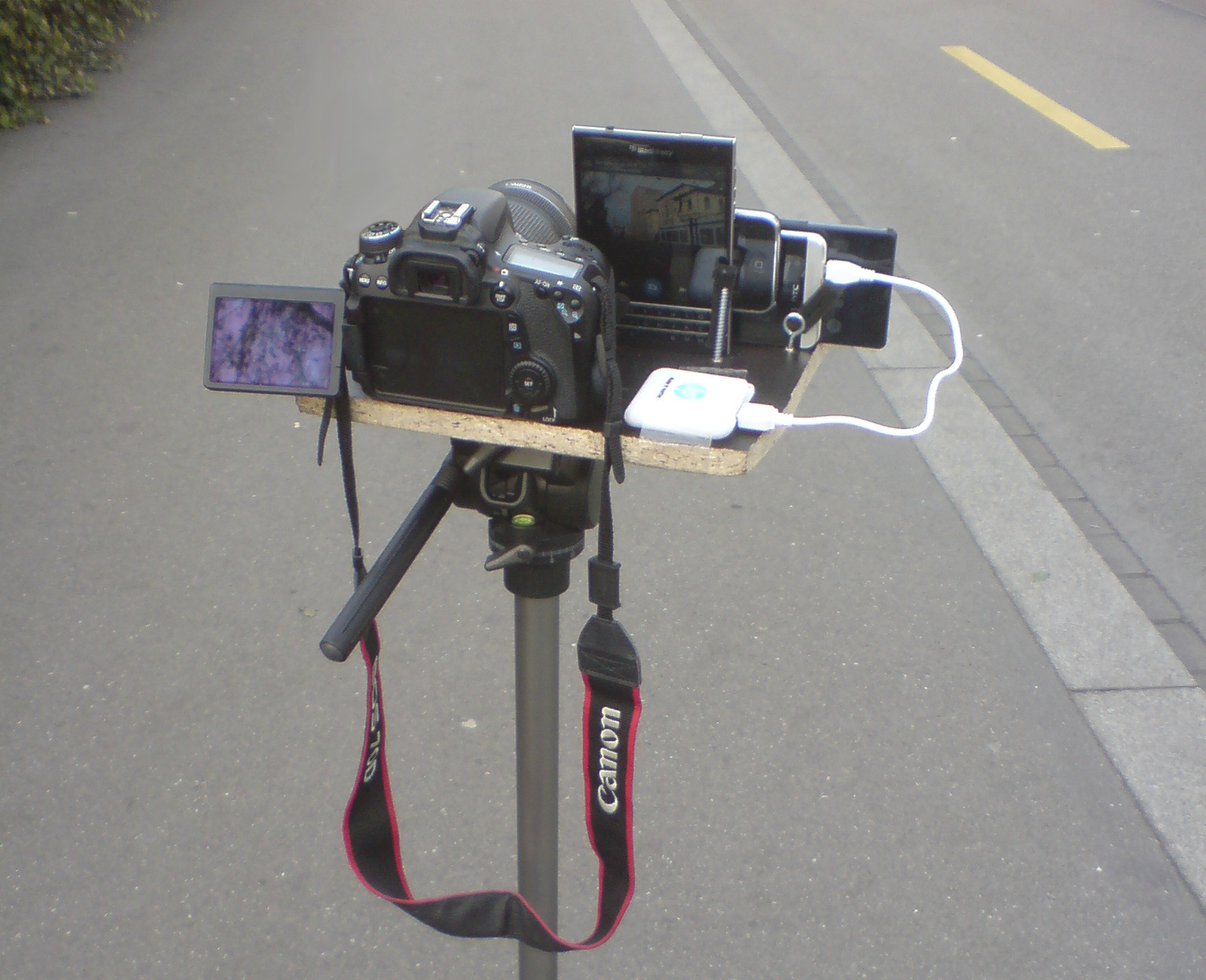}
   \caption{The rig with the four DPED cameras from Table~\ref{tab:DPED_cameras}.}
\label{fig:tripod}
\vspace{-0.4cm}
\end{figure}

\section{DSLR Photo Enhancement Dataset}
\label{sec:dataset}
\begin{figure}[t!]
\centering
\setlength{\tabcolsep}{1pt}
\resizebox{\linewidth}{!}
{
\begin{tabular}{cccc}
iPhone & BlackBerry & Sony & Canon\\
    \includegraphics[width=0.24\linewidth]{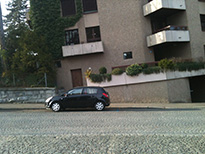}&
    \includegraphics[width=0.24\linewidth]{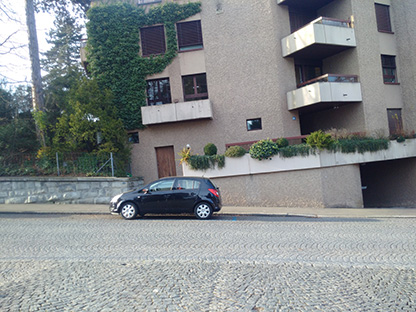}&
    \includegraphics[width=0.24\linewidth]{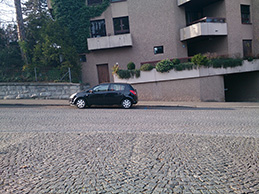}&
    \includegraphics[width=0.24\linewidth]{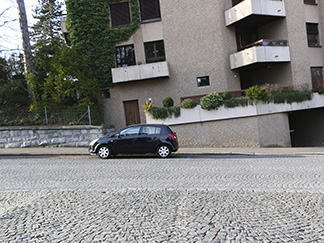}\\
    \includegraphics[width=0.24\linewidth]{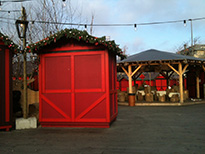}&
    \includegraphics[width=0.24\linewidth]{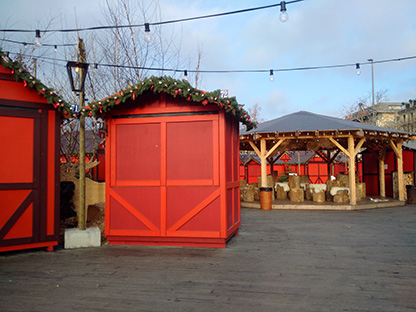}&
    \includegraphics[width=0.24\linewidth]{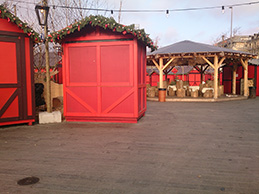}&
    \includegraphics[width=0.24\linewidth]{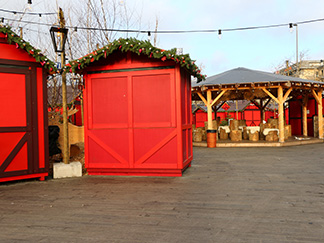}\\
    \includegraphics[width=0.24\linewidth]{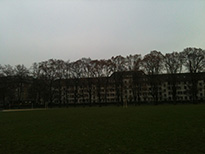}&
    \includegraphics[width=0.24\linewidth]{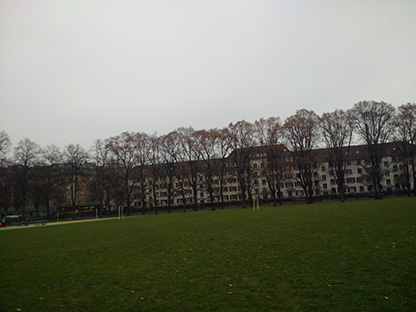}&
    \includegraphics[width=0.24\linewidth]{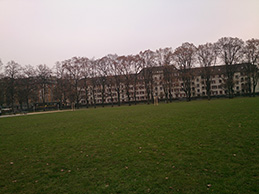}&
    \includegraphics[width=0.24\linewidth]{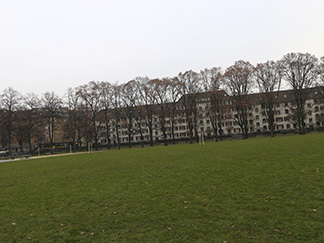}\\
    \includegraphics[width=0.24\linewidth]{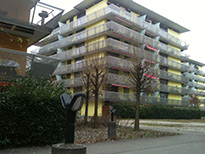}&
    \includegraphics[width=0.24\linewidth]{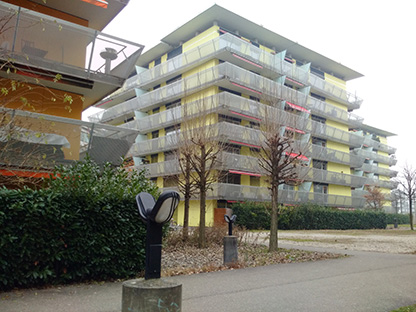}&
    \includegraphics[width=0.24\linewidth]{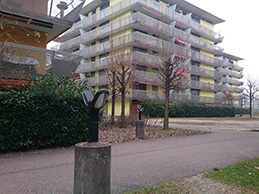}&
    \includegraphics[width=0.24\linewidth]{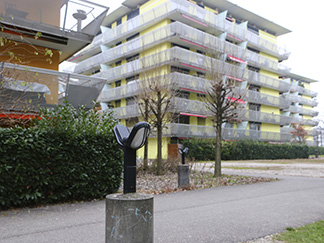}
\end{tabular}
}
\vspace{-0.25cm}
   \caption{Example quadruplets of images taken synchronously by the DPED four cameras.}
\label{fig:example_quadruplet}
\vspace{-0.1cm}
\end{figure}

In order to tackle the problem of image translation from poor quality images captured by smartphone cameras to superior quality images achieved by a professional DSLR camera,
we introduce a large-scale real-world dataset, namely the ``DSLR Photo Enhancement Dataset'' (DPED)\footnote{\url{http://dped-photos.vision.ee.ethz.ch}}, that can be used for the general photo quality enhancement task.
DPED consists of photos taken in the wild synchronously by three smartphones and one DSLR camera. The devices used to collect the data are described in Table~\ref{tab:DPED_cameras} and example quadruplets can be seen in Figure~\ref{fig:example_quadruplet}.

To ensure that all cameras were capturing photos simultaneously, the devices were mounted on a tripod and activated remotely by a wireless control system (see Figure~\ref{fig:tripod}).
In total, over 22K photos were collected during 3 weeks, including 4549 photos from Sony smartphone, 5727 from iPhone and 6015 photos from each Canon and BlackBerry cameras. The photos were taken during the daytime in a wide variety of places and in various illumination and weather conditions. The photos were captured in automatic mode, and we used default settings for all cameras throughout the whole collection procedure.

\begin{figure}[t!]
\vspace{0.08cm}
\centering
    \includegraphics[width=\linewidth]{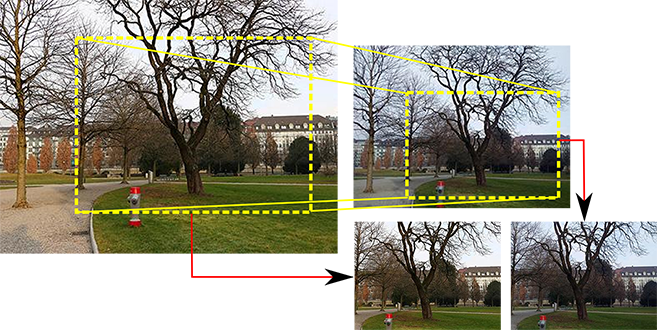}
\vspace{-0.2cm}
   \caption{Matching algorithm: an overlapping region is determined by SIFT descriptor matching, followed by a non-linear transform and a crop resulting in two images of the same resolution representing the same scene. Here: Canon and BlackBerry images, respectively.}
\label{fig:matching}
\vspace{-0.2cm}
\end{figure}

\paragraph{Matching algorithm.}
The synchronously captured images are not perfectly aligned since the cameras have different viewing angles and positions as can be seen in Figure~\ref{fig:example_quadruplet}.
To address this, we performed additional non-linear transformations resulting in a fixed-resolution image that our network takes as an input.
The algorithm goes as follows (see Fig.~\ref{fig:matching}).
First, for each (phone-DSLR) image pair, we compute and match SIFT keypoints~\cite{Lowe2004} across the images.
These are used to estimate a homography using RANSAC~\cite{vlfeat}.
We then crop both images to the intersection part and downscale the DSLR image crop to the size of the phone crop.

Training CNN on the aligned high-resolution images is infeasible, thus patches of size 100$\times$100px were extracted from these photos. Our preliminary experiments revealed that larger patch sizes do not lead to better performance, while requiring considerably more computational resources. We extracted patches using a non-overlapping sliding window. The window was moving in parallel along both images from each phone-DSLR image pair, and its position on the phone image was additionally adjusted by shifts and rotations based on the cross-correlation metrics. To avoid significant displacements, only patches with cross-correlation greater than 0.9 were included in the dataset. Around 100 original images were reserved for testing, the rest of the photos were used for training and validation. This procedure resulted in 139K, 160K and 162K training and 2.4-4.3K test patches for BlackBerry-Canon, iPhone-Canon and Sony-Canon pairs, respectively. It should be emphasized that both training and test patches are precisely matched, the potential shifts do not exceed 5 pixels. In the following we assume that these patches of size 3$\times$100$\times$100 constitute the input data to our CNNs.

\section{Method}
\label{sec:methods}

Given a low-quality photo $I_{s}$ (source image), the goal of the considered enhancement task is to reproduce the image $I_{t}$ (target image) taken by a DSLR camera. A deep residual CNN $F_{\textbf{W}}$ parameterized by weights $\textbf{W}$ is used to learn the underlying translation function.
Given the training set $\{I_{s}^j, \, I_{t}^j\}_{j=1}^N$ consisting of $N$ image pairs, it is trained to minimize:
\begin{equation}
\textbf{W}^{*} = \arg\min_{\textbf{W}} \frac{1}{N} \sum_{j=1}^N \, \mathcal{L}\bigl(F_{\textbf{W}}(I_{s}^j), \, I_{t}^j\bigr),
\end{equation}
where $\mathcal{L}$ denotes a multi-term loss function we detail in section~\ref{ssc:losses}.
We then define the system architecture of our solution in Section~\ref{ssc:system}.

\subsection{Loss function}
\label{ssc:losses}
The main difficulty of the image enhancement task is that input and target photos cannot be matched densely (i.e., pixel-to-pixel): different optics and sensors cause specific local non-linear distortions and aberrations, leading to a non-constant shift of pixels between each image pair even after precise alignment.
Hence, the standard per-pixel losses, besides being doubtful as a perceptual quality metric, are not applicable in our case.
We build our loss function under the assumption that the overall perceptual image quality can be decomposed into three independent parts: i) color quality, ii) texture quality and iii) content quality.
We now define loss functions for each component, and ensure invariance to local shifts by design.

\begin{figure}[t]
\centering
   \includegraphics[width=\linewidth]{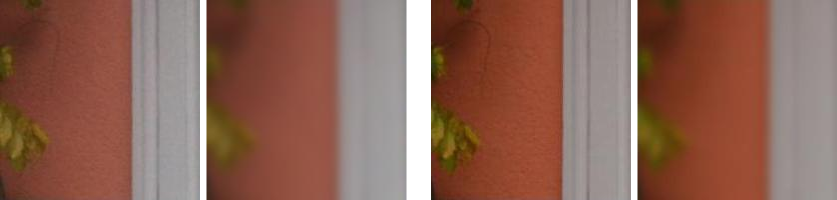}
\vspace{-0.25cm}
   \caption{Fragments from the original and blurred images taken by the phone (two left-most) and DSLR (two right-most) camera. Blurring removes high-frequencies and makes color comparison easier.}
\label{fig:blur_ex}
\end{figure}

\subsubsection{Color loss}
\label{sec:methods:color}
To measure the color difference between the enhanced and target images, we propose applying a Gaussian blur (see Figure~\ref{fig:blur_ex}) and computing Euclidean distance between the obtained representations. In the context of CNNs, this is equivalent to using one additional convolutional layer with a fixed Gaussian kernel followed by the mean squared error (MSE) function. Color loss can be written as:
\begin{equation}
\mathcal{L}_{\text{color}}(X, Y) = \|X_b - Y_b\|_{2}^2,
\end{equation}
where $X_b$ and $Y_b$ are the blurred images of $X$ and $Y$, resp.:
\begin{equation}
X_b(i,j) = \sum_{k,l} X(i+k,j+l) \cdot G(k,l), \\
\end{equation}
and the 2D Gaussian blur operator is given by
\begin{equation}
G(k,l) = A \exp\biggl(-\frac{(k - \mu_x)^2}{2\sigma_x} -\frac{(l - \mu_y)^2}{2\sigma_y}\biggr)
\end{equation}
where we defined $A=0.053$, $\mu_{x,y}=0$, and $\sigma_{x,y}=3$.

\begin{figure}[t]
	\centering
   		\includegraphics[width=\linewidth]{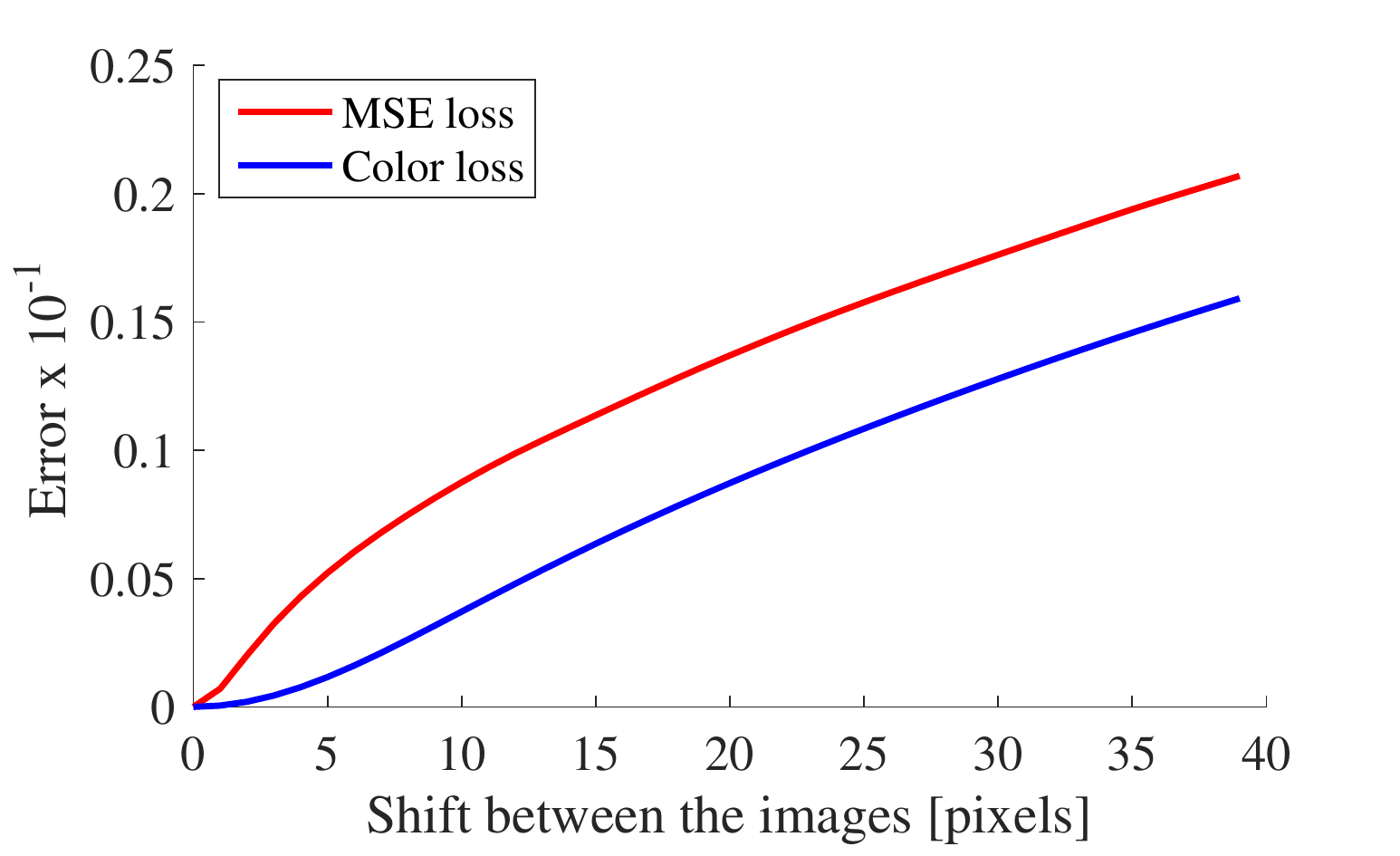}
    \vspace{-0.25cm}
   	\caption{Comparison between MSE and color loss as a function of the magnitude of shift between images. Results were averaged over 50K images.
    \label{fig:color_loss}}
\end{figure}

The idea behind this loss is to evaluate the difference in brightness, contrast and major colors between the images while eliminating texture and content comparison. Hence, we fixed a constant $\sigma$ by visual inspection as the smallest value that ensures that texture and content are dropped.
The crucial property of this loss is its invariance to small distortions. Figure~\ref{fig:color_loss} demonstrates the MSE and Color losses for image pairs (X, Y), where Y equals X shifted in a random direction by $n$ pixels. As one can see, color loss is nearly insensitive to small distortions ($\leqslant2$ pixels).
For higher shifts (3-5px), it is still about 5-10 times smaller compared to the MSE, whereas for larger displacements it demonstrates similar magnitude and behavior.
As a result, color loss forces the enhanced image to have the same color distribution as the target one, while being tolerant to small mismatches.

\subsubsection{Texture loss}
Instead of using a pre-defined loss function, we build upon generative adversarial networks (GANs)~\cite{GAN} to directly learn a suitable metric for measuring texture quality.
The discriminator CNN is applied to grayscale images so that it is targeted specifically on texture processing. It observes both fake (improved) and real (target) images, and its goal is to predict whether the input image is real or not. It is trained to minimize the cross-entropy loss function, and the texture loss is defined as a standard generator objective:
\begin{equation}
\mathcal{L}_{\text{texture}} = -\sum_{i} \log{D(F_{\textbf{W}}(I_s), I_t)},
\end{equation}
where $F_{\textbf{W}}$ and $D$ denote the generator and discriminator networks, respectively. The discriminator is pre-trained on the \{phone, DSLR\} image pairs, and then trained jointly with the proposed network as is conventional for GANs. It should be noted that this loss is shift-invariant by definition since no alignment is required in this case.

\subsubsection{Content loss}
Inspired by~\cite{Johnson2016,Ledig2016}, we define our content loss based on the activation maps produced by the ReLU layers of the pre-trained VGG-19 network. Instead of measuring per-pixel difference between the images, this loss encourages them to have similar feature representation that comprises various aspects of their content and perceptual quality. In our case it is used to preserve image semantics since other losses don't consider it. Let $\psi_j()$ be the feature map obtained after the $j$-th convolutional layer of the VGG-19 CNN, then our content loss is defined as Euclidean distance between feature representations of the enhanced and target images:
\begin{equation}
\mathcal{L}_{\text{content}} = \frac{1}{C_jH_jW_j} \|\psi_j\bigl(F_{\textbf{W}}(I_s)\bigr) - \psi_j\bigl(I_t\bigr)\|,
\end{equation}
where $C_j$, $H_j$ and $W_j$ denotes the number, height and width of the feature maps, and $F_{\textbf{W}}(I_s)$ the enhanced image.

\subsubsection{Total variation loss}
In addition to previous losses, we add total variation (TV) loss~\cite{TV2005} to enforce spatial smoothness of the produced images:
\begin{equation}
\mathcal{L}_{\text{tv}} = \frac{1}{C H W} \|\nabla_x F_{\textbf{W}}(I_s) + \nabla_y F_{\textbf{W}}(I_s)\|,
\end{equation}
where $C$, $H$ and $W$ are the dimensions of the generated image $F_{\textbf{W}}(I_s)$.
As it is relatively lowly weighted (see Eqn.~\ref{eq:loss}), it does not harm high-frequency components while it is quite effective at removing salt-and-pepper noise.

\subsubsection{Total loss} Our final loss is defined as a weighted sum of previous losses with the following coefficients:
\begin{equation}
\label{eq:loss}
\mathcal{L}_{\text{total}} = \mathcal{L}_{\text{content}} + 0.4 \cdot \mathcal{L}_{\text{texture}} + 0.1 \cdot \mathcal{L}_{\text{color}} + 400 \cdot \mathcal{L}_{\text{tv}},
\end{equation}
where the content loss is based on the features produced by the \textit{relu}\makebox[1mm]{\hrulefill}5\makebox[1mm]{\hrulefill}4 layer of the VGG-19 network. The coefficients were chosen based on preliminary experiments on the DPED training data.

\subsection{Generator and Discriminator CNNs}
\label{ssc:system}
\begin{figure}[t!]
\centering
   \includegraphics[width=\linewidth]{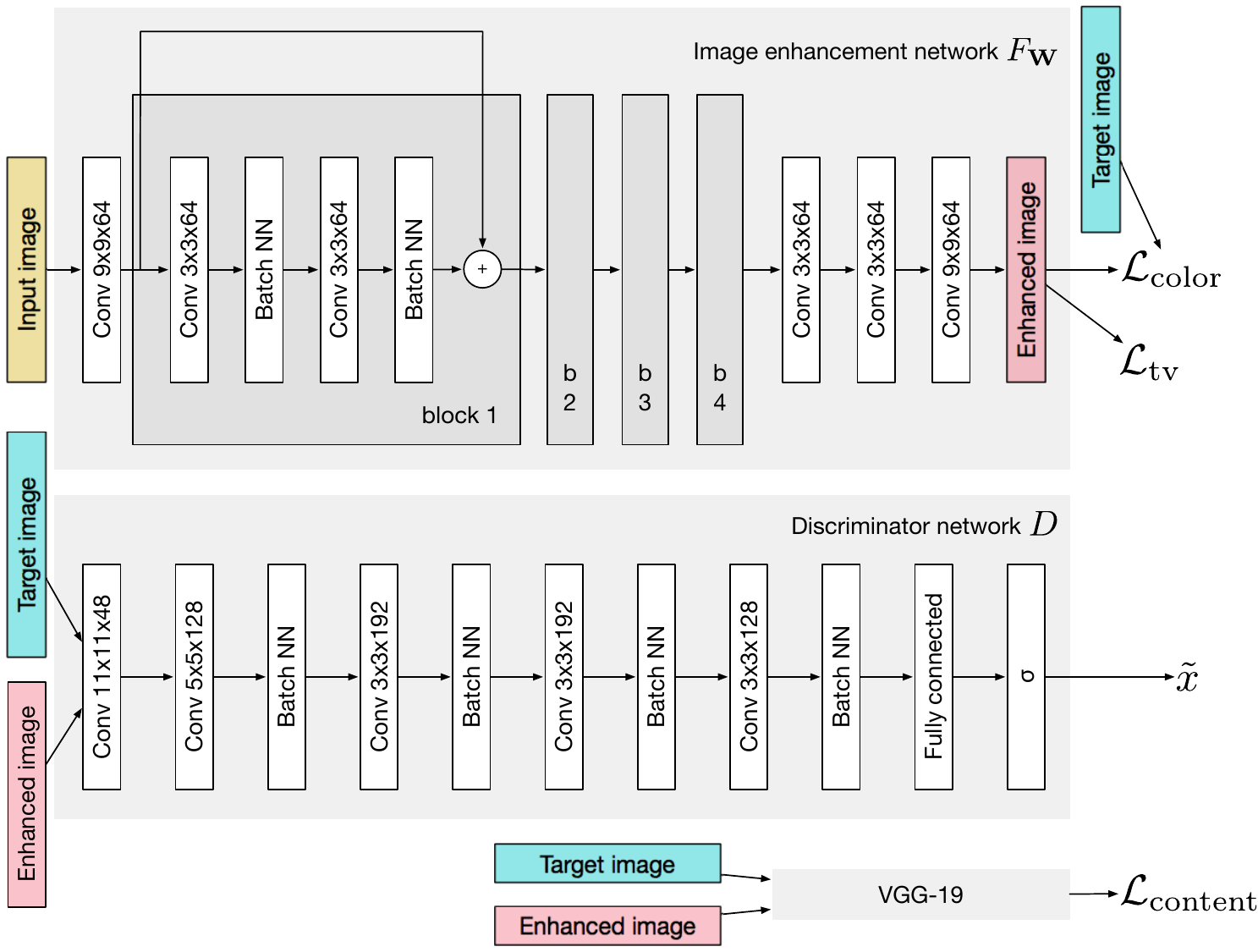}
\vspace{-0.2cm}
   \caption{The overall architecture of the proposed system.}
\label{fig:architecture}
\vspace{0.2cm}
\end{figure}

Figure~\ref{fig:architecture} illustrates the overall architecture of the proposed CNNs. Our image transformation network is fully-convolutional, and starts with a 9$\times$9 layer followed by four residual blocks. Each residual block consists of two 3$\times$3 layers alternated with batch-normalization layers. We use two additional layers with kernels of size 3$\times$3 and one with 9$\times$9 kernels after the residual blocks. All layers in the transformation network have 64 channels and are followed by a \textit{ReLU} activation function, except for the last one, where a scaled \textit{tanh} is applied to the outputs.

The discriminator CNN consists of five convolutional layers each followed by a \textit{LeakyReLU} nonlinearity and batch normalization. The first, second and fifth convolutional layers are strided with a step size of 4, 2 and 2, respectively. A sigmoidal activation function is applied to the outputs of the last fully-connected layer containing 1024 neurons and produces a probability that the input image was taken by the target DSLR camera.

\subsection{Training details}

The network was trained on a \textit{NVidia Titan X} GPU for 20K iterations using a batch size of 50. The parameters of the network were optimized using \textit{Adam}~\cite{Adam14} modification of stochastic gradient descent with a learning rate of 5e-4. The whole pipeline and experimental setup was identical for all cameras.

\begin{figure*}
   \includegraphics[width=0.33\linewidth]{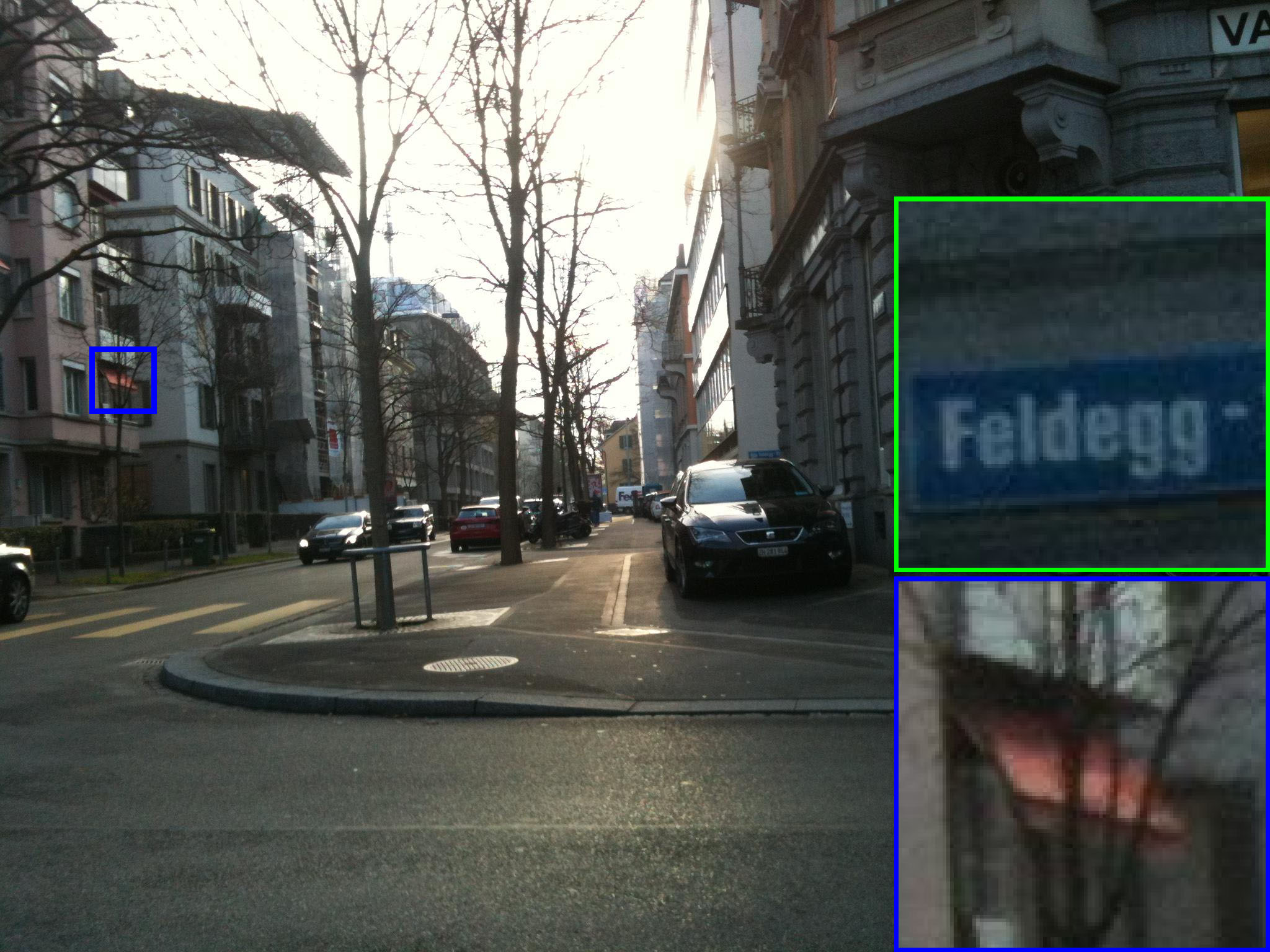}
   \includegraphics[width=0.33\linewidth]{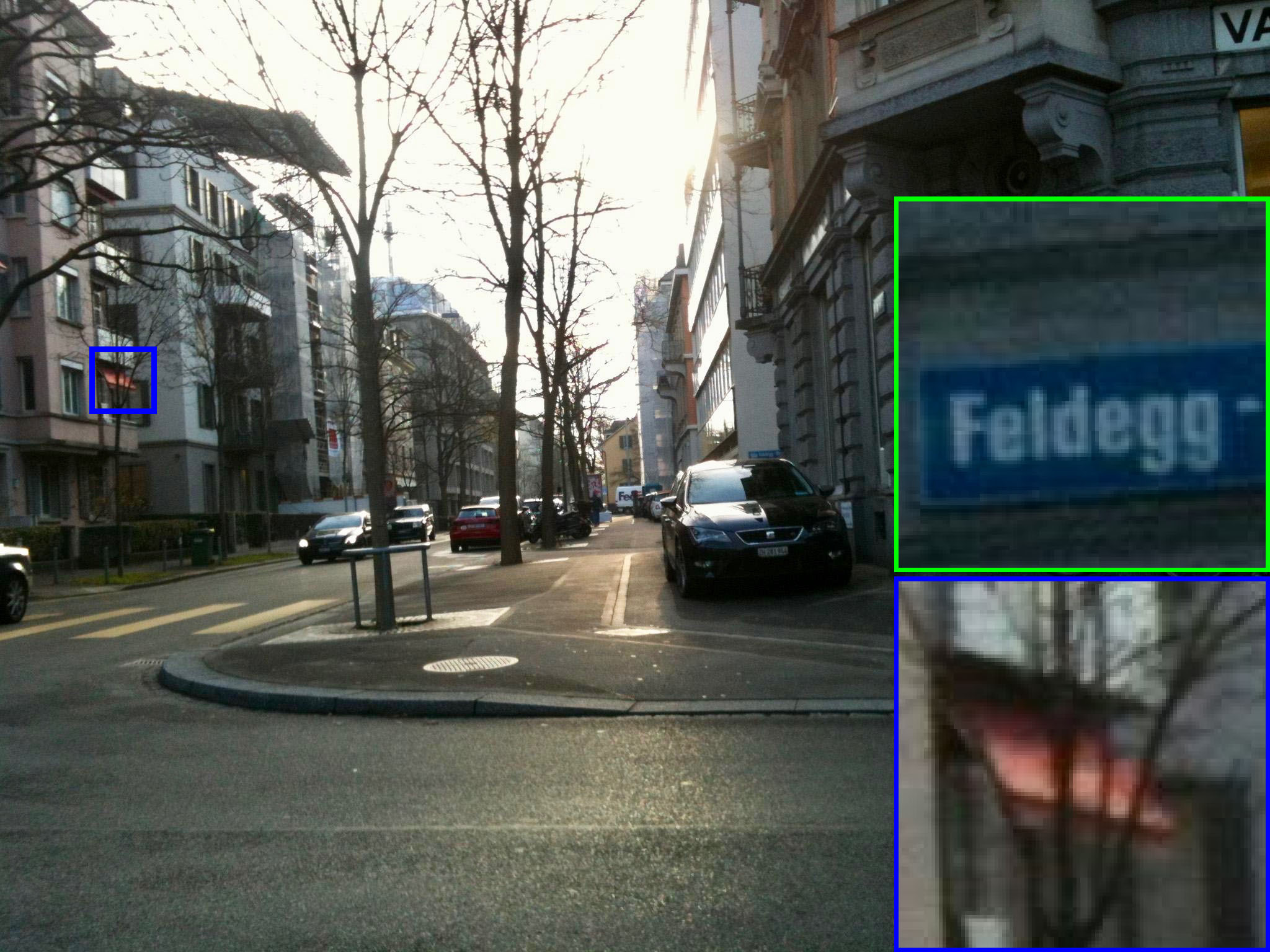}
   \includegraphics[width=0.33\linewidth]{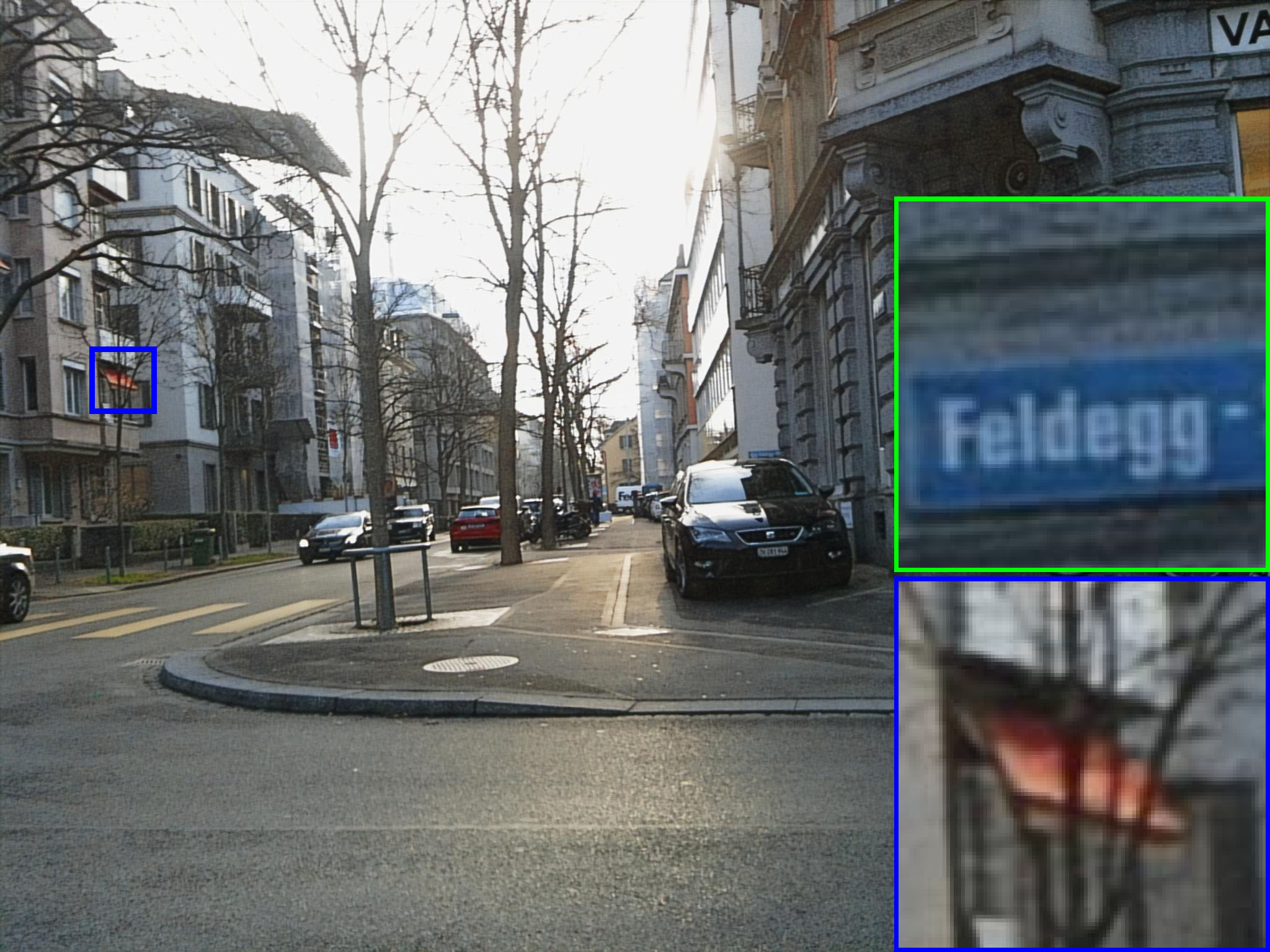}

   \vspace{0.5mm}

   \includegraphics[width=0.33\linewidth]{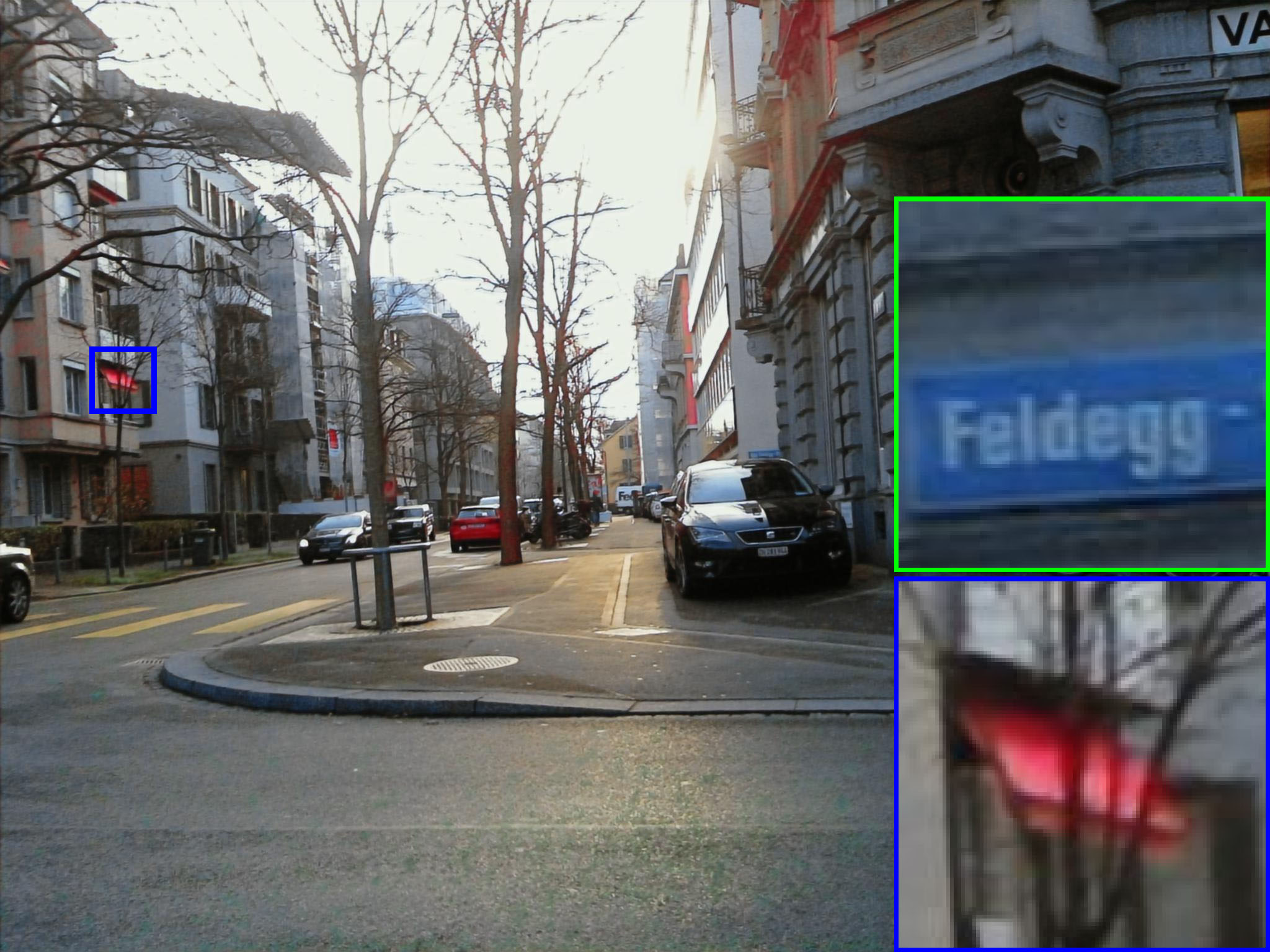}
   \includegraphics[width=0.33\linewidth]{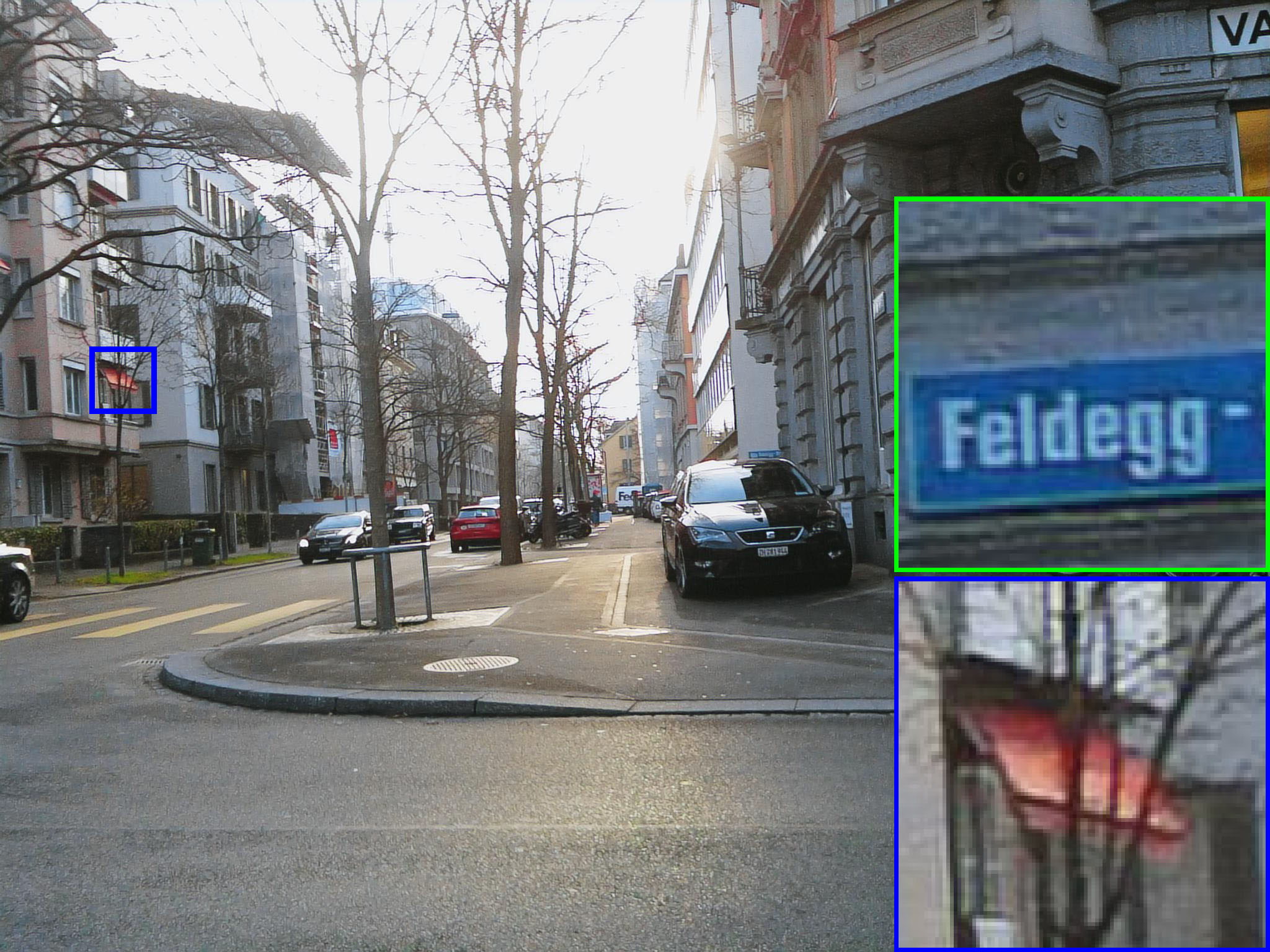}
   \includegraphics[width=0.33\linewidth]{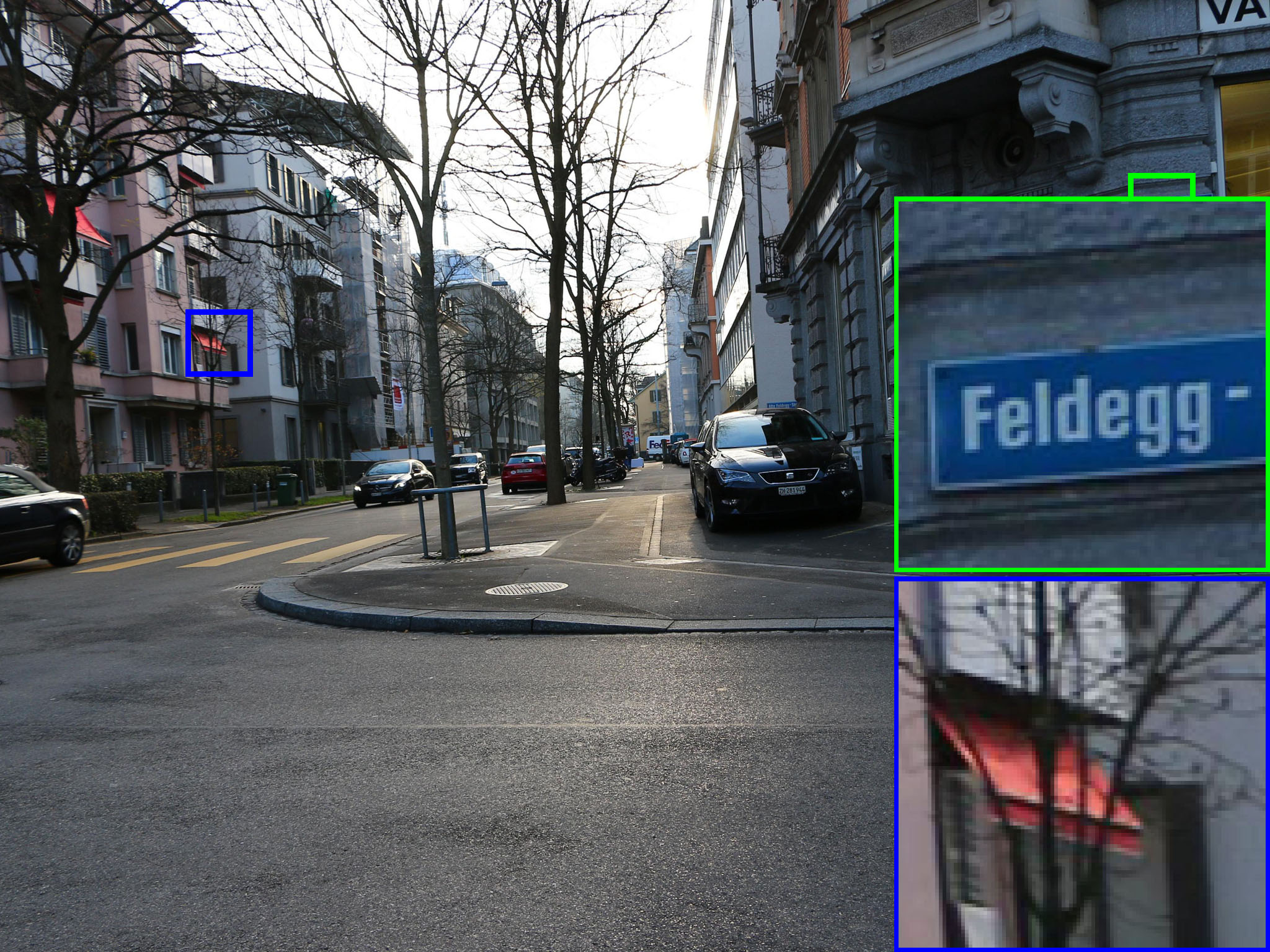}
 \caption{From left to right, top to bottom: original iPhone photo and the same image after applying, respectively: APE, Dong~et al.~\cite{Dong2014}, Johnson~et al.~\cite{Johnson2016}, our generator network, and the corresponding DSLR image.}
 \label{fig:all_enhancements1}
\end{figure*}

\section{Experiments}
\label{sec:experiments}

Our general goal to ``improve image quality'' is subjective and hard to evaluate quantitatively.
We suggest a set of tools and methods from the literature that are most relevant to our problem. We use them, as well as our proposed method, on a set of test images taken by mobile devices and compare how close the results are to the DSRL shots.

\begin{table*}[th!]
\caption{Average PSNR/SSIM results on DPED test images.\label{tab:scores}}
\centering
\begin{tabular}{l||*{2}{c}|*{2}{c}|*{2}{c}|*{2}{c}}
Phone &  \multicolumn{2}{|c|}{APE} & \multicolumn{2}{|c|}{Dong et al.~\cite{Dong2014}} & \multicolumn{2}{|c|}{Johnson et al.~\cite{Johnson2016}} &\multicolumn{2}{|c}{\textbf{Ours}} \\
 & PSNR & SSIM & PSNR & SSIM & PSNR & SSIM & PSNR & SSIM \\
\hline
iPhone & 17.28 & 0.8631 & 19.27 & 0.8992 & \textBF{20.32} & 0.9161 & 20.08 & \textBF{0.9201} \\
BlackBerry & 18.91 & 0.8922 & 18.89 & 0.9134 & \textBF{20.11} & 0.9298 & 20.07 & \textBF{0.9328} \\
Sony & 19.45 & 0.9168 & 21.21 & 0.9382 & 21.33 & 0.9434 & \textBF{21.81} & \textBF{0.9437} \\
\end{tabular}
\end{table*}

In section~\ref{ssc:methods}, we present the methods we compare to.
Then we present both objective and subjective evaluations: the former w.r.t. the ground truth reference (i.e., the DSLR images) in section~\ref{ssc:exp:quantitative}, the latter with no-reference subjective quality scores in section~\ref{ssc:exp:subjective}. Finally, section~\ref{ssc:exp:limitations} analyzes the limitations of the proposed solution.

\subsection{Benchmark methods}
\label{ssc:methods}
In addition to our proposed photo enhancement solution, we compare with the following tools and methods.

\textbf{Apple Photo Enhancer (APE)} is a commercial product known to generate among the best visual results, while the algorithm is unpublished. We trigger the method using the automatic \textit{Enhance} function from the \textit{Photos} app.
It performs image improvement without taking any parameters.

\textbf{Dong~et al.~\cite{Dong2014}} is a fundamental baseline super-resolution method, thus addredding a task related to end-to-end image-to-image mapping.
Hence we chose it to apply on our task and compare with.
The method relies on a standard 3-layer CNN and MSE loss function and maps from low resolution / corrupted image to the restored image.

\textbf{Johnson~et al.~\cite{Johnson2016}} is one of the latest state of the art in photo-realistic super-resolution and style transferring tasks. The method is based on a deep residual network (with four residual blocks, each consisting of two convolutional layers) that is trained to minimize a VGG-based loss function.

\textbf{Manual enhancement}. We asked a graphical artist to enhance color, sharpness and general look-and-feel of 9 images using professional software (Adobe Photoshop CS6).
A time limit of one workday was given, so as to simulate a realistic scenario.
Figure~\ref{fig:all_enhancements1} illustrates the ensemble of enhancement methods we consider for comparison in our experiments.
Dong~et al.~\cite{Dong2014} and Johnson~et al.~\cite{Johnson2016} are trained using the same train image pairs as for our solution for each of the smartphones from the DPED dataset.

\begin{figure*}[t]
\setlength{\tabcolsep}{1pt}
\resizebox{\linewidth}{!}
{
\begin{tabular}{cccc}
BlackBerry & BlackBerry & Sony & Sony \\
   \includegraphics[width=0.245\linewidth]{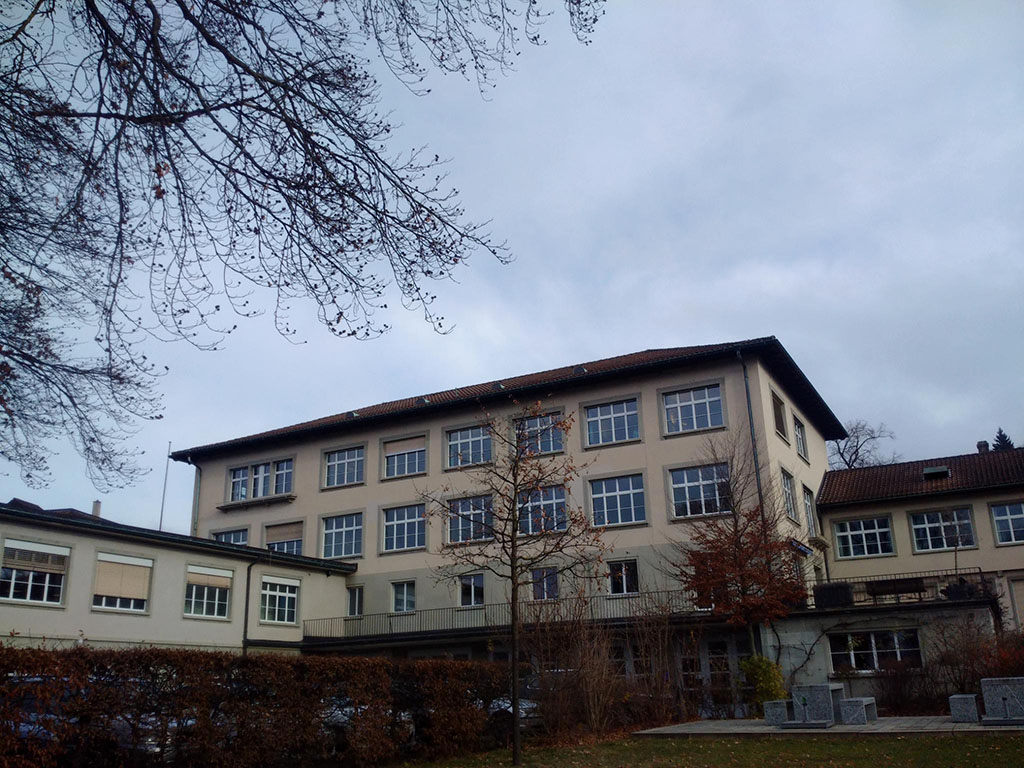}&
   \includegraphics[width=0.245\linewidth]{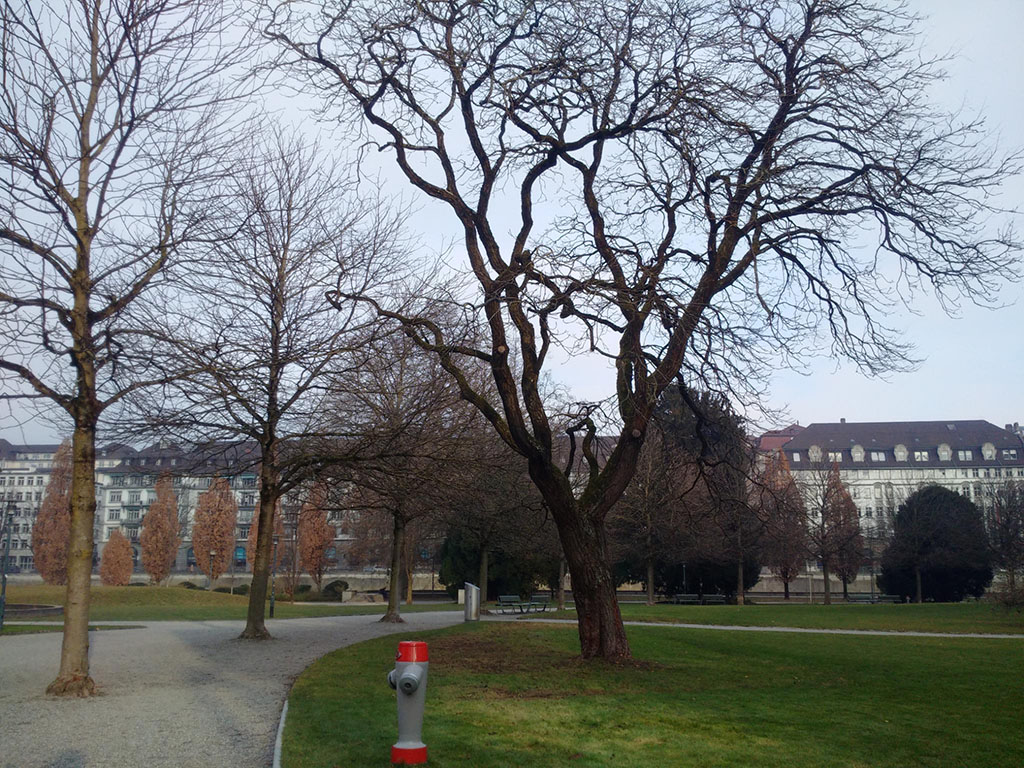}&
   \includegraphics[width=0.245\linewidth]{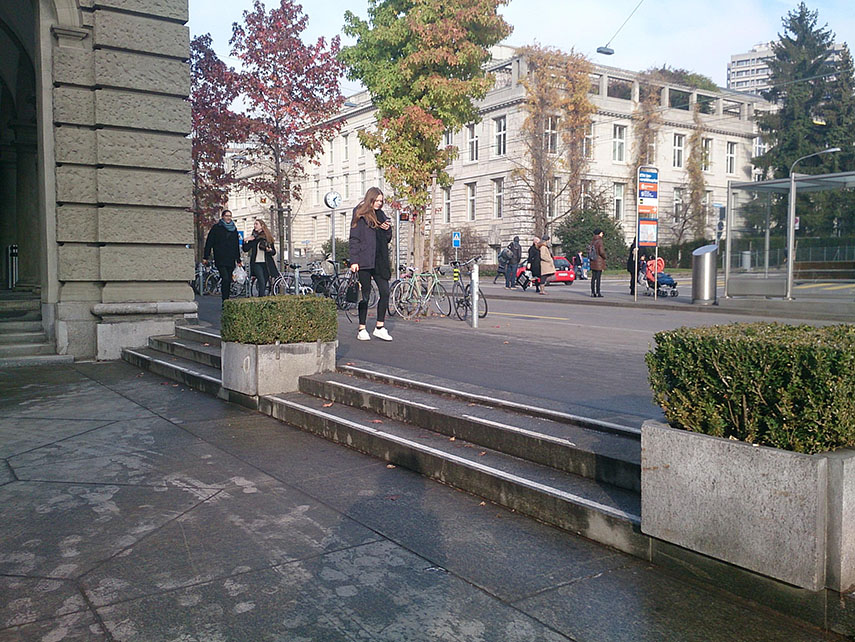}&
   \includegraphics[width=0.245\linewidth]{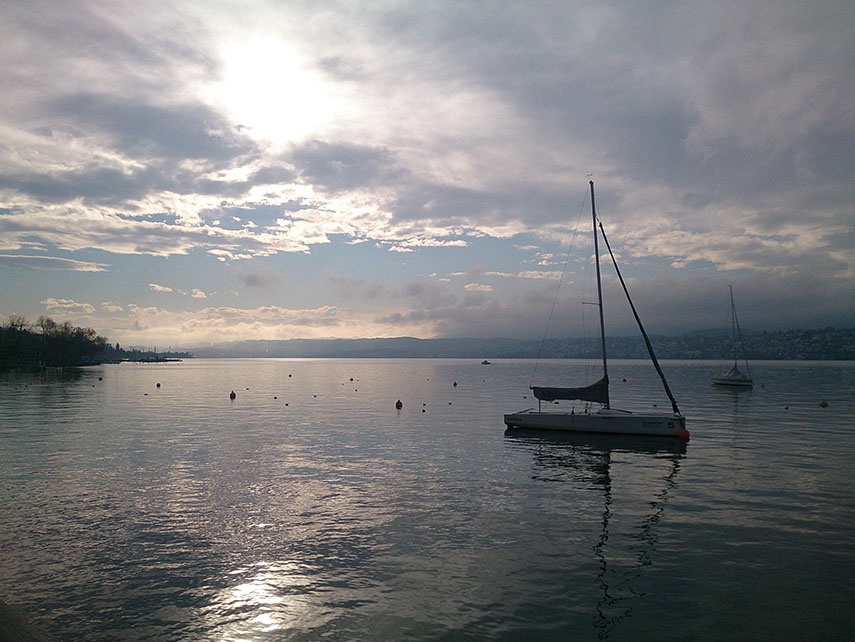}
   \\
   \includegraphics[width=0.245\linewidth]{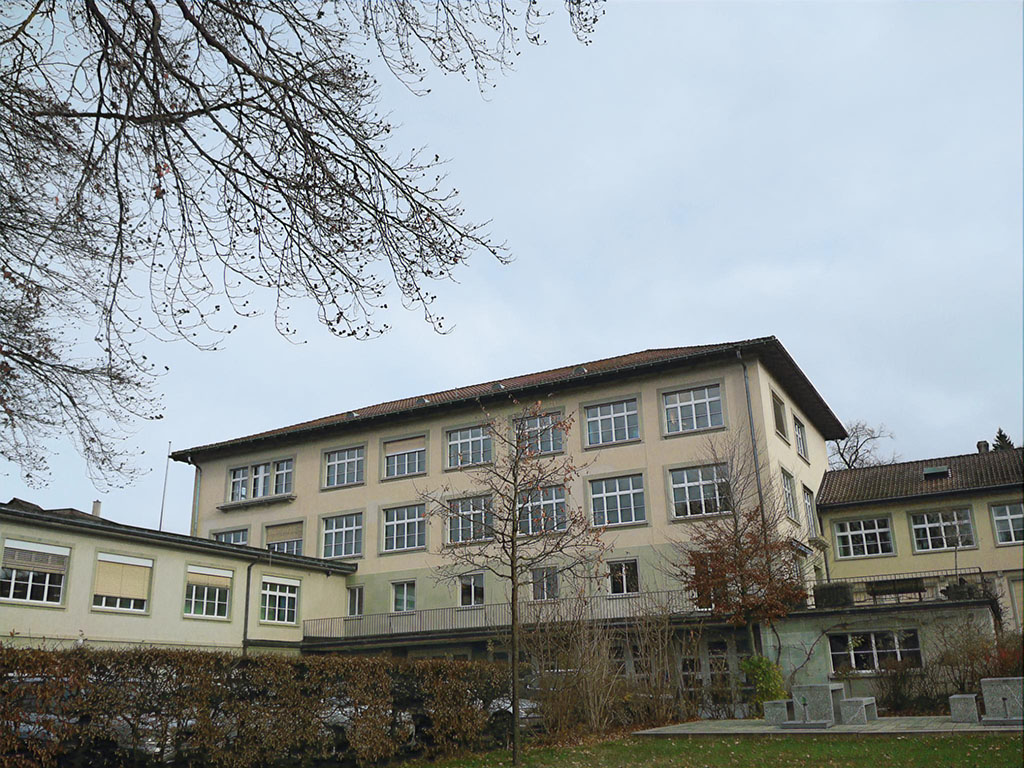}&
   \includegraphics[width=0.245\linewidth]{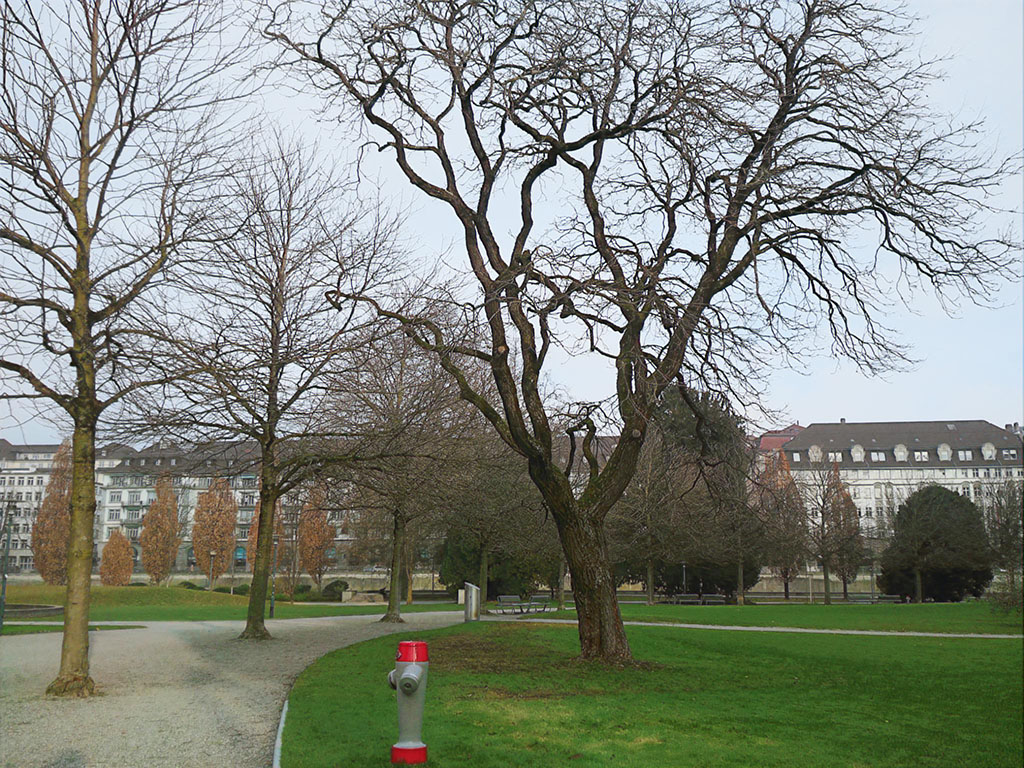}&
   \includegraphics[width=0.245\linewidth]{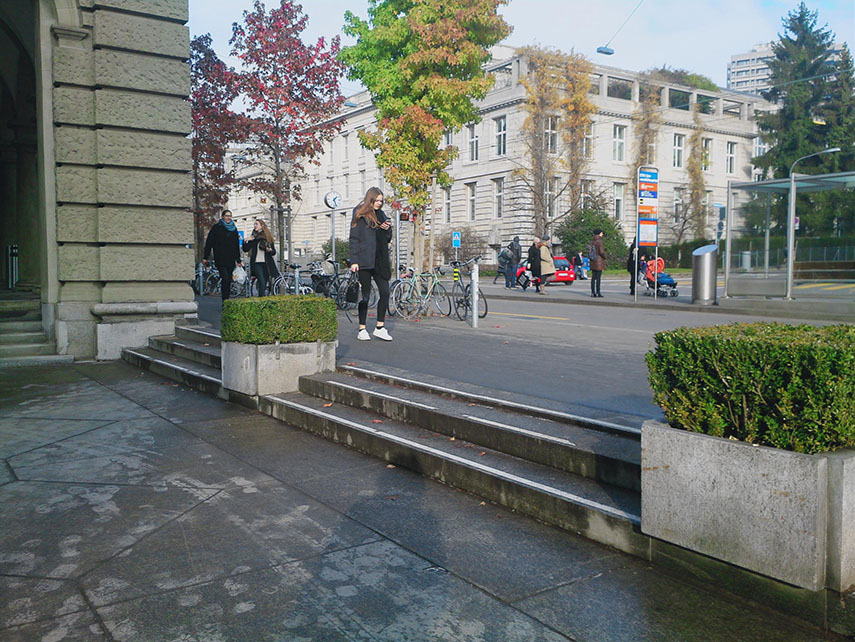}&
   \includegraphics[width=0.245\linewidth]{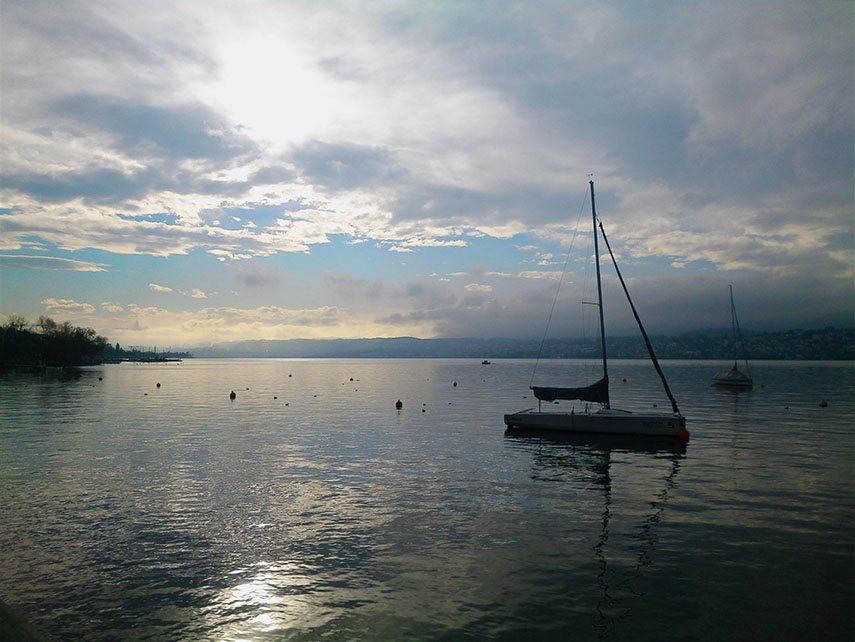} \\
\end{tabular}
}
\vspace{-0.2cm}
 \caption{Four examples of original (top) vs. enhanced (bottom) images captured by BlackBerry and Sony cameras.}
 \label{fig:enhancedexamples}
 \vspace{-0.1cm}
\end{figure*}

\subsection{Quantitative evaluation}
\label{ssc:exp:quantitative}
We first quantitatively compare APE, Dong~et al.~\cite{Dong2014}, Johnson~et al.~\cite{Johnson2016} and our method on the task of mapping photos from three low-end cameras to the high-quality DSLR (Canon) images and report the results in Table~\ref{tab:scores}.
As such, we do not evaluate global image quality but, rather, we measure resemblance to a reference (the ground truth DSLR image).
We use classical distance metrics, namely PSNR and SSIM scores: the former measures signal distortion w.r.t. the reference, the latter measures structural similarity which is known to be a strong cue for perceived quality~\cite{SSIM}.
First, one can note that our method is the best in terms of SSIM, at the same time producing images that are cleaner and sharper, thus perceptually performs the best.
On PSNR terms, our method competes with the state of the art: it slightly improves or worsens depending on the dataset, i.e., on the actual phone used.
Alignment issues could be responsible for these minor variations, and thus we consider Johnson~et al.'s method~\cite{Dong2014} and ours equivalent here, while outperforming other methods.
In Fig.~\ref{fig:all_enhancements1} we show visual results comparing to the source photo (iPhone) and the target DSLR photo (Canon). More results are in the supplementary material.

\subsection{User study}
\label{ssc:exp:subjective}
Our goal is to produce DSLR-quality images for the end user of smartphone cameras.
To measure overall quality we designed a no-reference user study where subjects are repeatedly asked to choose the better looking picture out of a displayed pair. Users were instructed to ignore precise picture composition errors (e.g., field of view, perspective variation, \textit{etc.}). There was no time limit given to the participants, images were shown in full resolution and the users were allowed to zoom in and out at will.
In this setting, we did the following pairwise comparisons (every group of experiments contains 3 classes of pictures, the users were shown all possible pairwise combinations of these classes):

\smallskip

\noindent\textbf{(i)} Comparison between:
\vspace{-0.1cm}
\begin{itemize}
\setlength\itemsep{-0.3em}
\item original low-end phone photos,
\item DSLR photos,
\item photos enhanced by our proposed method.
\end{itemize}
\vspace{-0.07cm}

\begin{figure*}[t!]
\centering
   \includegraphics[width=\linewidth]{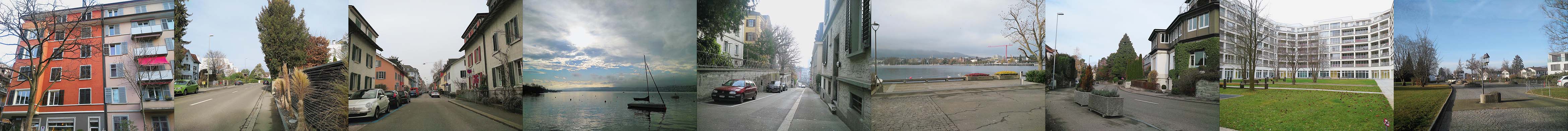}
   \vspace{-0.3cm}
   \caption{The 9 scenes shown to the participants of the user study. Here: BlackBerry images enhanced using our technique.}
\label{fig:strip}
\vspace{-0.2cm}
\end{figure*}

At every question, the user is shown two pictures from different categories (original, DSLR or enhanced).
9 scenes were used for each phone (e.g., see Fig.~\ref{fig:strip}).
In total, there are 27 questions for every phone, thus 81 in total.

\smallskip

\noindent\textbf{(ii)} Additionally, we compared (iPhone images only):
\vspace{-0.1cm}
\begin{itemize}
\setlength\itemsep{-0.3em}
\item photos enhanced by the proposed method,
\item photos enhanced manually (by a professional),
\item photos enhanced by APE.
\end{itemize}
\vspace{-0.09cm}

We again considered 9 images that resulted in 27 binary selection questions.
Thus, in total the study consists of 108 binary questions.
All pairs are shuffled randomly for every subject, as is the sequence of displayed images.
42 subjects unaware of the goal of this work participated.
They are mostly young scientists with a computer science background.

\begin{figure*}[th!]
    \centering
    \setlength{\tabcolsep}{0.5cm}
    \resizebox{\linewidth}{!}
    {
    \begin{tabular}{cccc}
    \includegraphics[trim={0cm 0cm 0cm 0cm}]{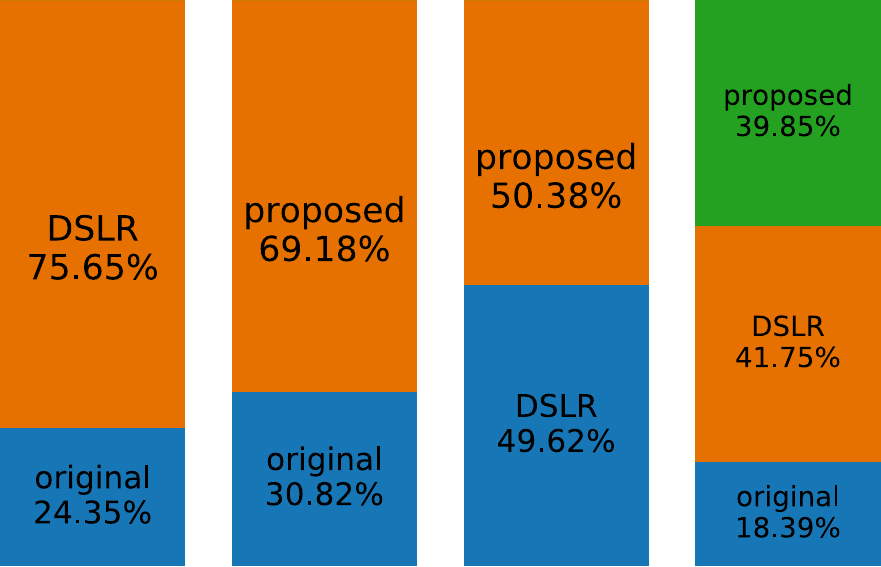}&
    \includegraphics[trim={0cm 0cm 0cm 0cm}]{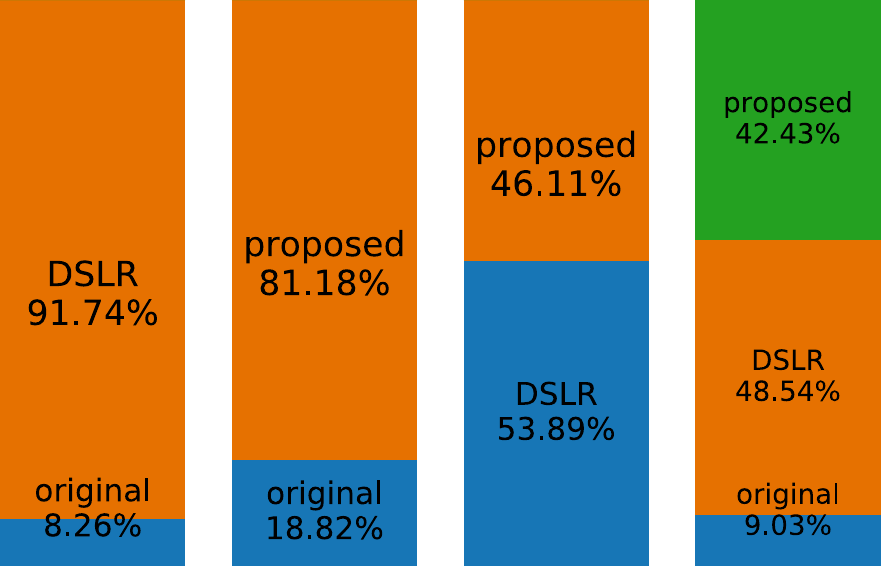}&
    \includegraphics[trim={0cm 0cm 0cm 0cm}]{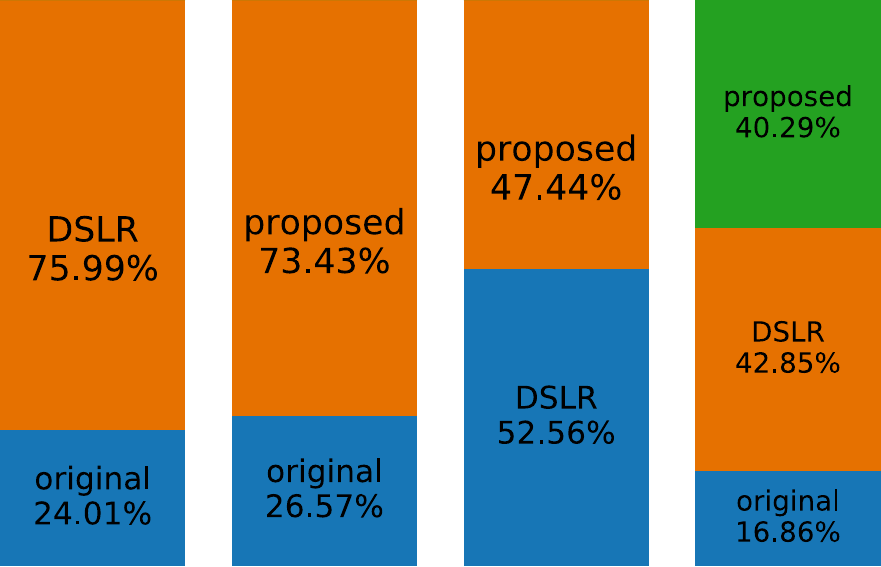}&
    \includegraphics[trim={0cm 0cm 0cm 0cm}]{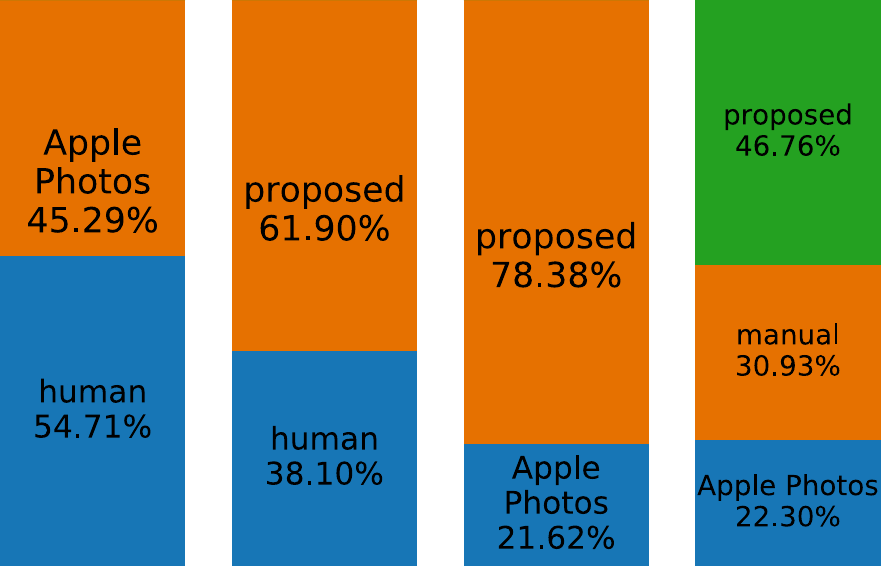} \\
    {\Large (a) BlackBerry phone} &
    {\Large (b) iPhone} &
    {\Large (c) Sony phone} &
    {\Large (d) Enhanced iPhone pictures}\\
    \end{tabular}
    }
    \vspace{-0.2cm}
    \caption{User study: results of pairwise comparisons. In every subfigure, the first three bars show the result of the pairwise experiments, while the last bar shows the distribution of the aggregated scores.}
    \label{fig:userstudy}
    \vspace{-0.1cm}
\end{figure*}

Figure~\ref{fig:userstudy} shows results: for every experiment the first 3 bars show the results of the pairwise comparison averaged over the 9 images shown, while the last bar shows the fraction of cases when the selected method was chosen over all experiments.

The subfigures~\ref{fig:userstudy}a-c show the results of enhancing photos from 3 different mobile devices. It can be seen that in all cases both pictures taken with a DSLR as well as pictures enhanced by the proposed CNN are picked much more often than the original ones taken with the mobile devices. When subjects are asked to select the better picture among the DSLR-picture and our enhanced picture, the choice is almost random (see the third bar in subfigures~\ref{fig:userstudy}a-c).
This means that the quality difference is inexistent or indistinguishable, and users resort to chance.

Subfigure~\ref{fig:userstudy}d shows user choices among our method, human artist work, and APE.
Although human enhancement turns out to be slightly preferred to the automatic APE, the images enhanced by our method are picked more often, outperforming even manual retouching.

We can conclude that our results are of on pair quality compared to DSLR images, while starting from low quality phone cameras. The human subjects are unable to distinguish between them -- the preferences are equally distributed.

\begin{figure*}[th!]
\centering
\setlength{\tabcolsep}{1pt}
\resizebox{\linewidth}{!}
{
\begin{tabular}{cccc}
\includegraphics[width=0.245\linewidth]{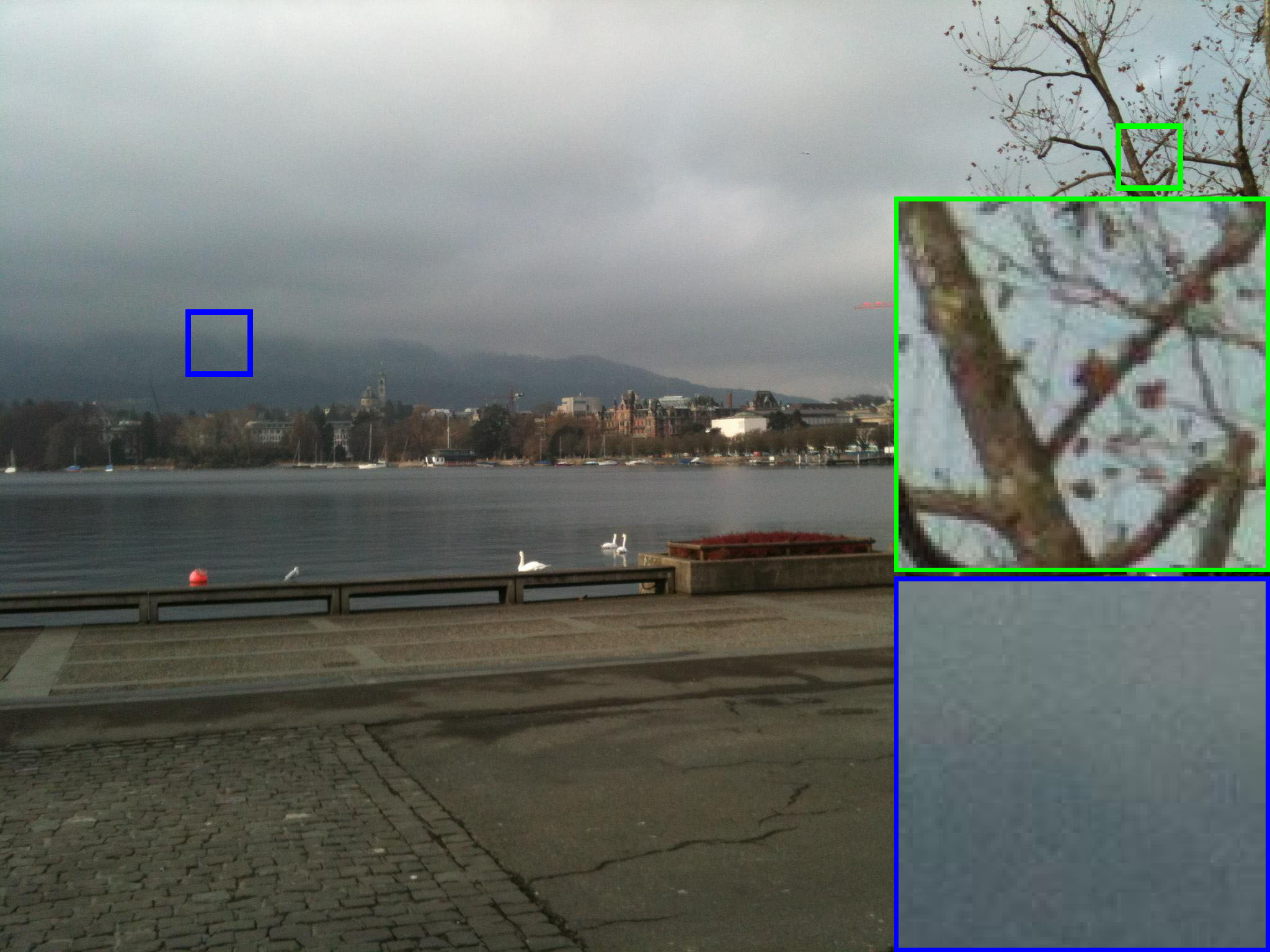}&
\includegraphics[width=0.245\linewidth]{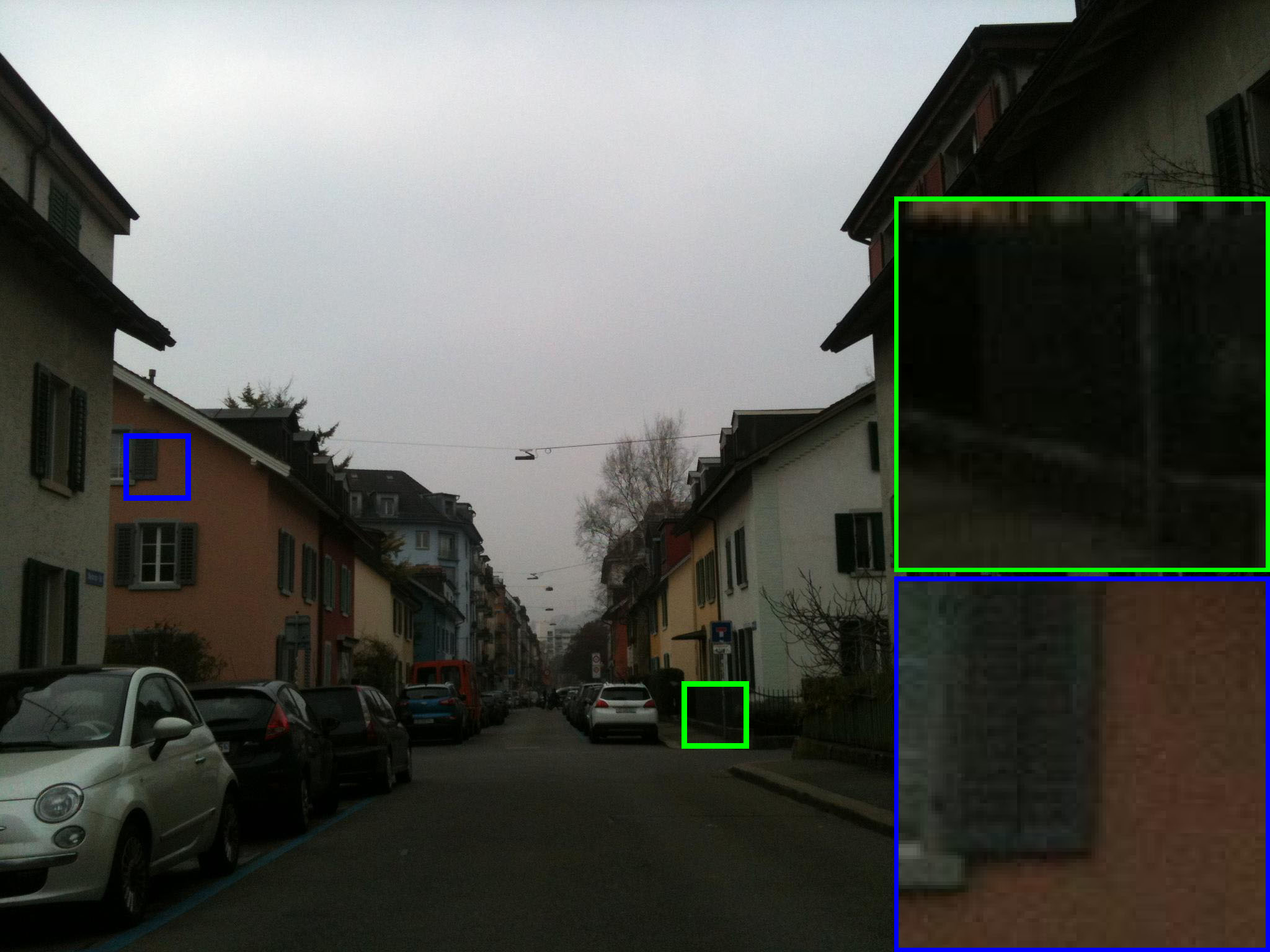}&
\includegraphics[width=0.245\linewidth]{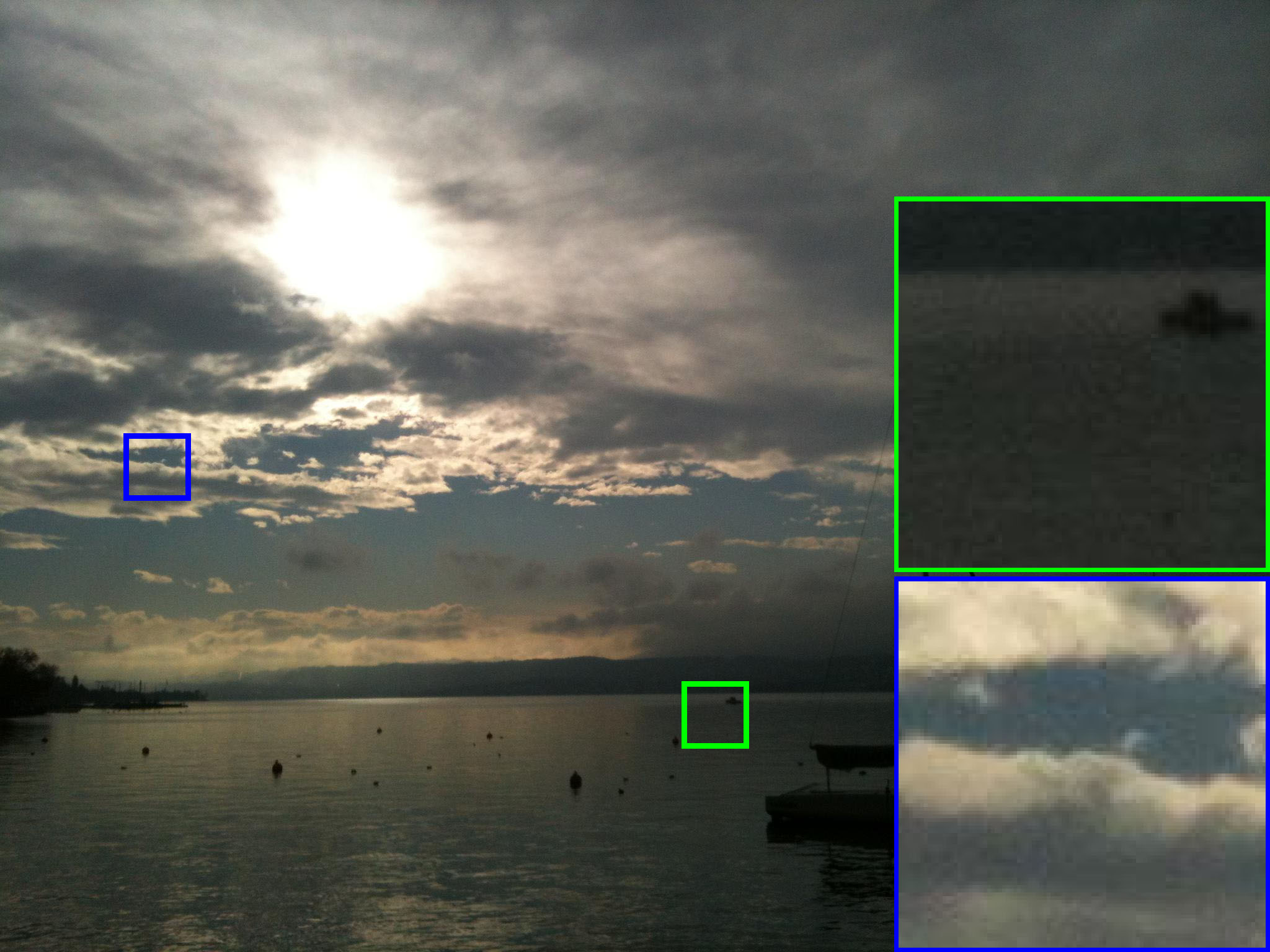}&
\includegraphics[width=0.245\linewidth]{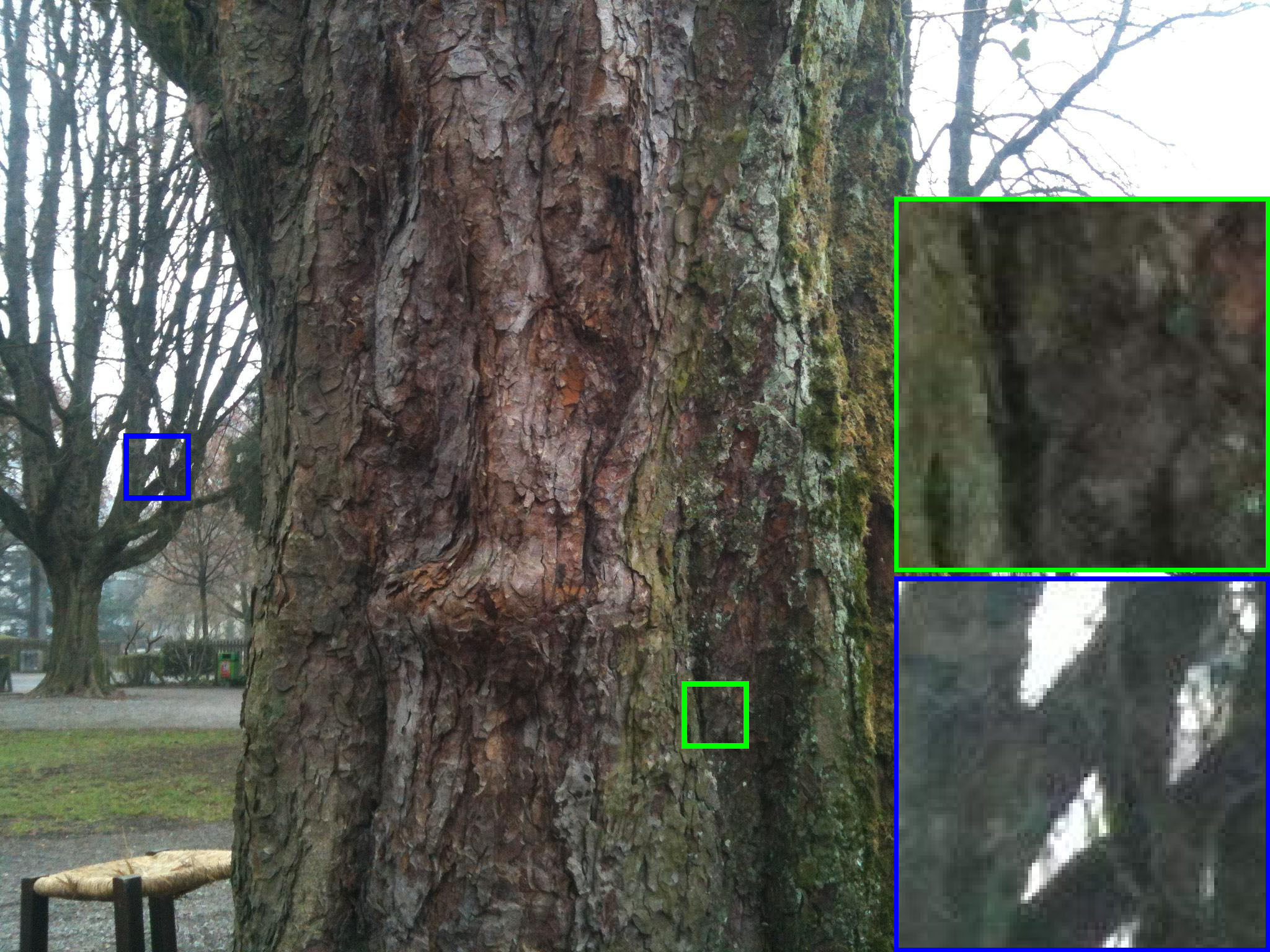}\\
\includegraphics[width=0.245\linewidth]{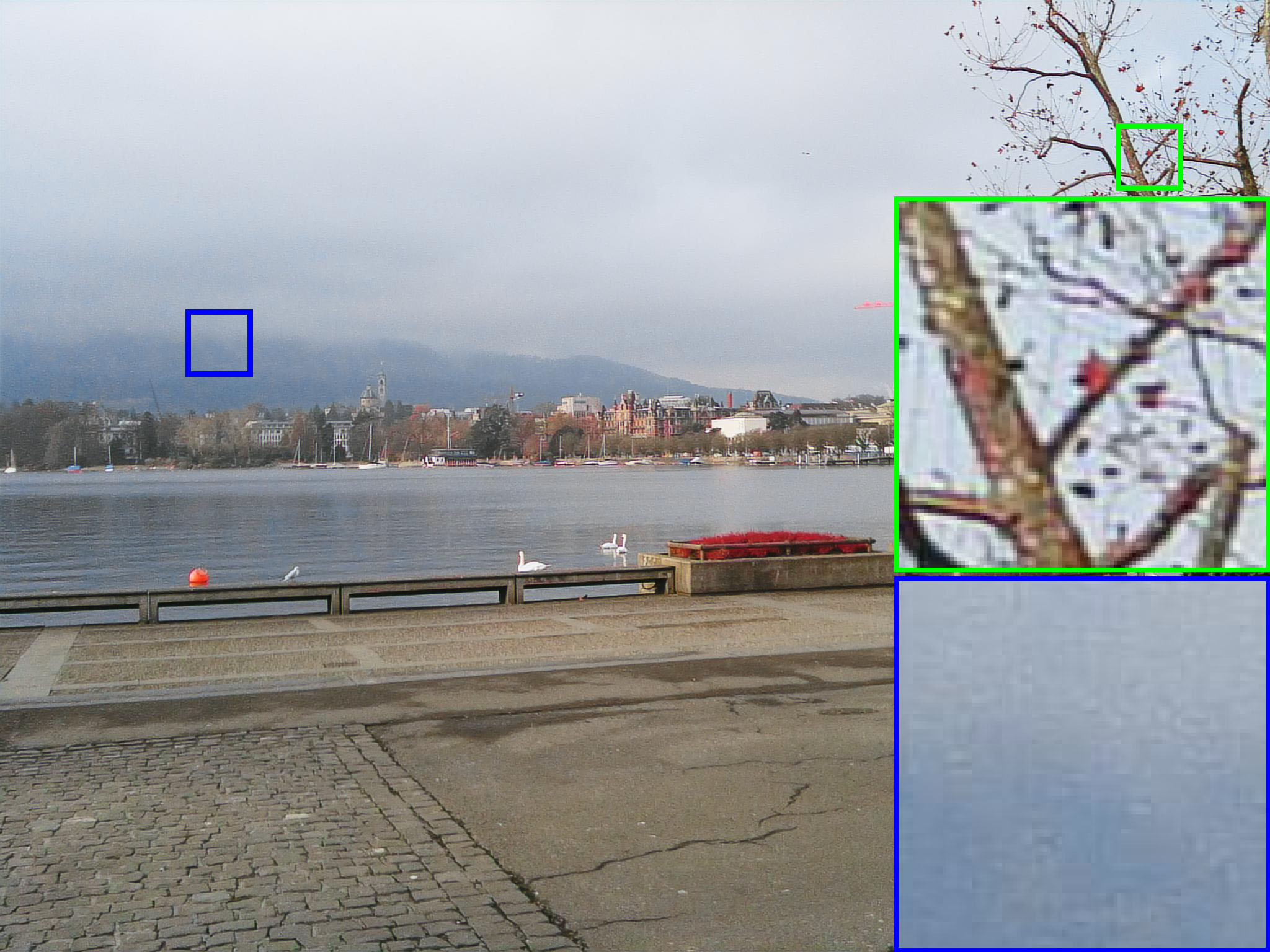}&
\includegraphics[width=0.245\linewidth]{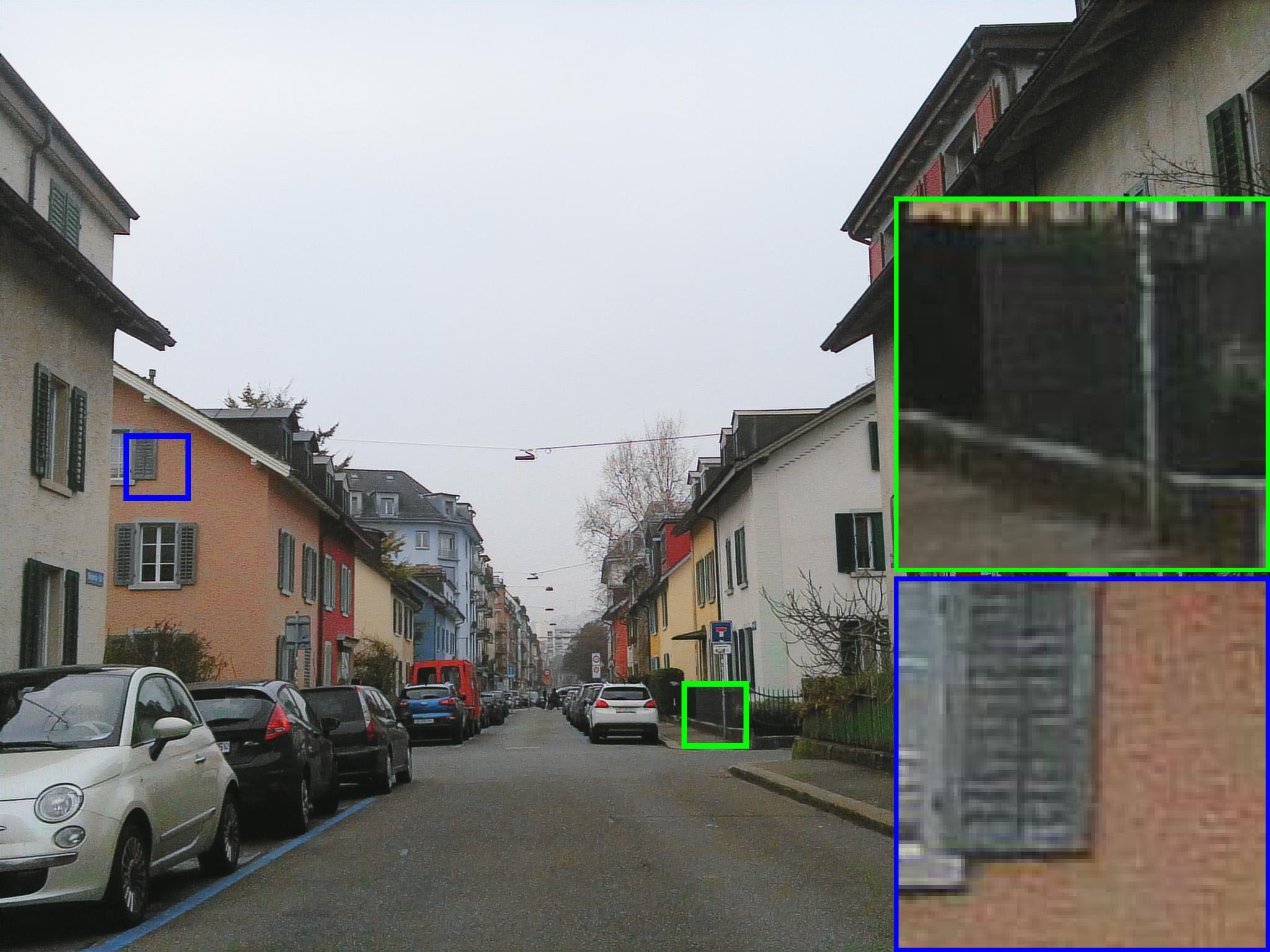}&
\includegraphics[width=0.245\linewidth]{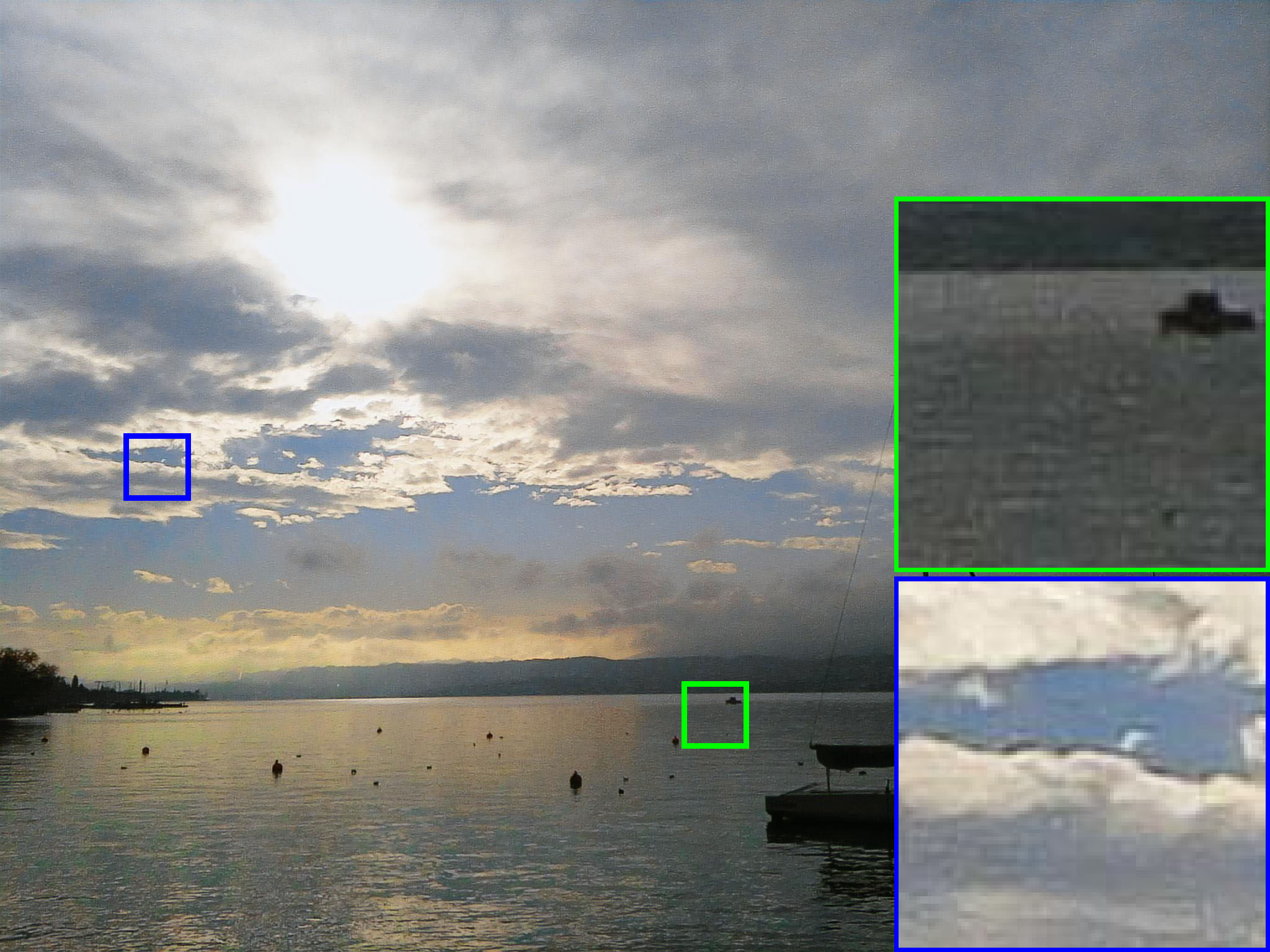}&
\includegraphics[width=0.245\linewidth]{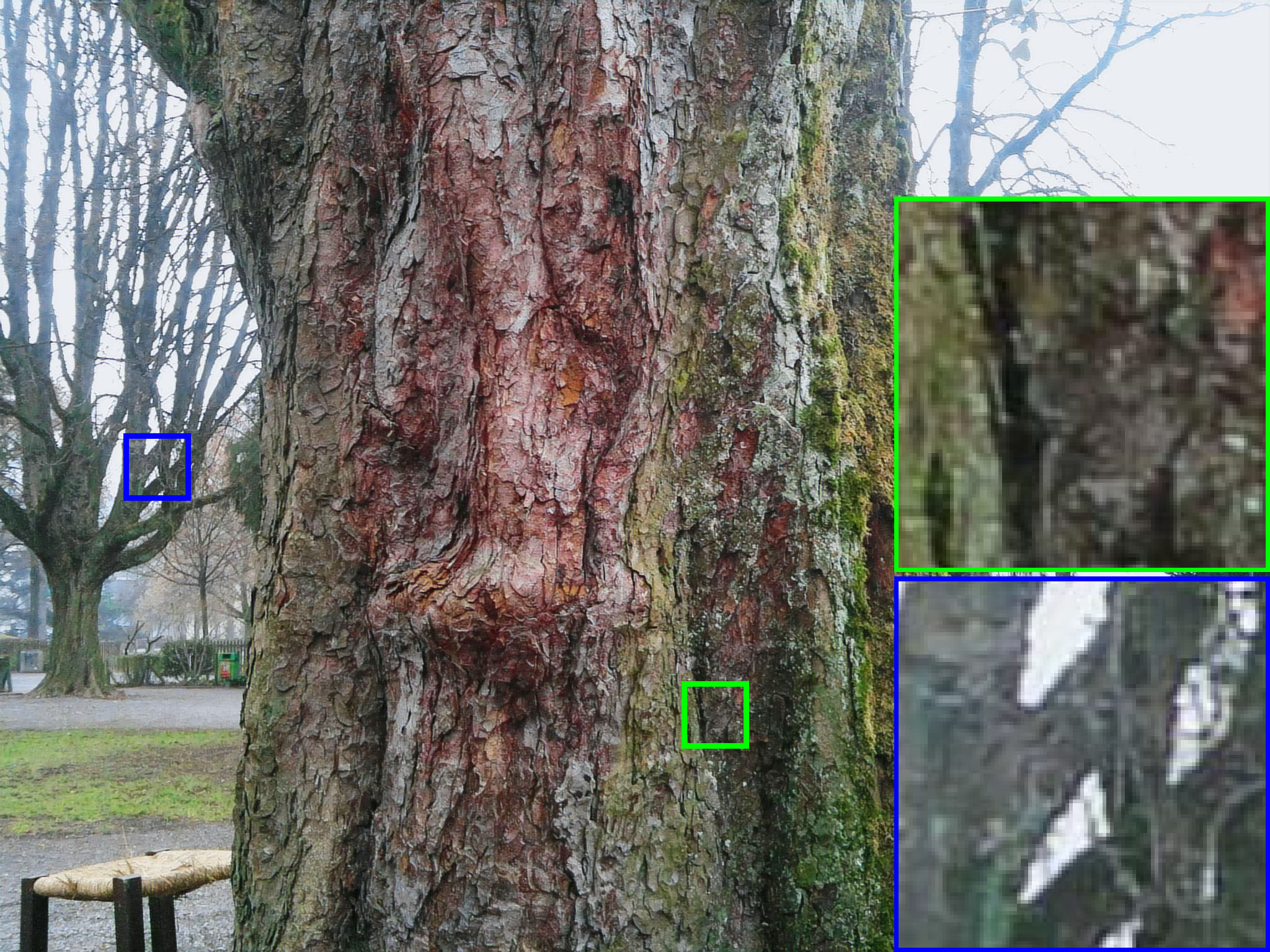}\\
\end{tabular}
}
\vspace{-0.3cm}
 \caption{Typical artifacts generated by our method (2nd row) compared with original iPhone images (1st row)}
 \label{fig:artifacts}
 \vspace{-0.2cm}
\end{figure*}

\subsection{Limitations}
\label{ssc:exp:limitations}

Since the proposed enhancement process is fully-automated, some flaws are inevitable. Two typical artifacts that can appear on the processed images are color deviations (see ground/mountains in first image of Fig.~\ref{fig:artifacts}) and too high contrast levels (second image).
Although they often cause rather plausible visual effects, in some situations this can lead to content changes that may look artificial, i.e. greenish asphalt in the second image of Fig.~\ref{fig:artifacts}. Another notable problem is noise amplification~-- due to the nature of GANs, they can effectively restore high frequency-components.
However, high-frequency noise is emphasized too. Fig.~\ref{fig:artifacts} (2nd and 3rd images) shows that a high noise in the original image is amplified in the enhanced image.
Note that this noise issue occurs mostly on the lowest-quality photos (i.e., from the iPhone), not on the better phone cameras.

Finally, the need of a strong supervision in the form of matched source/target training image pairs makes the process tedious to repeat for other cameras. To overcome this, we propose a weakly-supervised approach in~\cite{ignatov2017wespe} that does not require the mentioned correspondence.

\section{Conclusions}
\label{sec:conclusions}
We proposed a photo enhancement solution to effectively transform cameras from common smartphones into high quality DSLR cameras.
Our end-to-end deep learning approach uses a composite perceptual error function that combines content, color and texture losses.
To train and evaluate our method we introduced DPED -- a large-scale dataset that consists of real photos captured from three different phones and one high-end reflex camera, and suggested an efficient way of calibrating the images so that they are suitable for image-to-image learning.
Our quantitative and qualitative assessments reveal that the enhanced images demonstrate a quality comparable to DSLR-taken photos, and the method itself can be applied to cameras of various quality levels.
\\

\noindent\textbf{Acknowledgments.}
Work supported by the ETH Zurich General Fund (OK), Toyota via the project TRACE-Zurich, the ERC grant VarCity, and an NVidia GPU grant.

{\normalsize
\bibliographystyle{ieee}

\begin{thebibliography}{10}\itemsep=-1pt

\bibitem{TV2005}
H.~A. Aly and E.~Dubois.
\newblock Image up-sampling using total-variation regularization with a new
  observation model.
\newblock {\em IEEE Transactions on Image Processing}, 14(10):1647--1659, Oct
  2005.

\bibitem{Cai2016}
B.~Cai, X.~Xu, K.~Jia, C.~Qing, and D.~Tao.
\newblock Dehazenet: An end-to-end system for single image haze removal.
\newblock {\em IEEE Transactions on Image Processing}, 25(11):5187--5198, Nov
  2016.

\bibitem{Cheng2015}
Z.~Cheng, Q.~Yang, and B.~Sheng.
\newblock Deep colorization.
\newblock In {\em The IEEE International Conference on Computer Vision (ICCV)},
  December 2015.

\bibitem{Dong2014}
C.~Dong, C.~C. Loy, K.~He, and X.~Tang.
\newblock {\em Learning a Deep Convolutional Network for Image
  Super-Resolution}, pages 184--199.
\newblock Springer International Publishing, Cham, 2014.

\bibitem{GAN}
I.~Goodfellow, J.~Pouget-Abadie, M.~Mirza, B.~Xu, D.~Warde-Farley, S.~Ozair,
  A.~Courville, and Y.~Bengio.
\newblock Generative adversarial nets.
\newblock In Z.~Ghahramani, M.~Welling, C.~Cortes, N.~D. Lawrence, and K.~Q.
  Weinberger, editors, {\em Advances in Neural Information Processing Systems
  27}, pages 2672--2680. Curran Associates, Inc., 2014.

\bibitem{Hradis2015}
M.~Hradi{\v{s}}, J.~Kotera, P.~Zem{\v{c}}{\'{i}}k, and F.~{\v{S}}roubek.
\newblock Convolutional neural networks for direct text deblurring.
\newblock In {\em Proceedings of BMVC 2015}. The British Machine Vision
  Association and Society for Pattern Recognition, 2015.

\bibitem{ignatov2017wespe}
A.~Ignatov, N.~Kobyshev, R.~Timofte, K.~Vanhoey, and L.~Van~Gool.
\newblock Wespe: Weakly supervised photo enhancer for digital cameras.
\newblock 2017.

\bibitem{Isola2016}
P.~Isola, J.-Y. Zhu, T.~Zhou, and A.~A. Efros.
\newblock Image-to-image translation with conditional adversarial networks.
\newblock {\em arxiv}, 2016.

\bibitem{Johnson2016}
J.~Johnson, A.~Alahi, and L.~Fei-Fei.
\newblock {\em Perceptual Losses for Real-Time Style Transfer and
  Super-Resolution}, pages 694--711.
\newblock Springer International Publishing, Cham, 2016.

\bibitem{Kim2016}
J.~Kim, J.~K. Lee, and K.~M. Lee.
\newblock Accurate image super-resolution using very deep convolutional
  networks.
\newblock In {\em 2016 IEEE Conference on Computer Vision and Pattern
  Recognition (CVPR)}, pages 1646--1654, June 2016.

\bibitem{Adam14}
D.~P. Kingma and J.~Ba.
\newblock Adam: {A} method for stochastic optimization.
\newblock {\em CoRR}, abs/1412.6980, 2014.

\bibitem{Ledig2016}
C.~Ledig, L.~Theis, F.~Huszar, J.~Caballero, A.~Cunningham, A.~Acosta,
  A.~Aitken, A.~Tejani, J.~Totz, Z.~Wang, and W.~Shi.
\newblock Photo-realistic single image super-resolution using a generative
  adversarial network.
\newblock In {\em The IEEE Conference on Computer Vision and Pattern
  Recognition (CVPR)}, July 2017.

\bibitem{Lee2016}
J.-Y. Lee, K.~Sunkavalli, Z.~Lin, X.~Shen, and I.~So~Kweon.
\newblock Automatic content-aware color and tone stylization.
\newblock In {\em The IEEE Conference on Computer Vision and Pattern
  Recognition (CVPR)}, June 2016.

\bibitem{Ling2016}
Z.~Ling, G.~Fan, Y.~Wang, and X.~Lu.
\newblock Learning deep transmission network for single image dehazing.
\newblock In {\em 2016 IEEE International Conference on Image Processing
  (ICIP)}, pages 2296--2300, Sept 2016.

\bibitem{Lowe2004}
D.~G. Lowe.
\newblock Distinctive image features from scale-invariant keypoints.
\newblock {\em International Journal of Computer Vision}, 60(2):91--110, 2004.

\bibitem{Mao2016}
X.~Mao, C.~Shen, and Y.-B. Yang.
\newblock Image restoration using very deep convolutional encoder-decoder
  networks with symmetric skip connections.
\newblock In D.~D. Lee, M.~Sugiyama, U.~V. Luxburg, I.~Guyon, and R.~Garnett,
  editors, {\em Advances in Neural Information Processing Systems 29}, pages
  2802--2810. Curran Associates, Inc., 2016.

\bibitem{Ren2016}
W.~Ren, S.~Liu, H.~Zhang, J.~Pan, X.~Cao, and M.-H. Yang.
\newblock {\em Single Image Dehazing via Multi-scale Convolutional Neural
  Networks}, pages 154--169.
\newblock Springer International Publishing, Cham, 2016.

\bibitem{Shi2016}
W.~Shi, J.~Caballero, F.~Huszar, J.~Totz, A.~P. Aitken, R.~Bishop, D.~Rueckert,
  and Z.~Wang.
\newblock Real-time single image and video super-resolution using an efficient
  sub-pixel convolutional neural network.
\newblock In {\em The IEEE Conference on Computer Vision and Pattern
  Recognition (CVPR)}, June 2016.

\bibitem{Svoboda2016}
P.~Svoboda, M.~Hradis, D.~Barina, and P.~Zemc{\'{\i}}k.
\newblock Compression artifacts removal using convolutional neural networks.
\newblock {\em CoRR}, abs/1605.00366, 2016.

\bibitem{Timofte2015}
R.~Timofte, V.~De~Smet, and L.~Van~Gool.
\newblock {\em A+: Adjusted Anchored Neighborhood Regression for Fast
  Super-Resolution}, pages 111--126.
\newblock Springer International Publishing, Cham, 2015.

\bibitem{vlfeat}
A.~Vedaldi and B.~Fulkerson.
\newblock {VLFeat}: An open and portable library of computer vision algorithms,
  2008.

\bibitem{SSIM}
Z.~Wang, A.~C. Bovik, H.~R. Sheikh, and E.~P. Simoncelli.
\newblock Image quality assessment: from error visibility to structural
  similarity.
\newblock {\em IEEE Transactions on Image Processing}, 13(4):600--612, April
  2004.

\bibitem{Yan2016}
Z.~Yan, H.~Zhang, B.~Wang, S.~Paris, and Y.~Yu.
\newblock Automatic photo adjustment using deep neural networks.
\newblock {\em ACM Trans. Graph.}, 35(2):11:1--11:15, Feb. 2016.

\bibitem{Yang2016}
W.~Yang, R.~T. Tan, J.~Feng, J.~Liu, Z.~Guo, and S.~Yan.
\newblock Joint rain detection and removal via iterative region dependent
  multi-task learning.
\newblock {\em CoRR}, abs/1609.07769, 2016.

\bibitem{Yuan2012}
L.~Yuan and J.~Sun.
\newblock {\em Automatic Exposure Correction of Consumer Photographs}, pages
  771--785.
\newblock Springer Berlin Heidelberg, Berlin, Heidelberg, 2012.

\bibitem{Kai2016}
K.~Zhang, W.~Zuo, Y.~Chen, D.~Meng, and L.~Zhang.
\newblock Beyond a gaussian denoiser: Residual learning of deep {CNN} for image
  denoising.
\newblock {\em IEEE Transactions on Image Processing}, 2017.

\bibitem{Richard2016}
R.~Zhang, P.~Isola, and A.~A. Efros.
\newblock Colorful image colorization.
\newblock {\em ECCV}, 2016.

\bibitem{Zhang2016}
X.~Zhang and R.~Wu.
\newblock Fast depth image denoising and enhancement using a deep convolutional
  network.
\newblock In {\em 2016 IEEE International Conference on Acoustics, Speech and
  Signal Processing (ICASSP)}, pages 2499--2503, March 2016.

\bibitem{Zhou2015}
E.~Zhou, H.~Fan, Z.~Cao, Y.~Jiang, and Q.~Yin.
\newblock Learning face hallucination in the wild.
\newblock In {\em Proceedings of the Twenty-Ninth AAAI Conference on Artificial
  Intelligence}, AAAI'15, pages 3871--3877. AAAI Press, 2015.

\end{thebibliography}

}

\clearpage

\onecolumn

\centering

\section{Appendix. Results of the proposed method: iPhone\protect\footnote{All visual results for iPhone are available at \url{http://people.ee.ethz.ch/~ihnatova/dped_iphone.html}}}
\label{sec:results_iPhone}

\begin{figure*}[h!]
\setlength{\tabcolsep}{1pt}
\centering
\begin{tabular}{cc}
iPhone original & Enhanced with our method\\
   \includegraphics[width=0.36\linewidth]{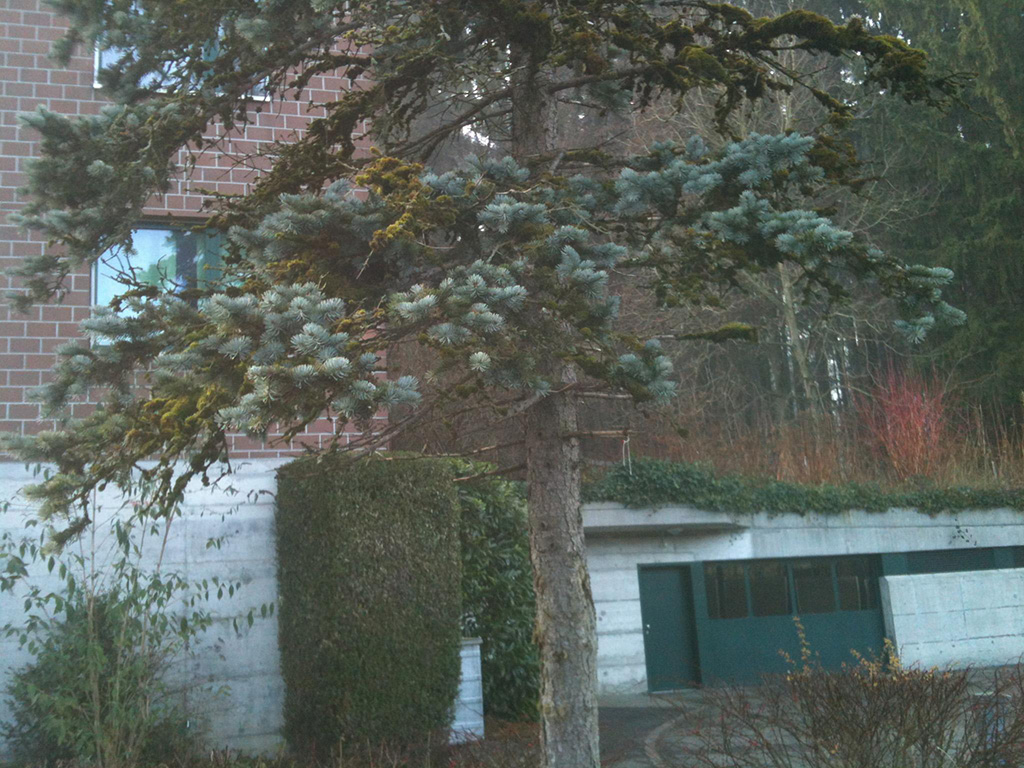}&
   \includegraphics[width=0.36\linewidth]{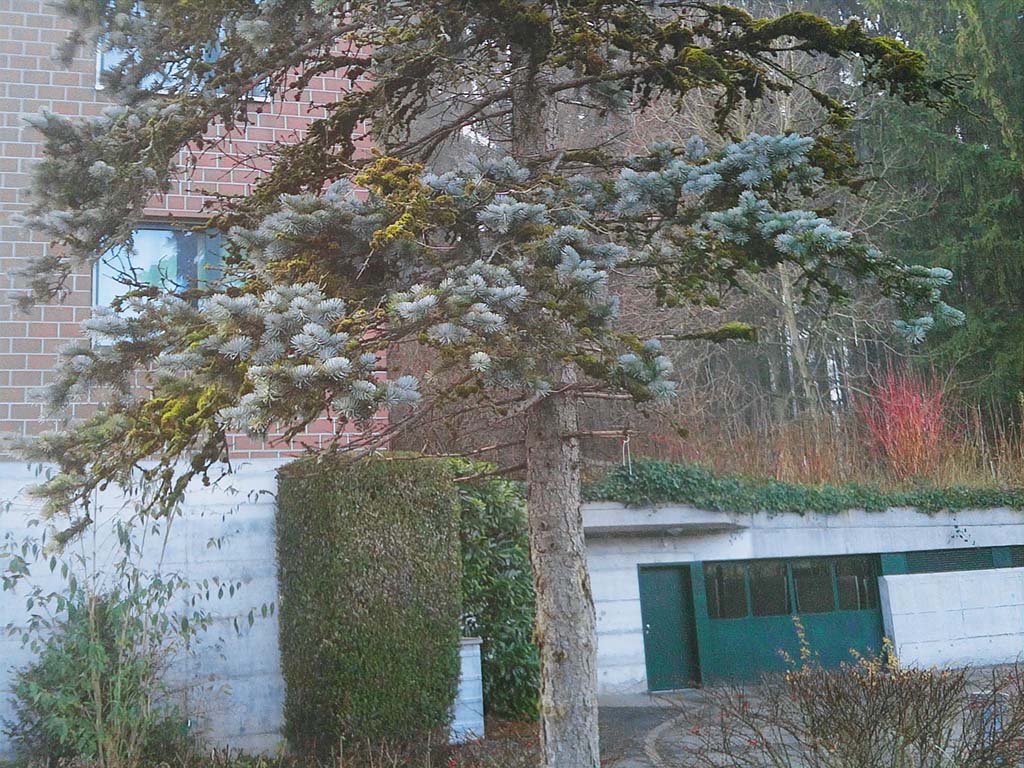}\\
   \includegraphics[width=0.36\linewidth]{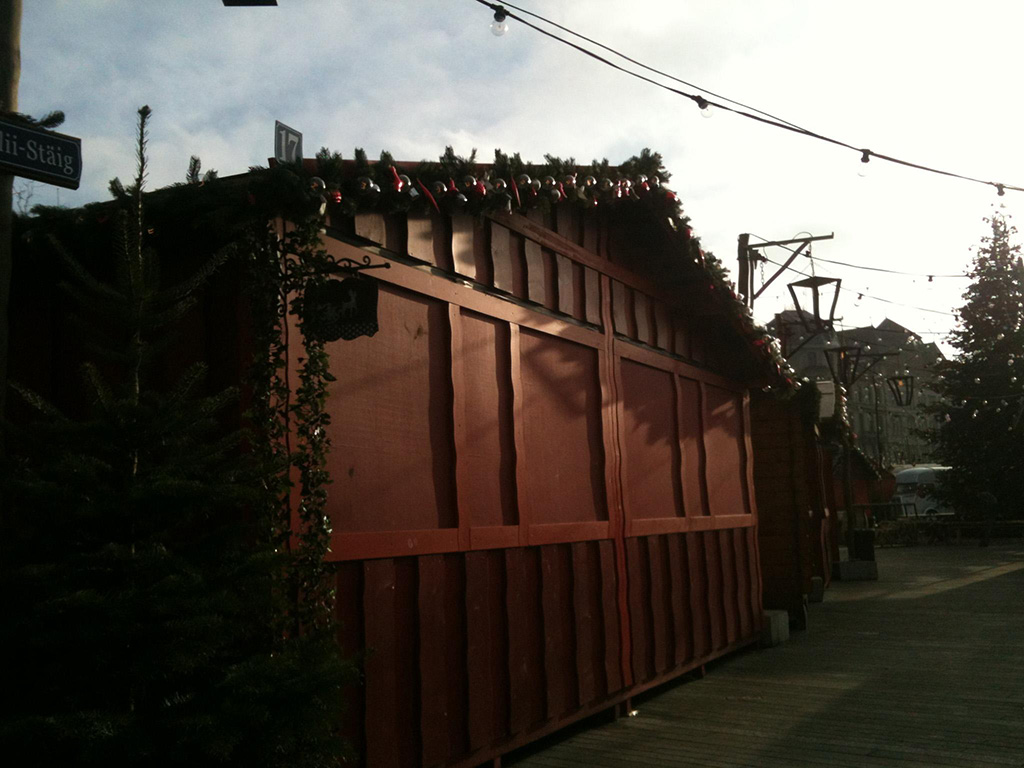}&
   \includegraphics[width=0.36\linewidth]{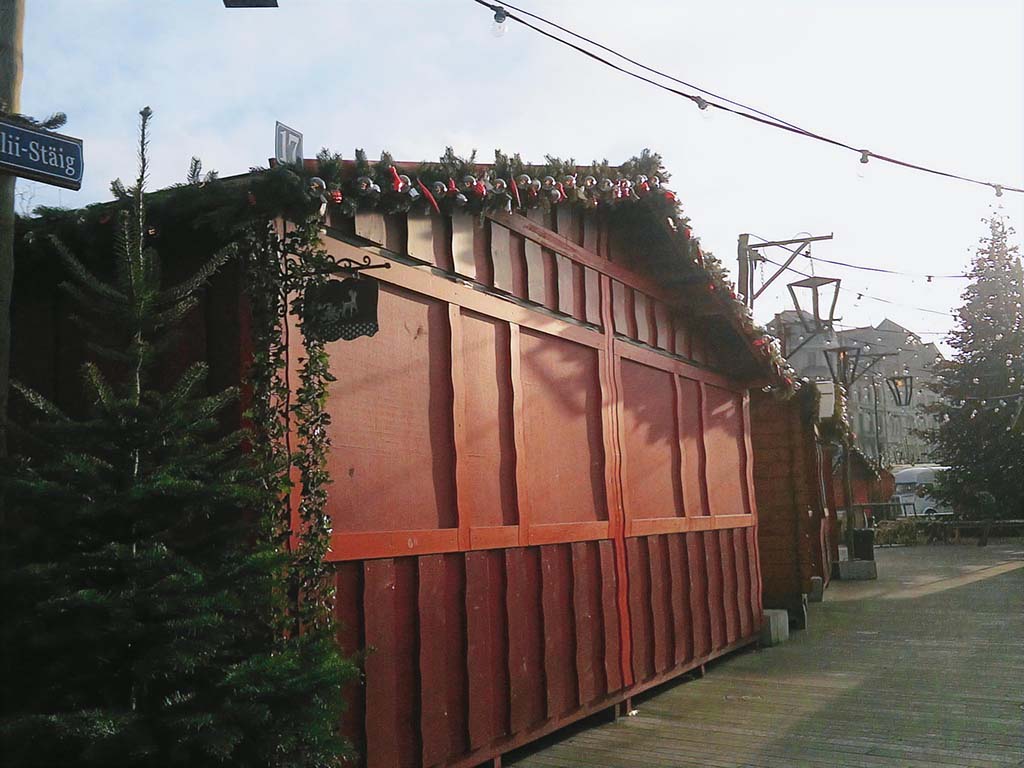}\\
   \includegraphics[width=0.36\linewidth]{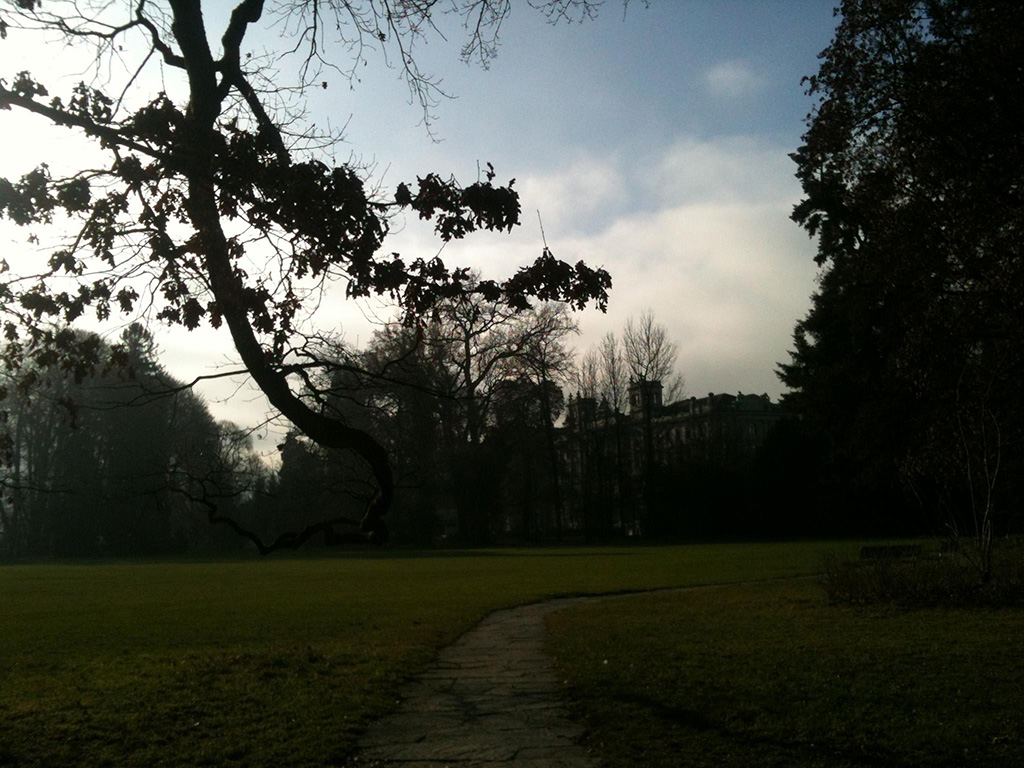}&
   \includegraphics[width=0.36\linewidth]{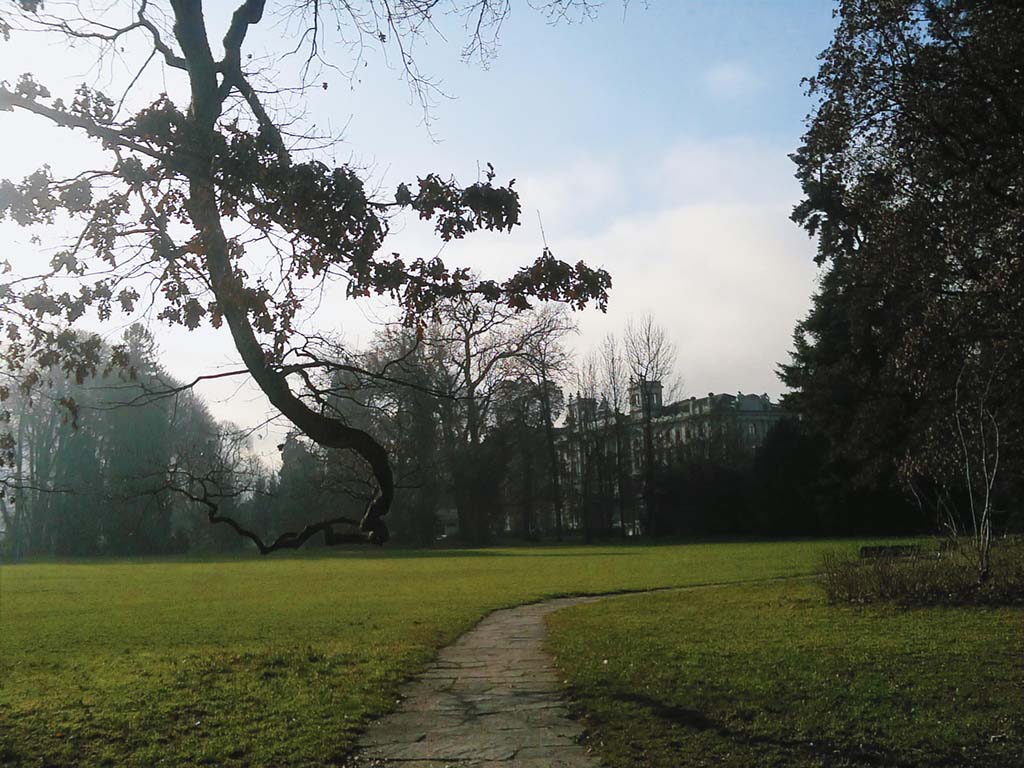}\\
   \includegraphics[width=0.36\linewidth]{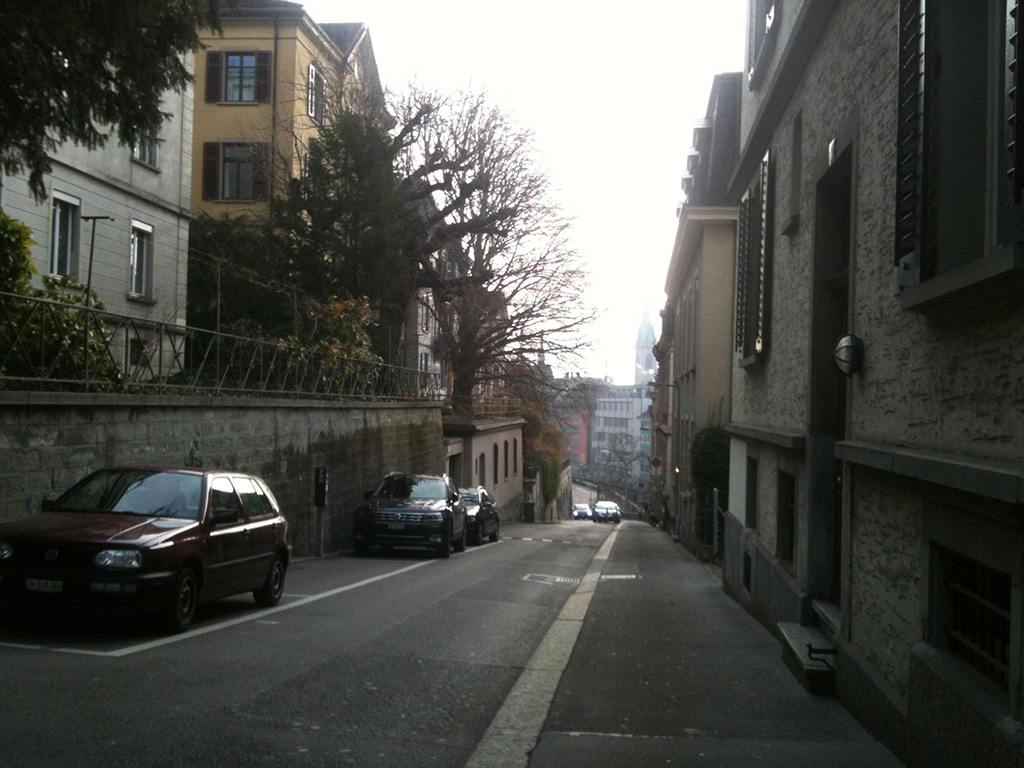}&
   \includegraphics[width=0.36\linewidth]{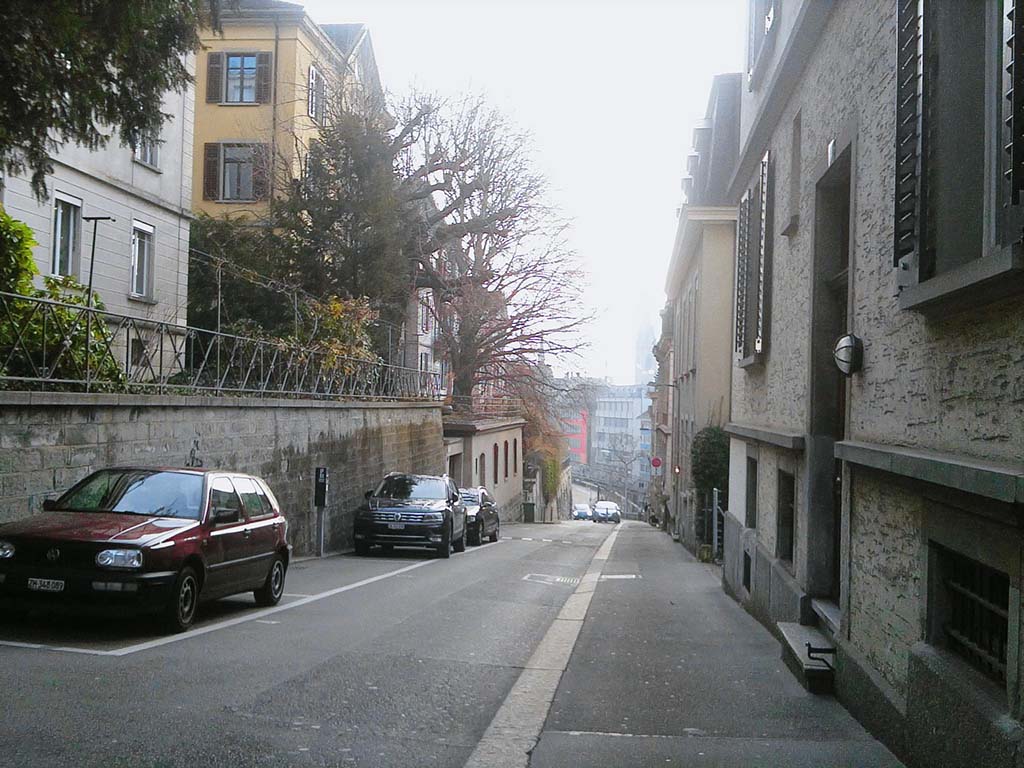}\\
\end{tabular}
\caption{Image results of our method for iPhone DPED test images.}
\label{tab:iphone_vs_ours1to4}
\end{figure*}

\begin{figure*}[h]
\setlength{\tabcolsep}{1pt}
\centering
\begin{tabular}{cc}
iPhone original & Enhanced with our method\\
   \includegraphics[width=0.39\linewidth]{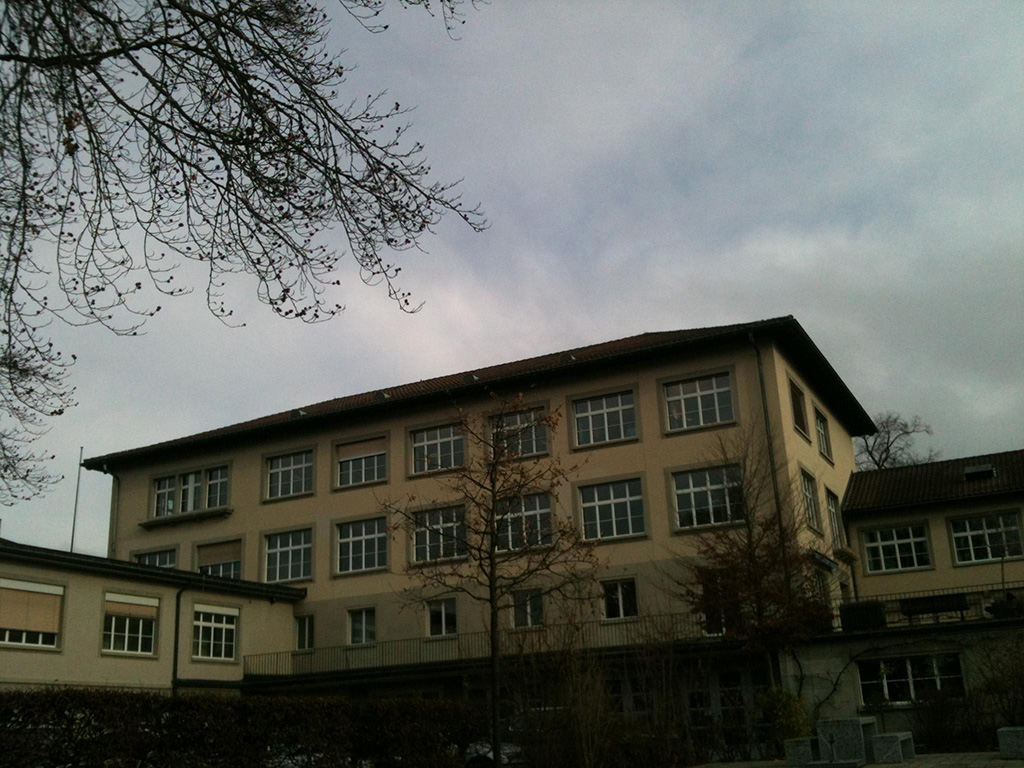}&
   \includegraphics[width=0.39\linewidth]{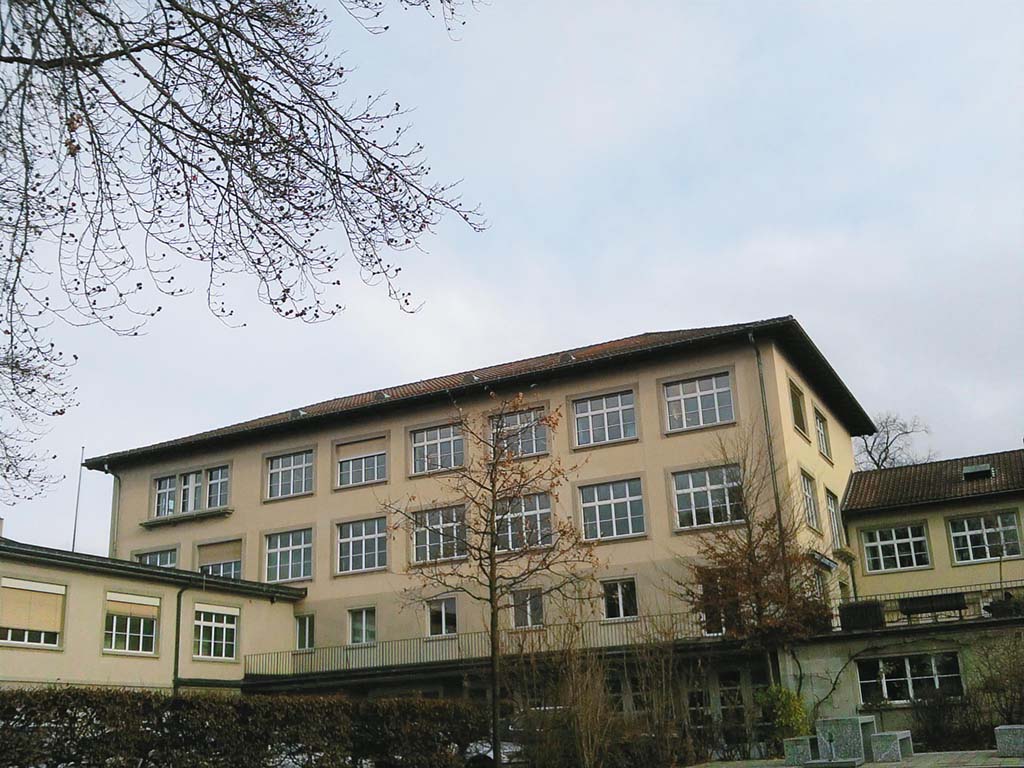}\\
   \includegraphics[width=0.39\linewidth]{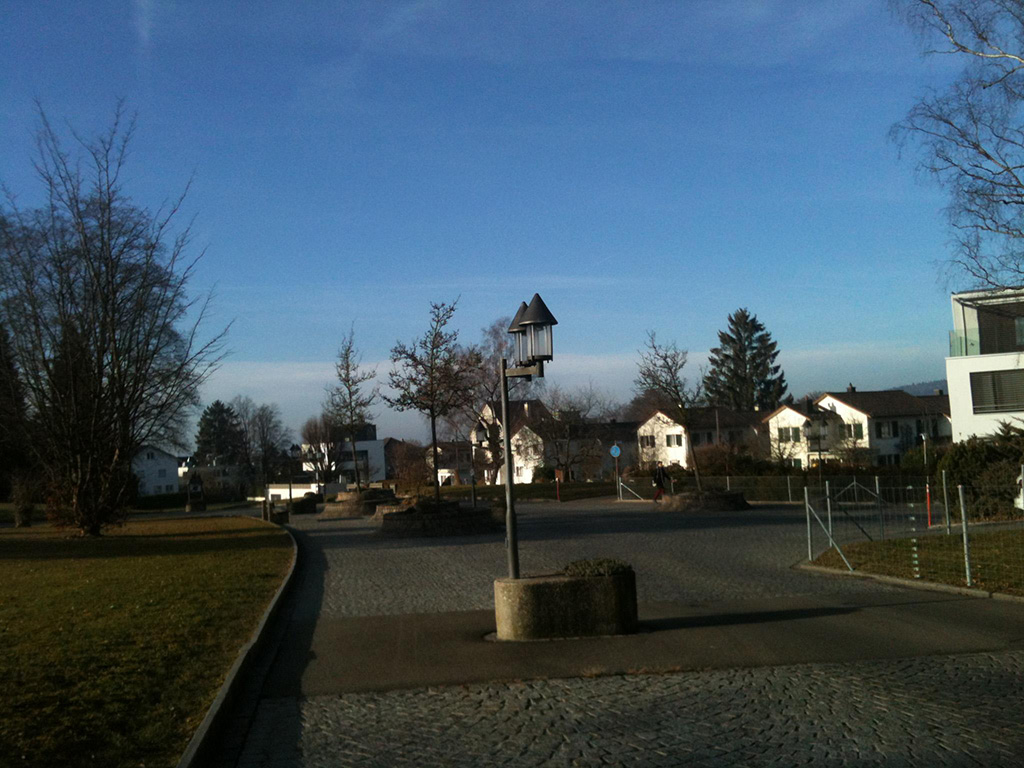}&
   \includegraphics[width=0.39\linewidth]{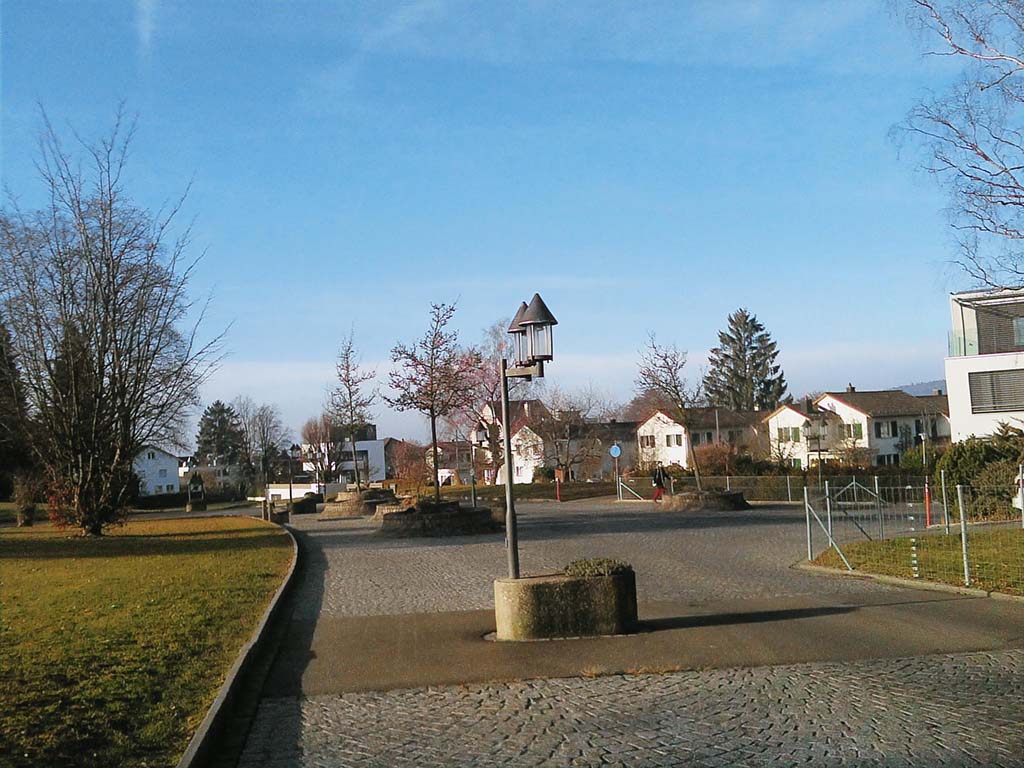}\\
   \includegraphics[width=0.39\linewidth]{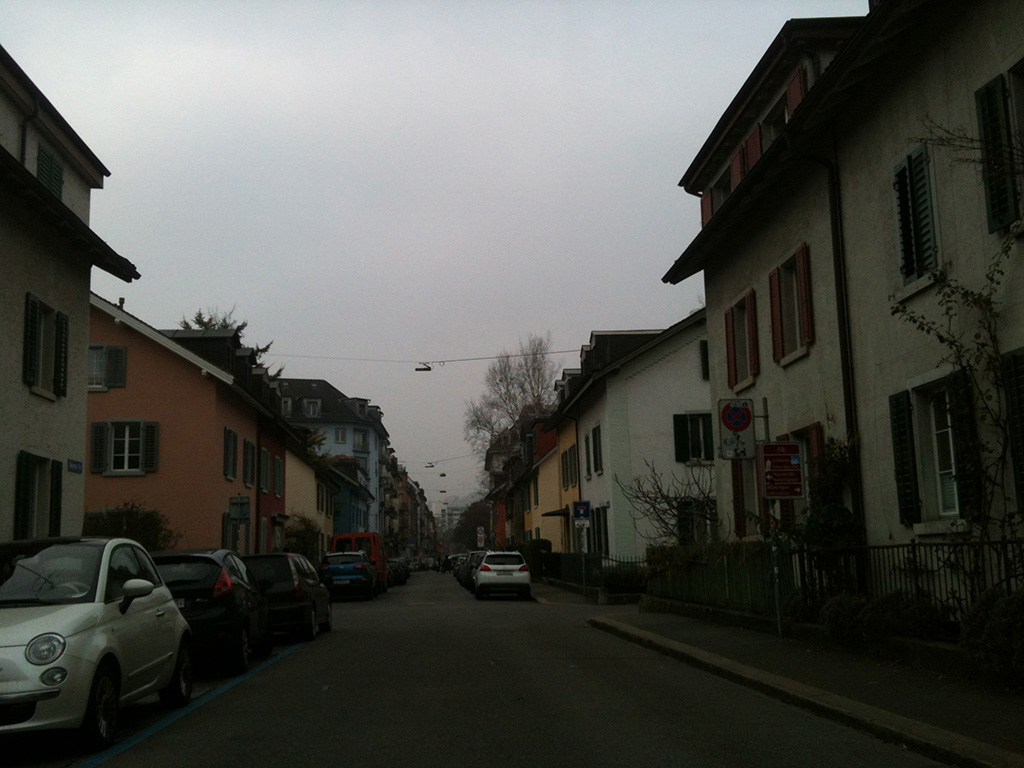}&
   \includegraphics[width=0.39\linewidth]{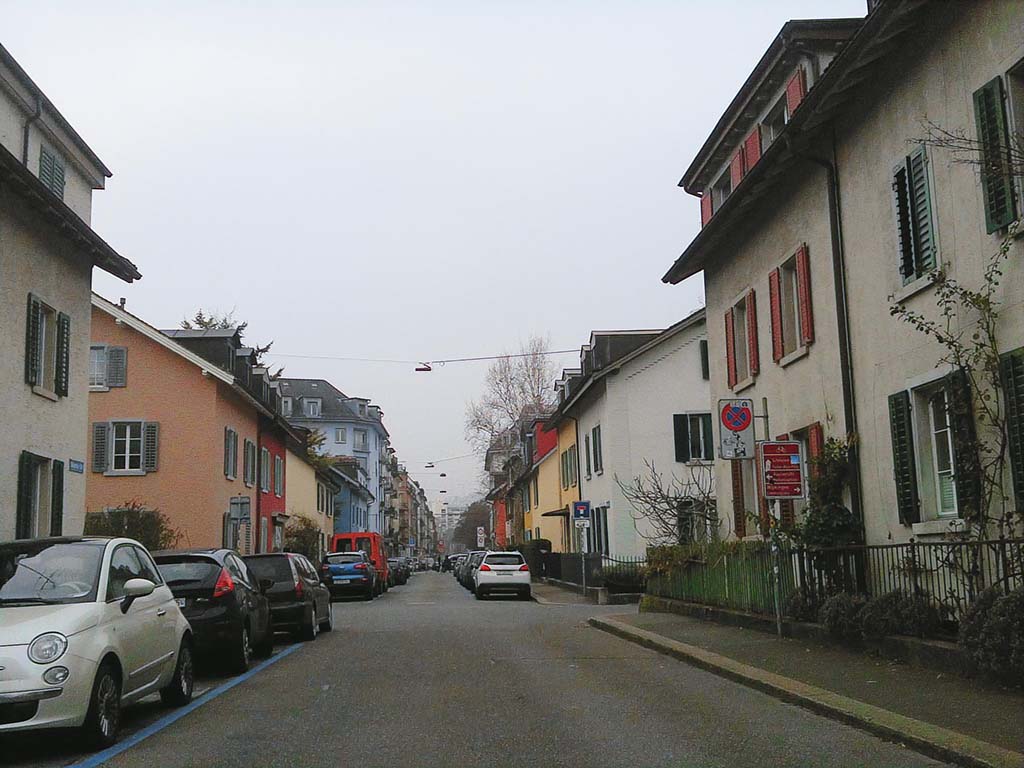}\\
   \includegraphics[width=0.39\linewidth]{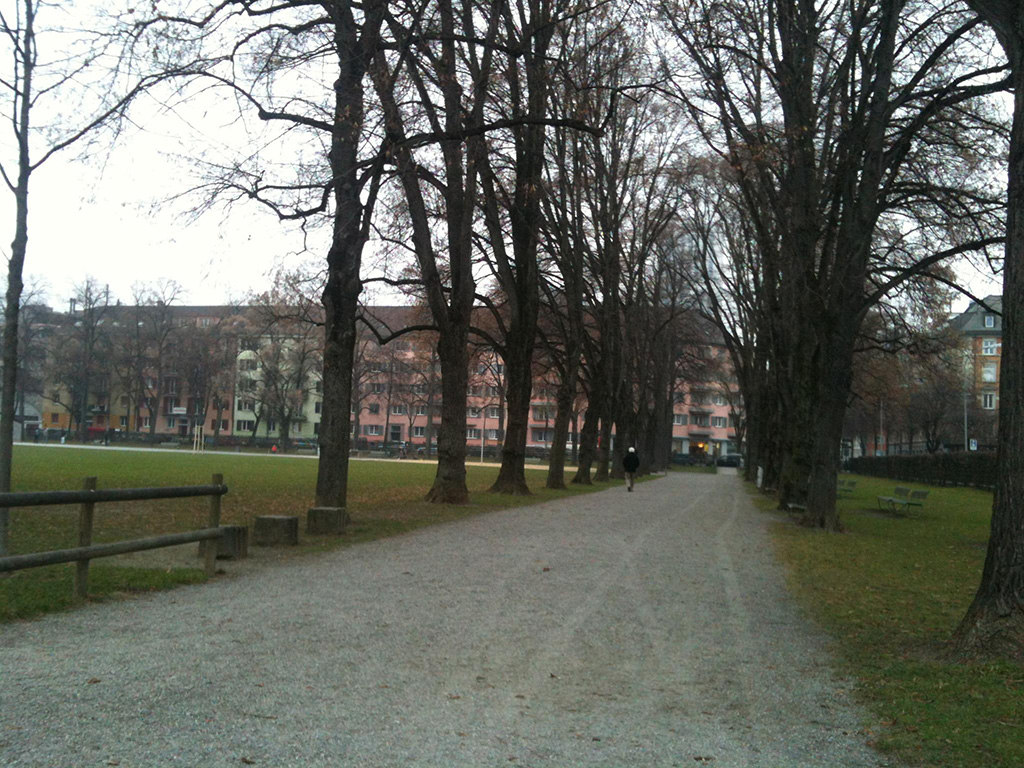}&
   \includegraphics[width=0.39\linewidth]{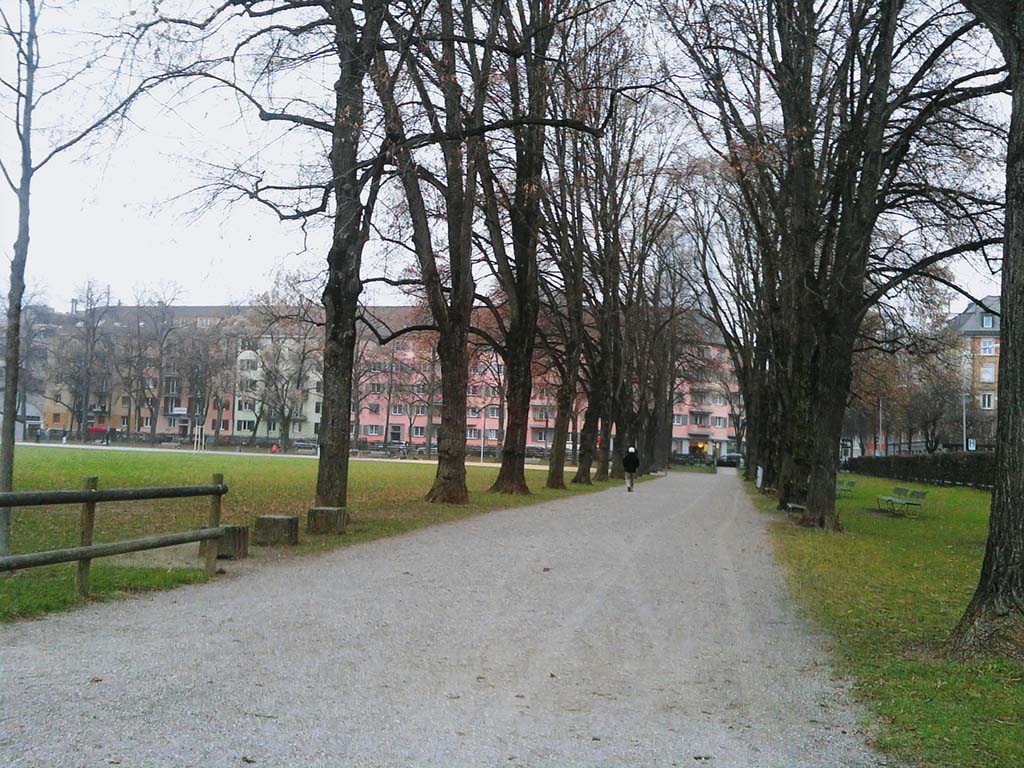}
\end{tabular}
\caption{Image results of our method for iPhone DPED test images.}
\end{figure*}

\clearpage
\newpage
\section{Appendix. Results of the proposed method: BlackBerry\protect\footnote{All visual results for BlackBerry are available at \url{http://people.ee.ethz.ch/~ihnatova/dped_blackberry.html}}}
\label{sec:results_BlackBerry}

\begin{figure*}[h!]
\setlength{\tabcolsep}{1pt}
\centering
\begin{tabular}{cc}
BlackBerry original & Enhanced with our method\\
   \includegraphics[width=0.36\linewidth]{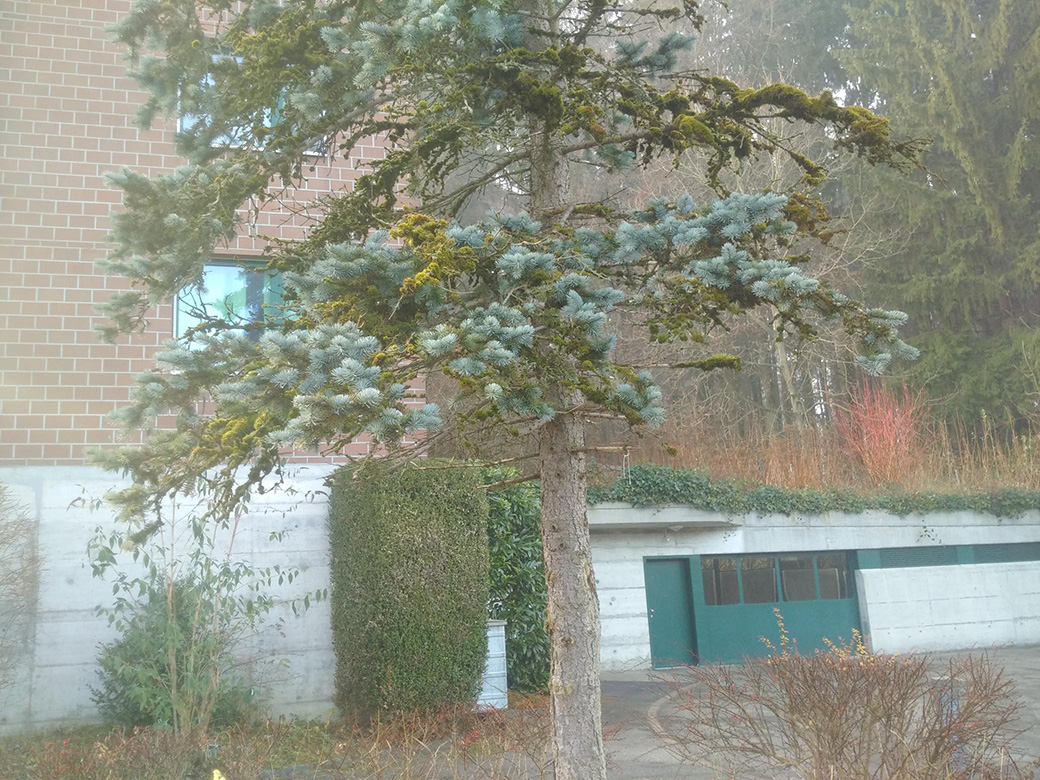}&
   \includegraphics[width=0.36\linewidth]{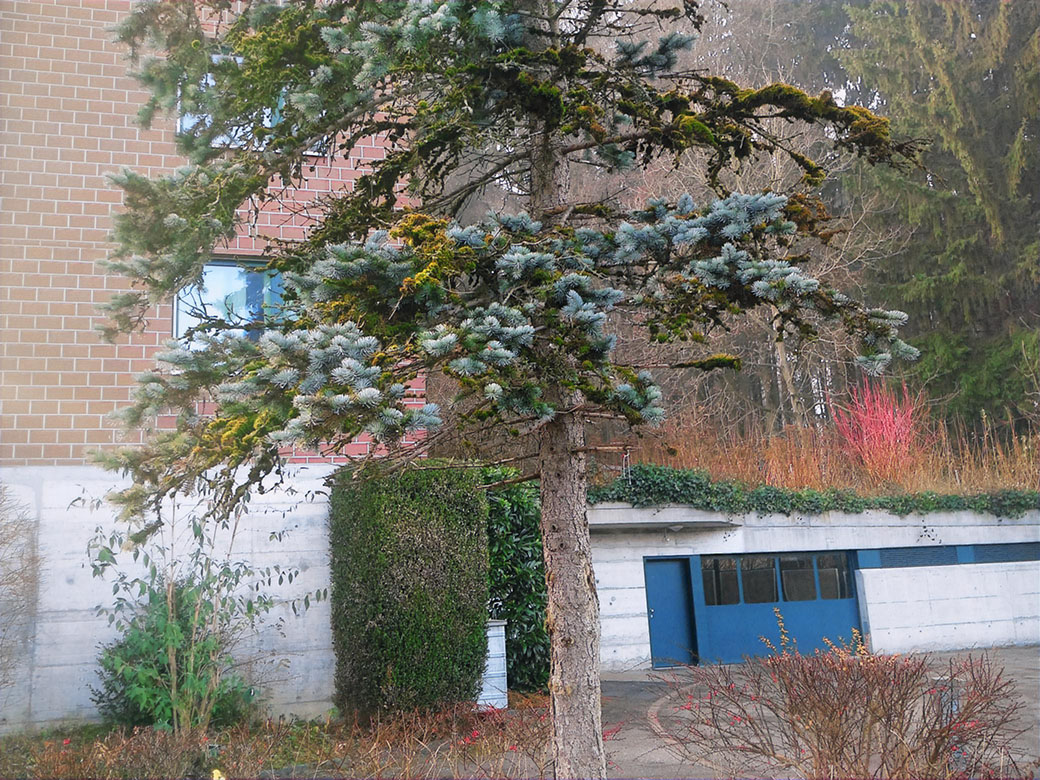}\\
   \includegraphics[width=0.36\linewidth]{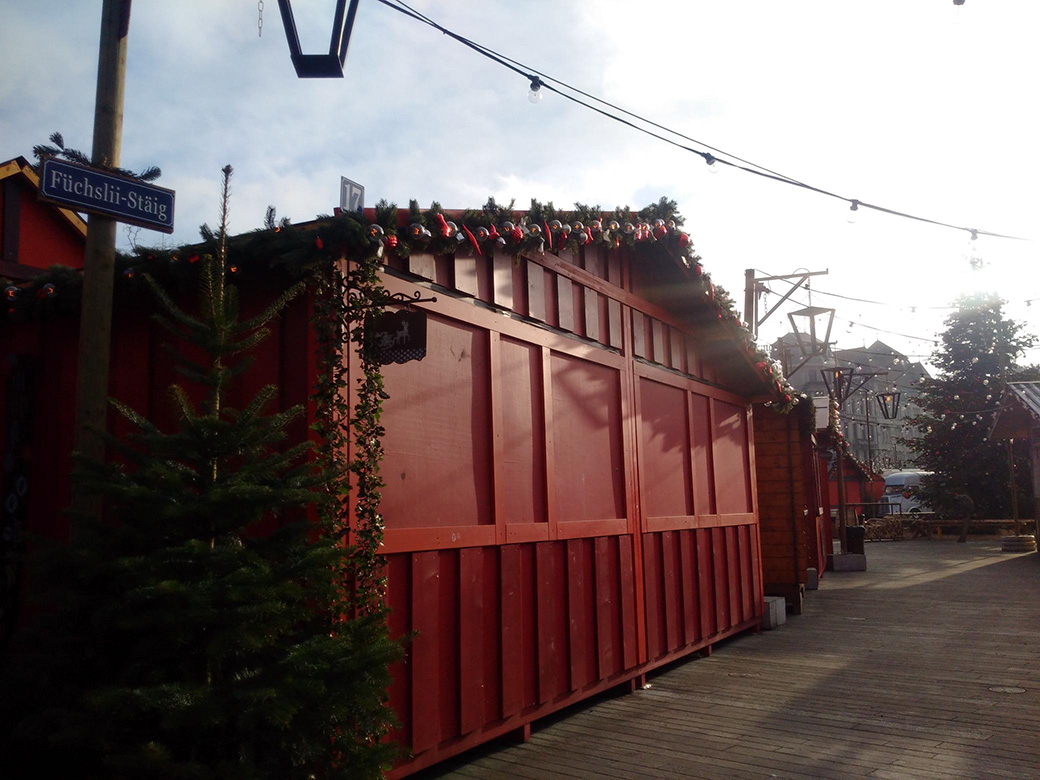}&
   \includegraphics[width=0.36\linewidth]{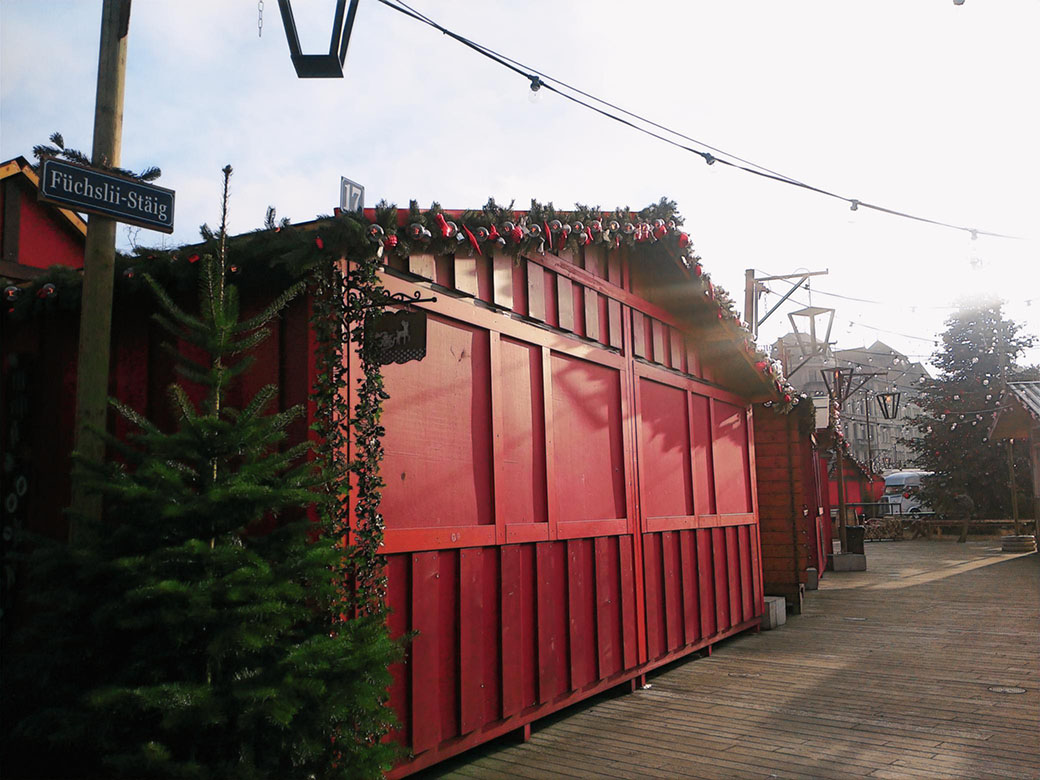}\\
   \includegraphics[width=0.36\linewidth]{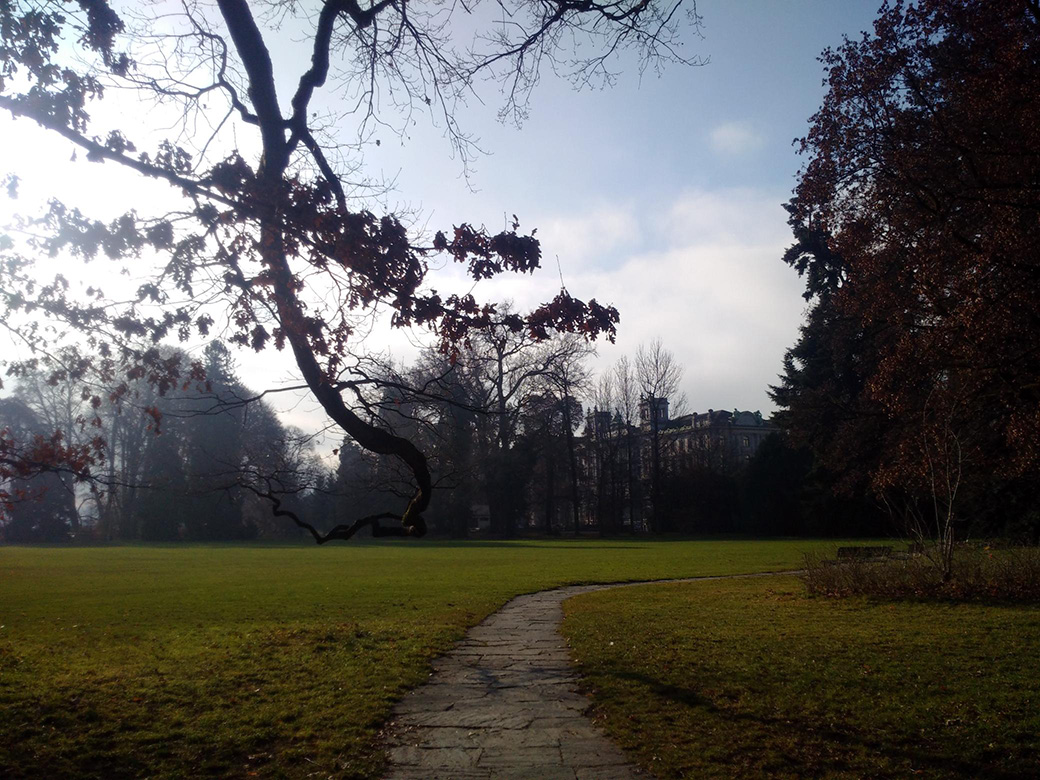}&
   \includegraphics[width=0.36\linewidth]{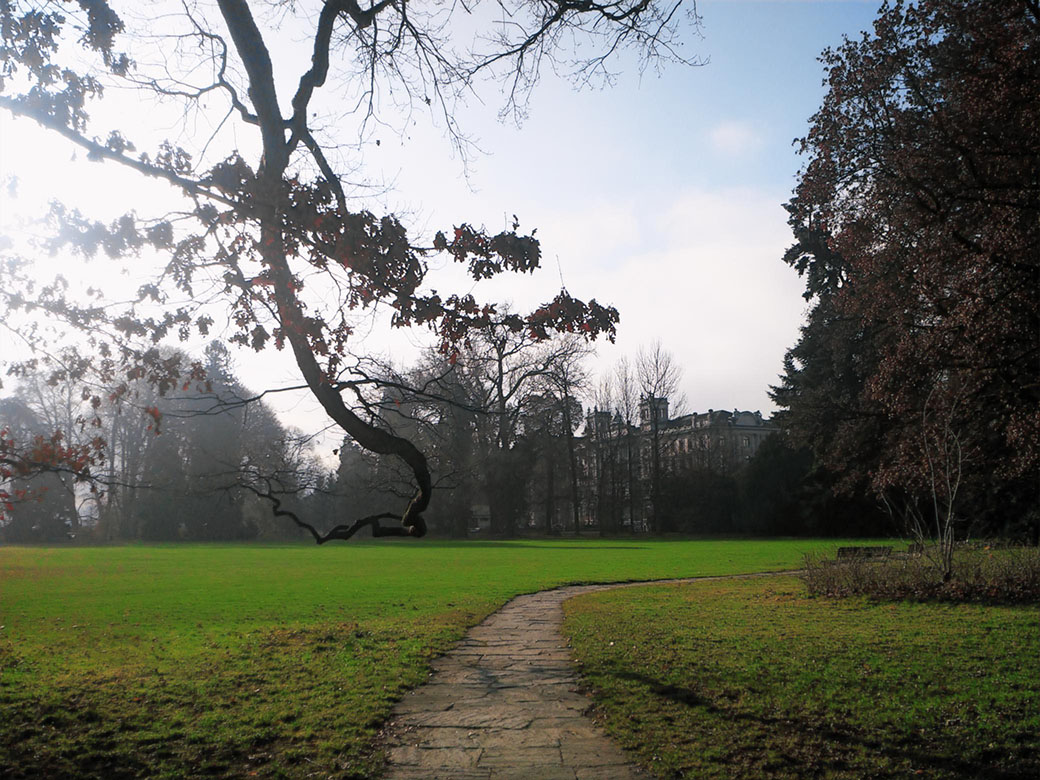}\\
   \includegraphics[width=0.36\linewidth]{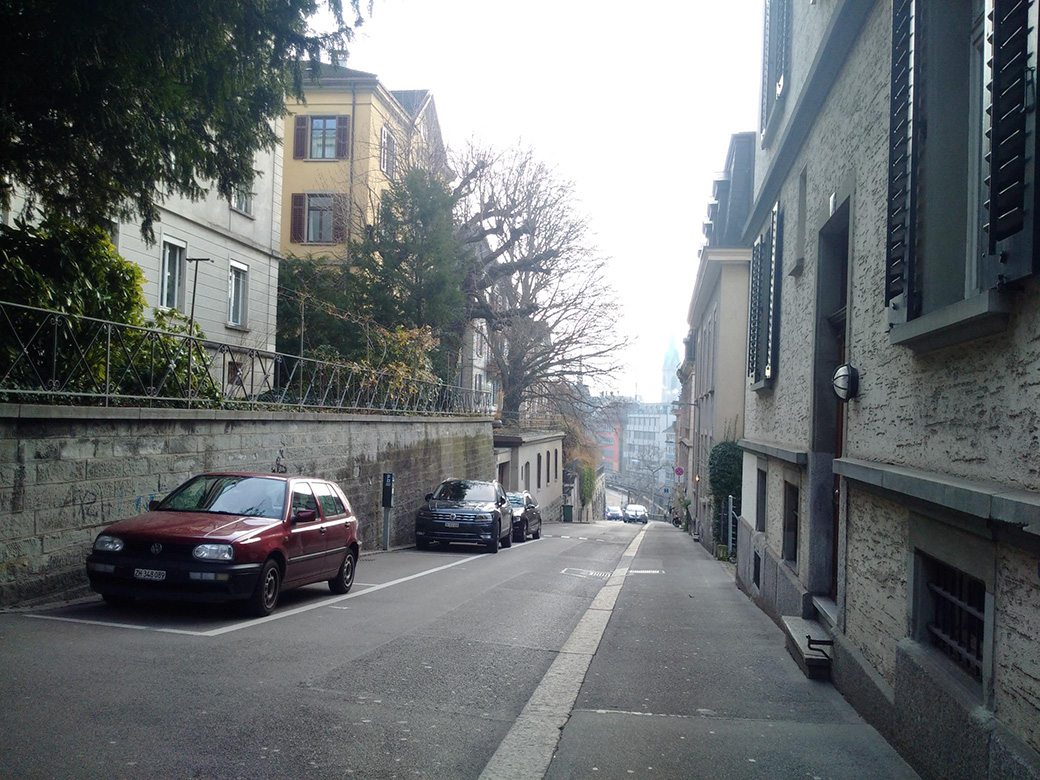}&
   \includegraphics[width=0.36\linewidth]{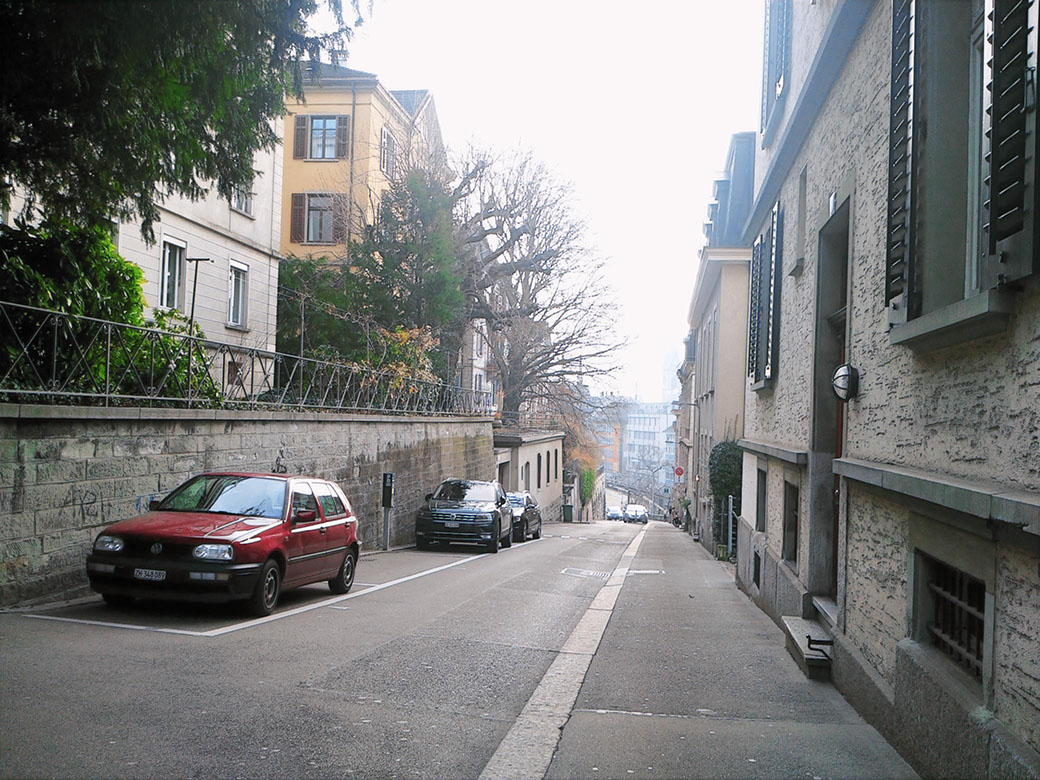}\\
\end{tabular}
\caption{Image results of our method for BlackBerry DPED test images.}
\end{figure*}

\begin{figure*}[h]
\setlength{\tabcolsep}{1pt}
\centering
\begin{tabular}{cc}
BlackBerry original & Enhanced with our method\\
   \includegraphics[width=0.39\linewidth]{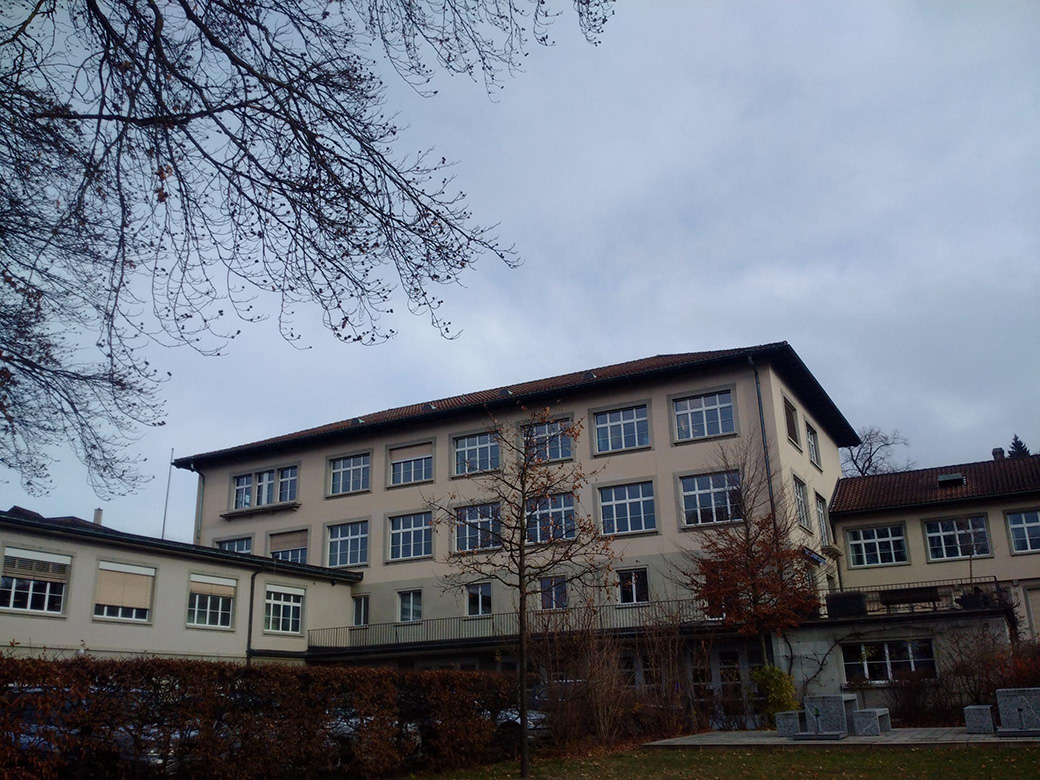}&
   \includegraphics[width=0.39\linewidth]{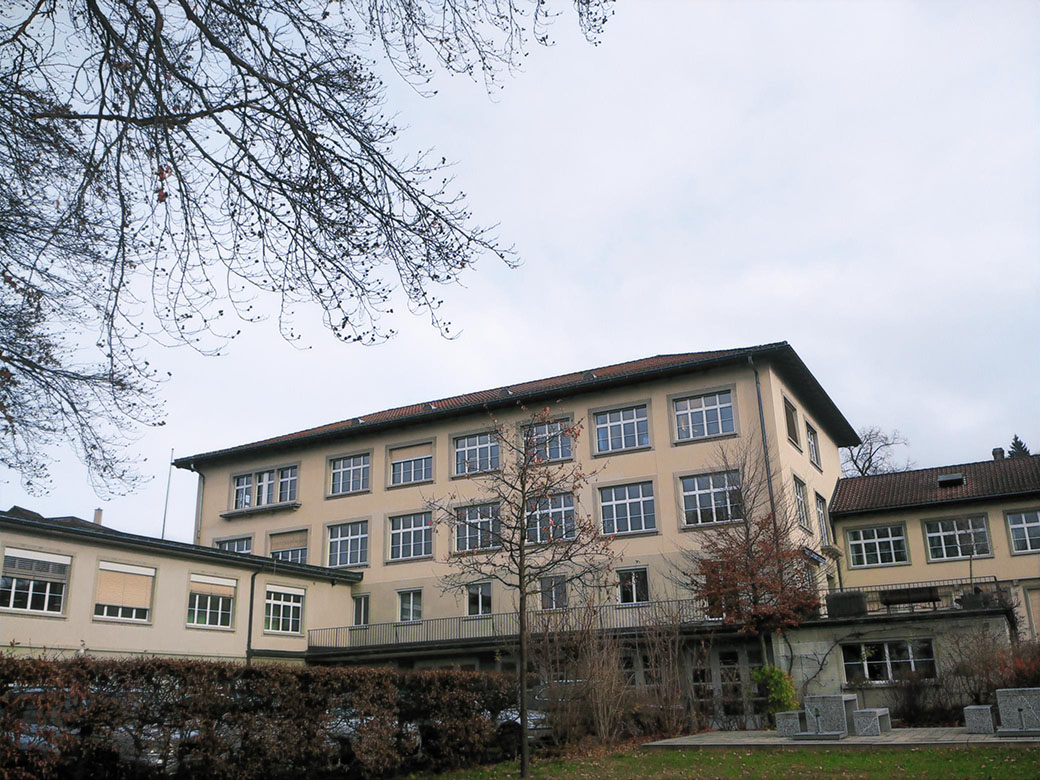}\\
   \includegraphics[width=0.39\linewidth]{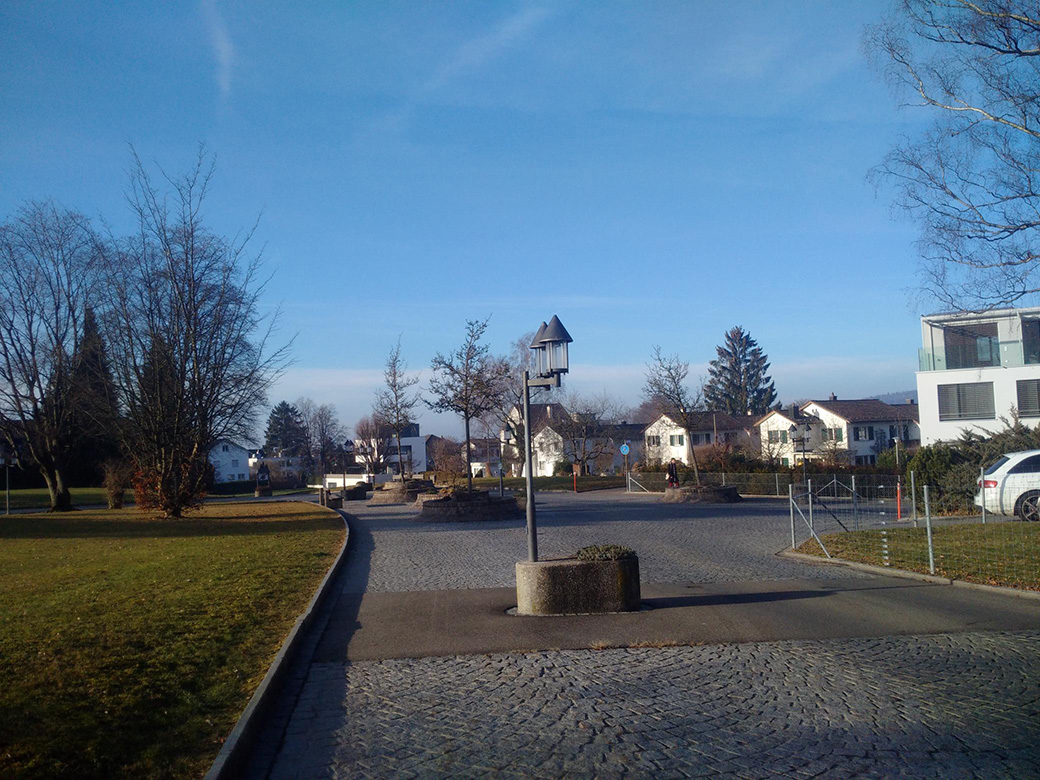}&
   \includegraphics[width=0.39\linewidth]{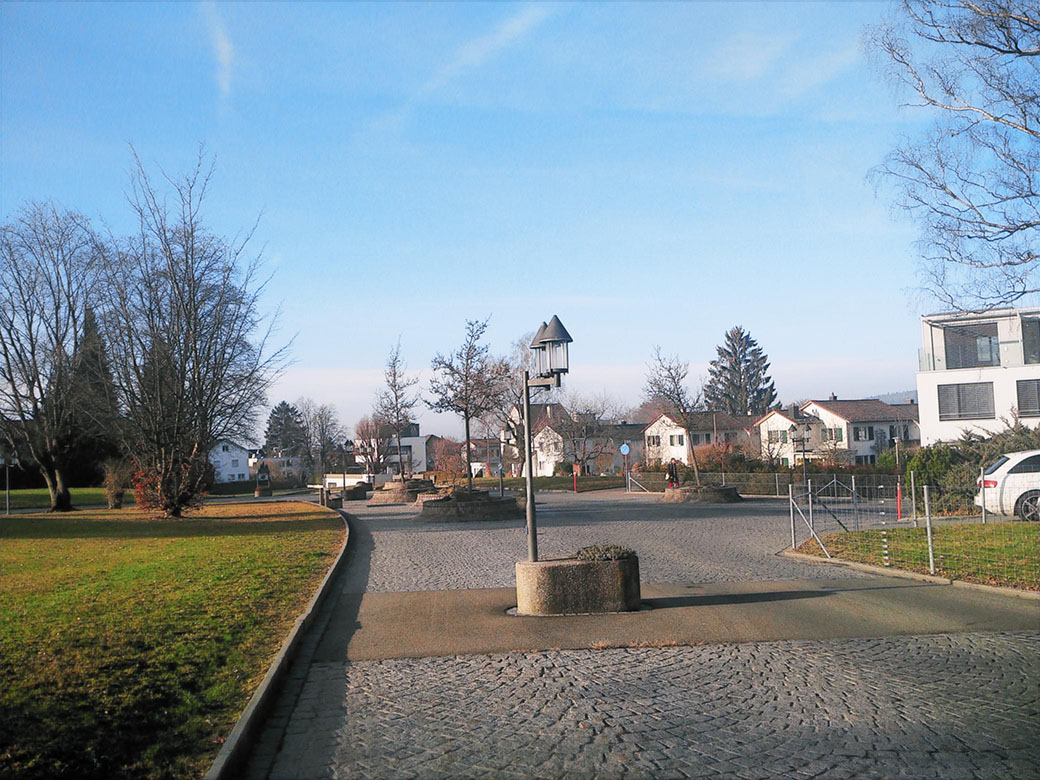}\\
   \includegraphics[width=0.39\linewidth]{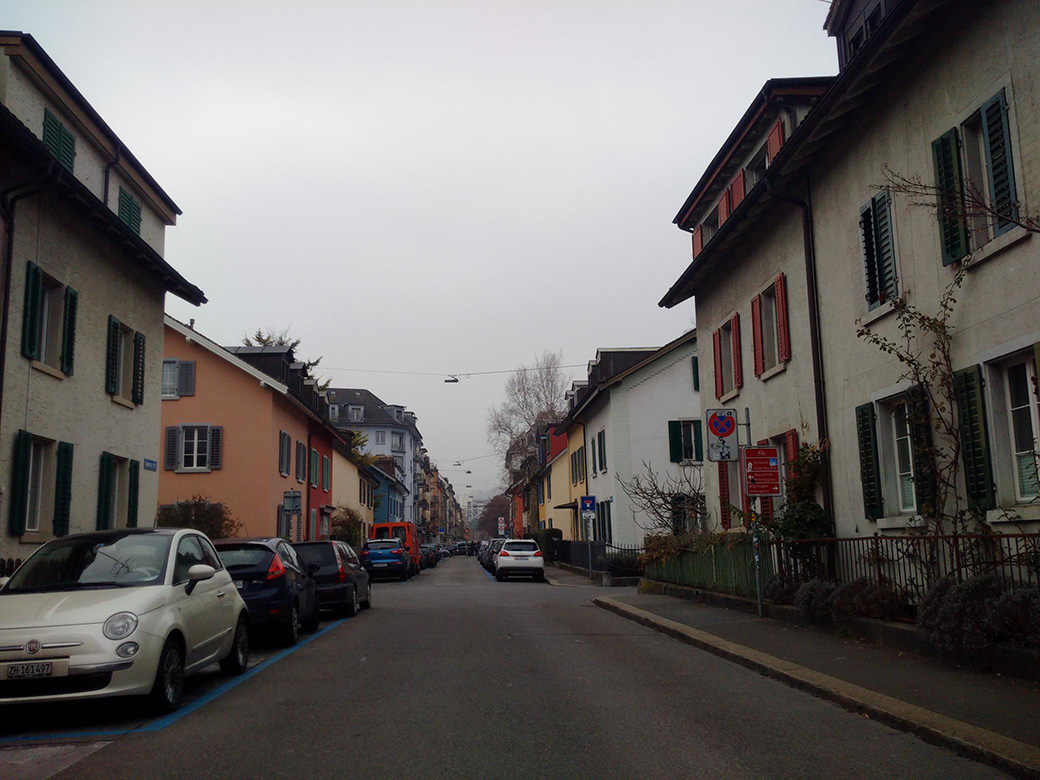}&
   \includegraphics[width=0.39\linewidth]{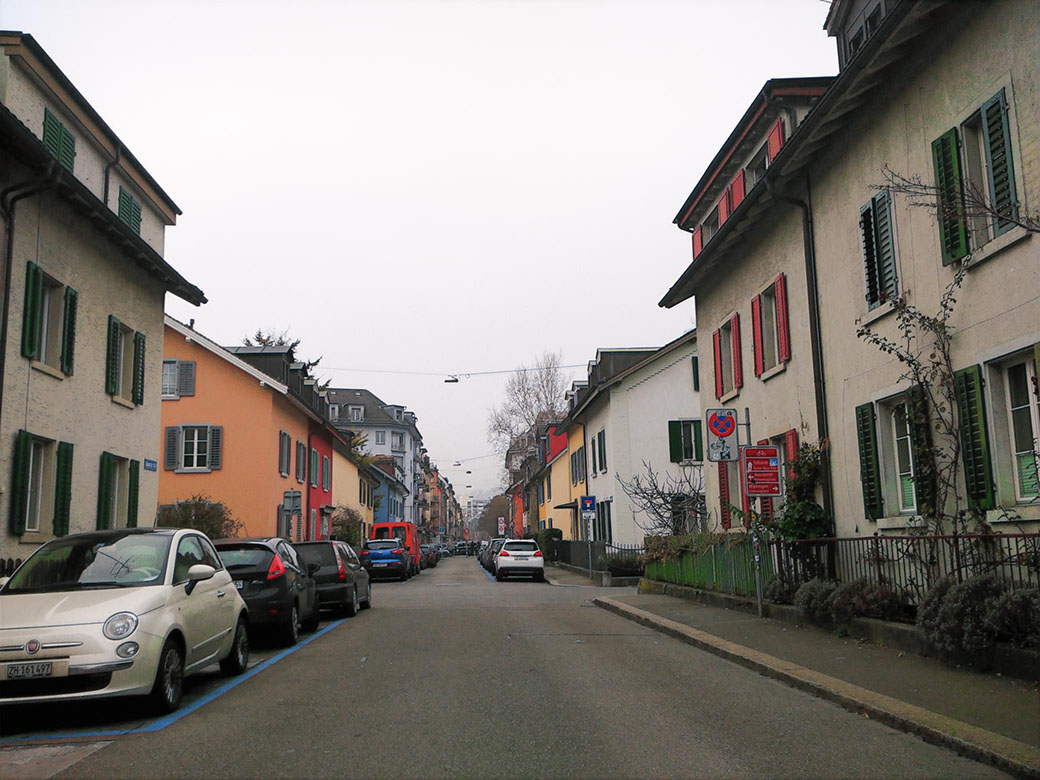}\\
   \includegraphics[width=0.39\linewidth]{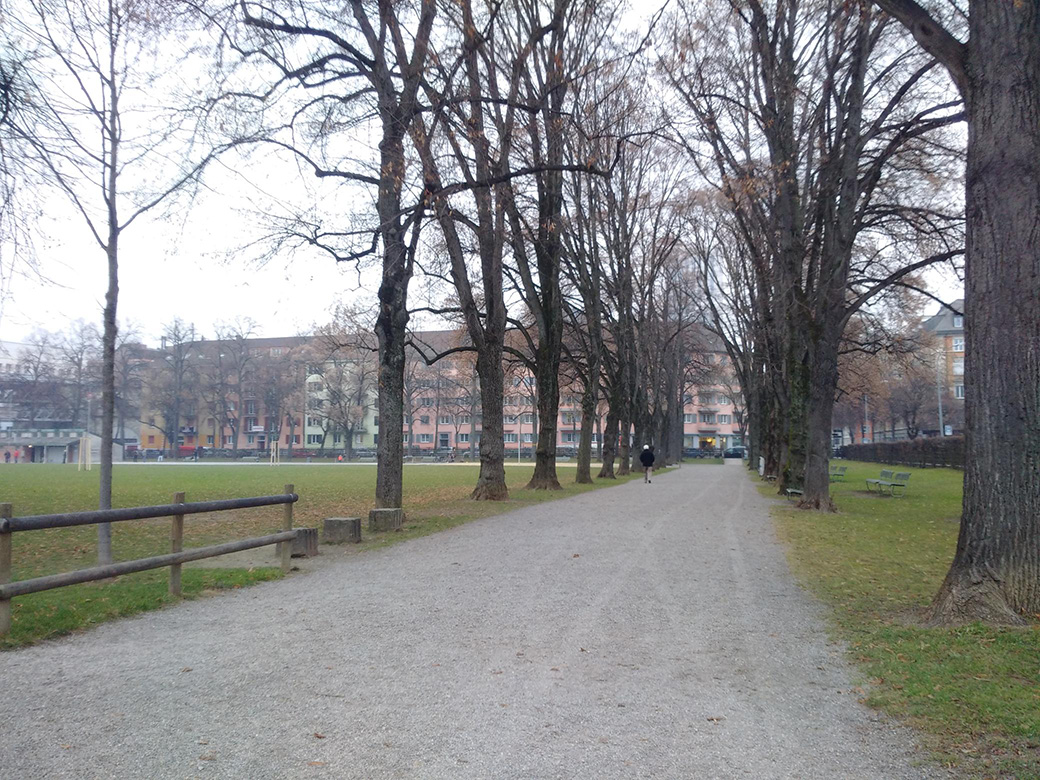}&
   \includegraphics[width=0.39\linewidth]{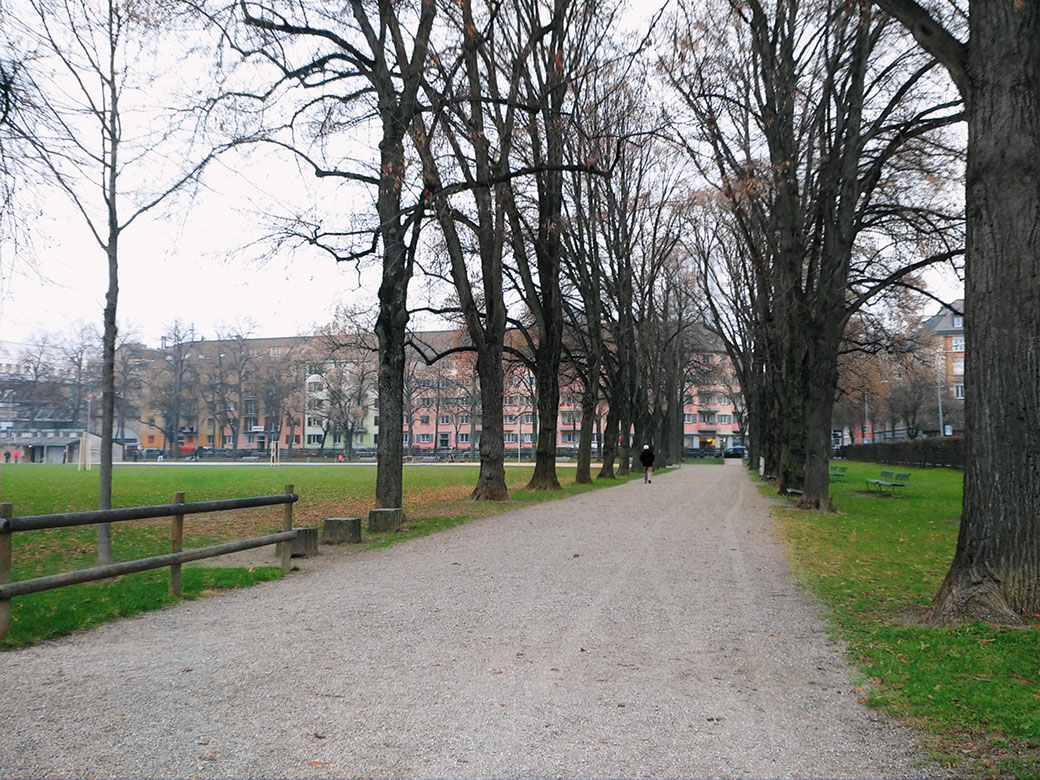}\\
\end{tabular}
\caption{Image results of our method for BlackBerry DPED test images.}
\end{figure*}

\clearpage
\newpage

\section{Appendix. Results of the proposed method: Sony\protect\footnote{All visual results for Sony are available at \url{http://people.ee.ethz.ch/~ihnatova/dped_sony.html}}}
\label{sec:results_Sony}

\begin{figure*}[h!]
\setlength{\tabcolsep}{1pt}
\centering
\begin{tabular}{cc}
Sony original & Enhanced with our method\\
   \includegraphics[width=0.36\linewidth]{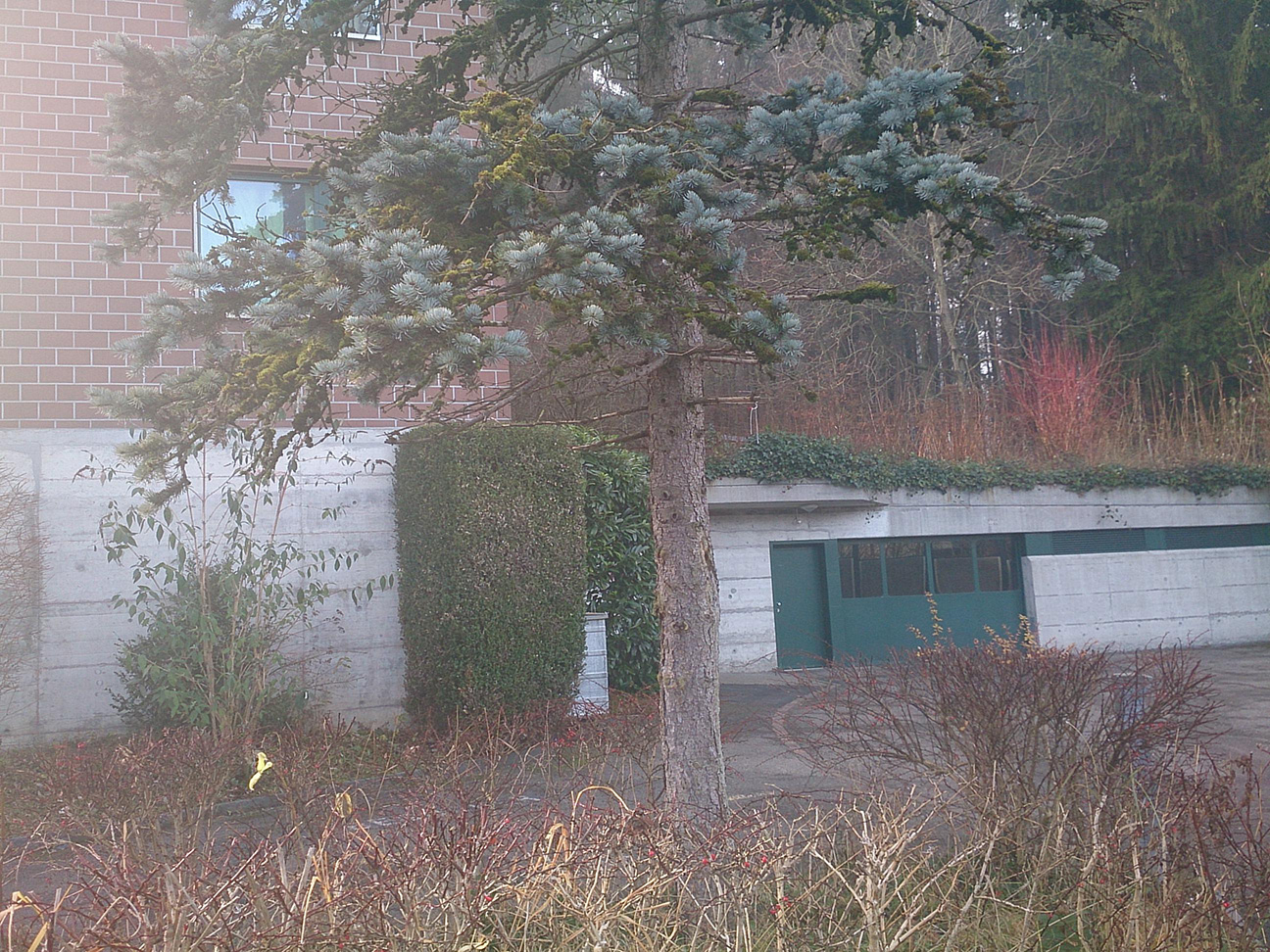}&
   \includegraphics[width=0.36\linewidth]{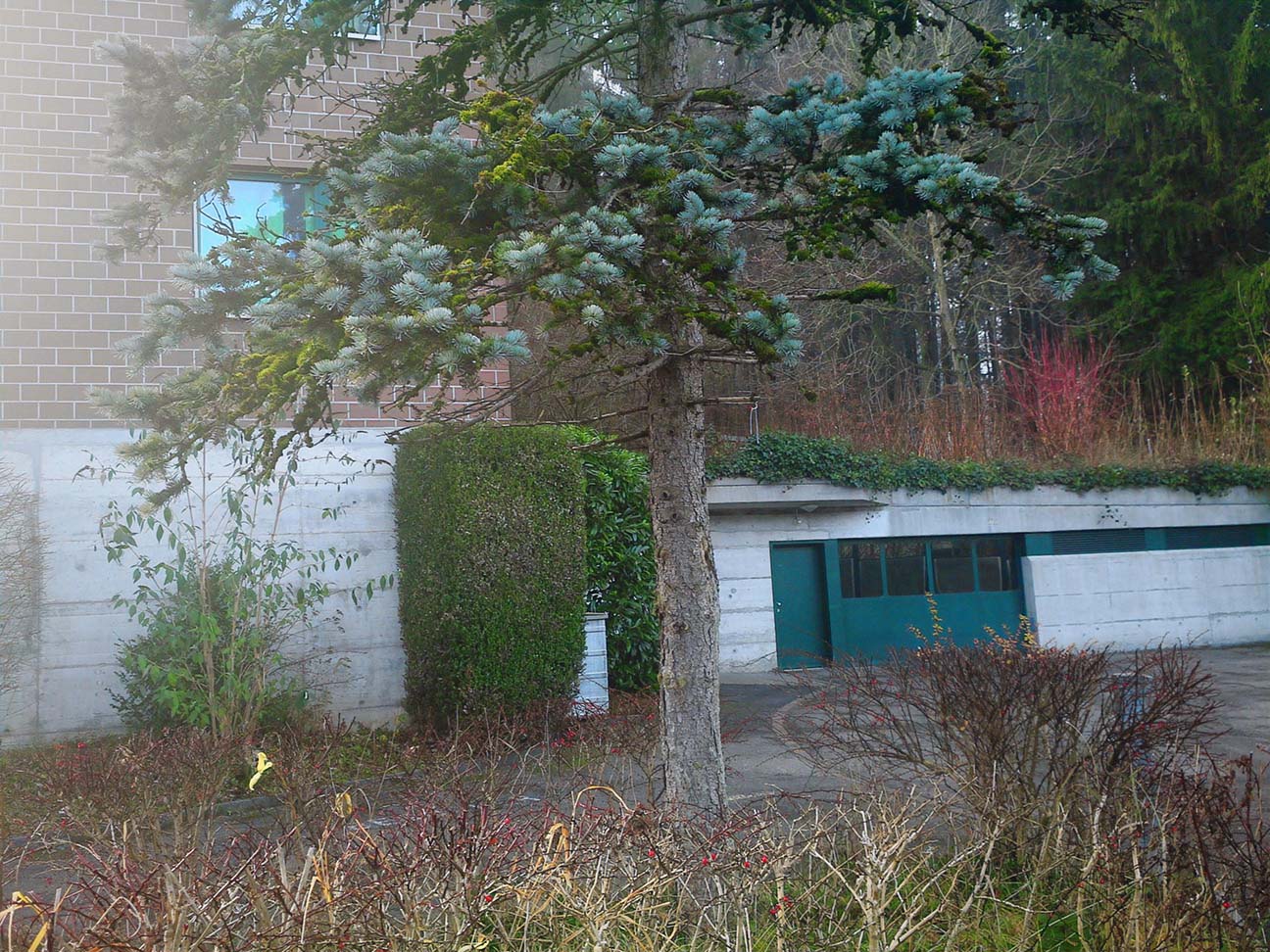}\\
   \includegraphics[width=0.36\linewidth]{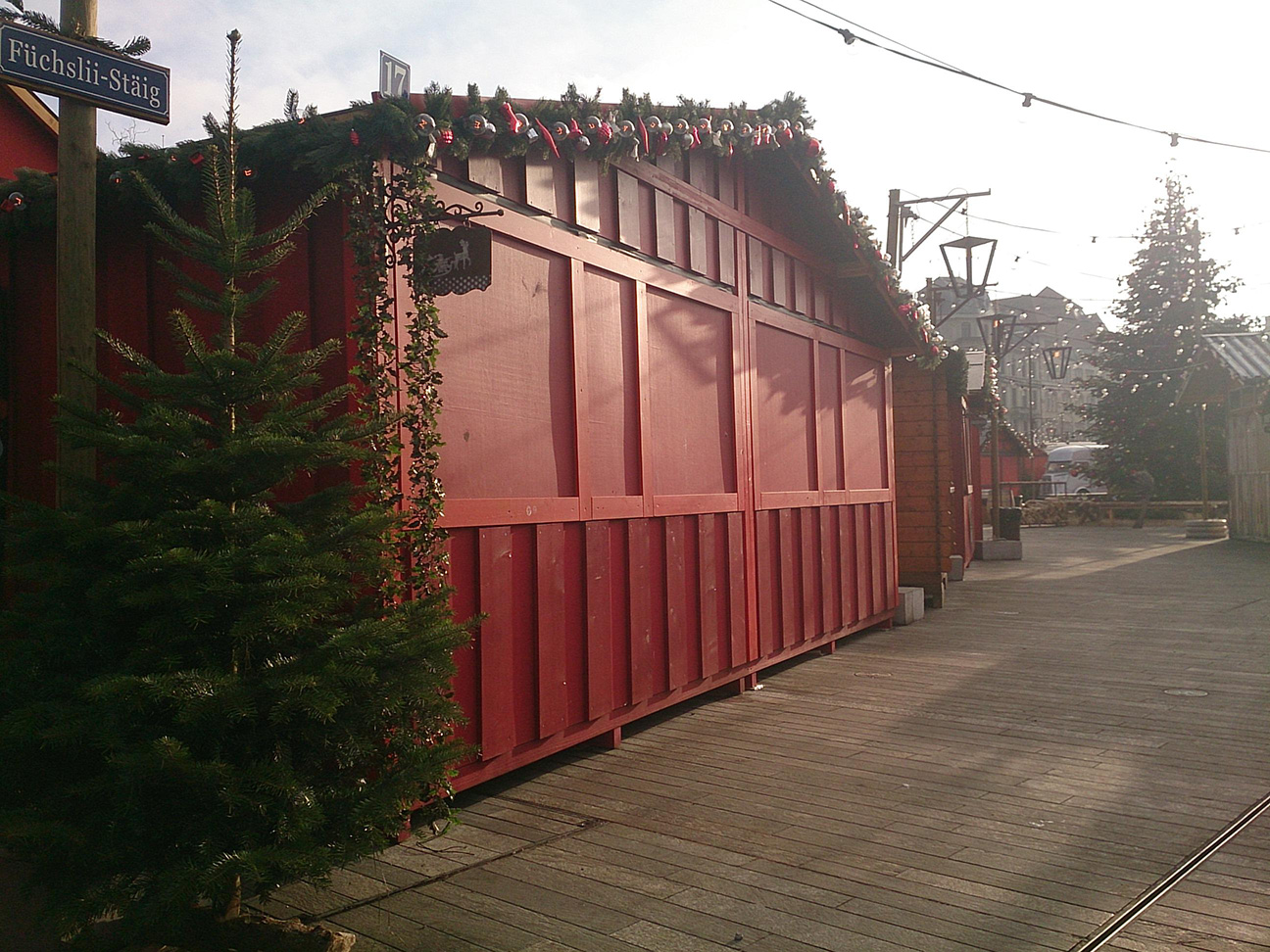}&
   \includegraphics[width=0.36\linewidth]{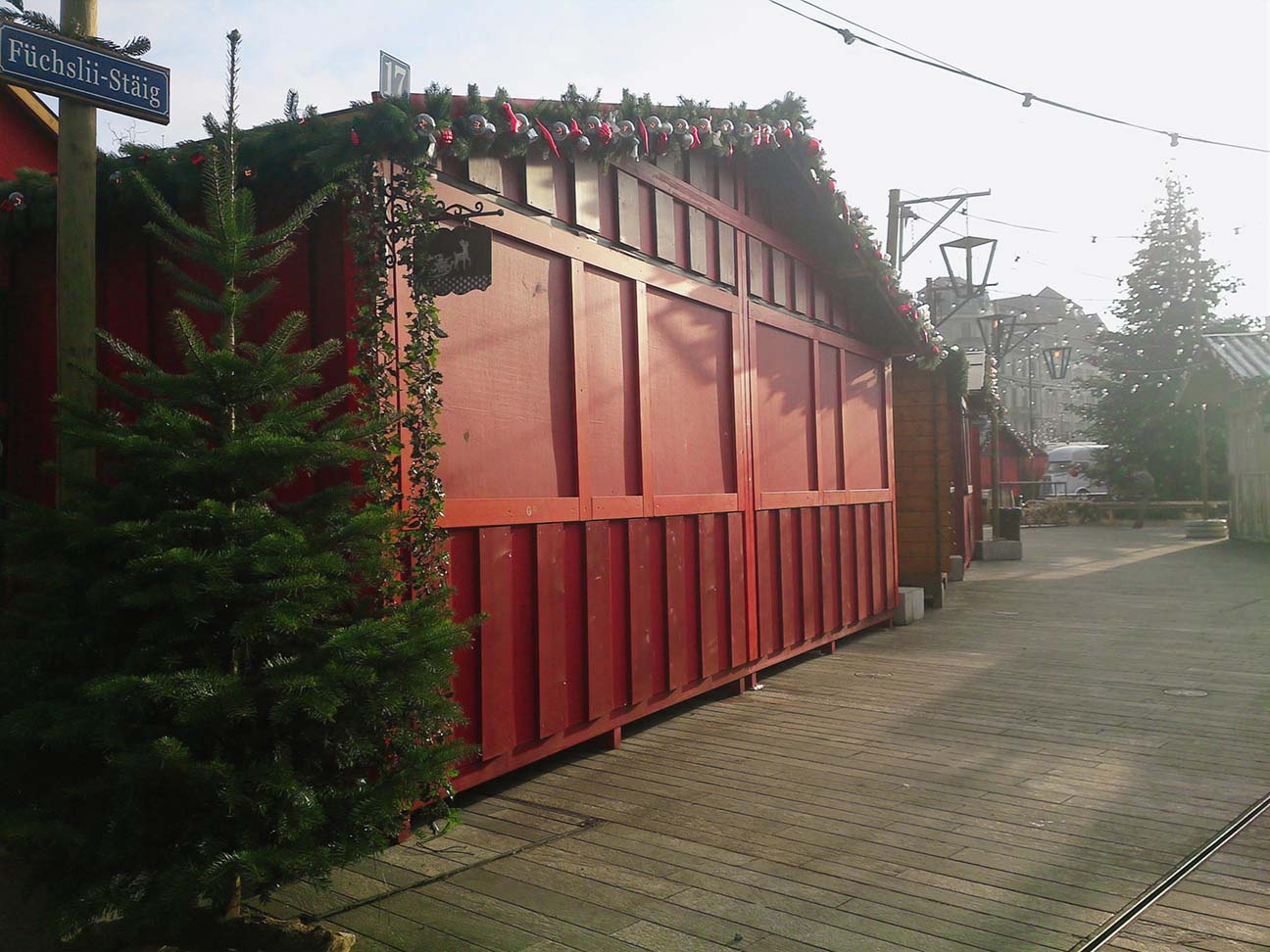}\\
   \includegraphics[width=0.36\linewidth]{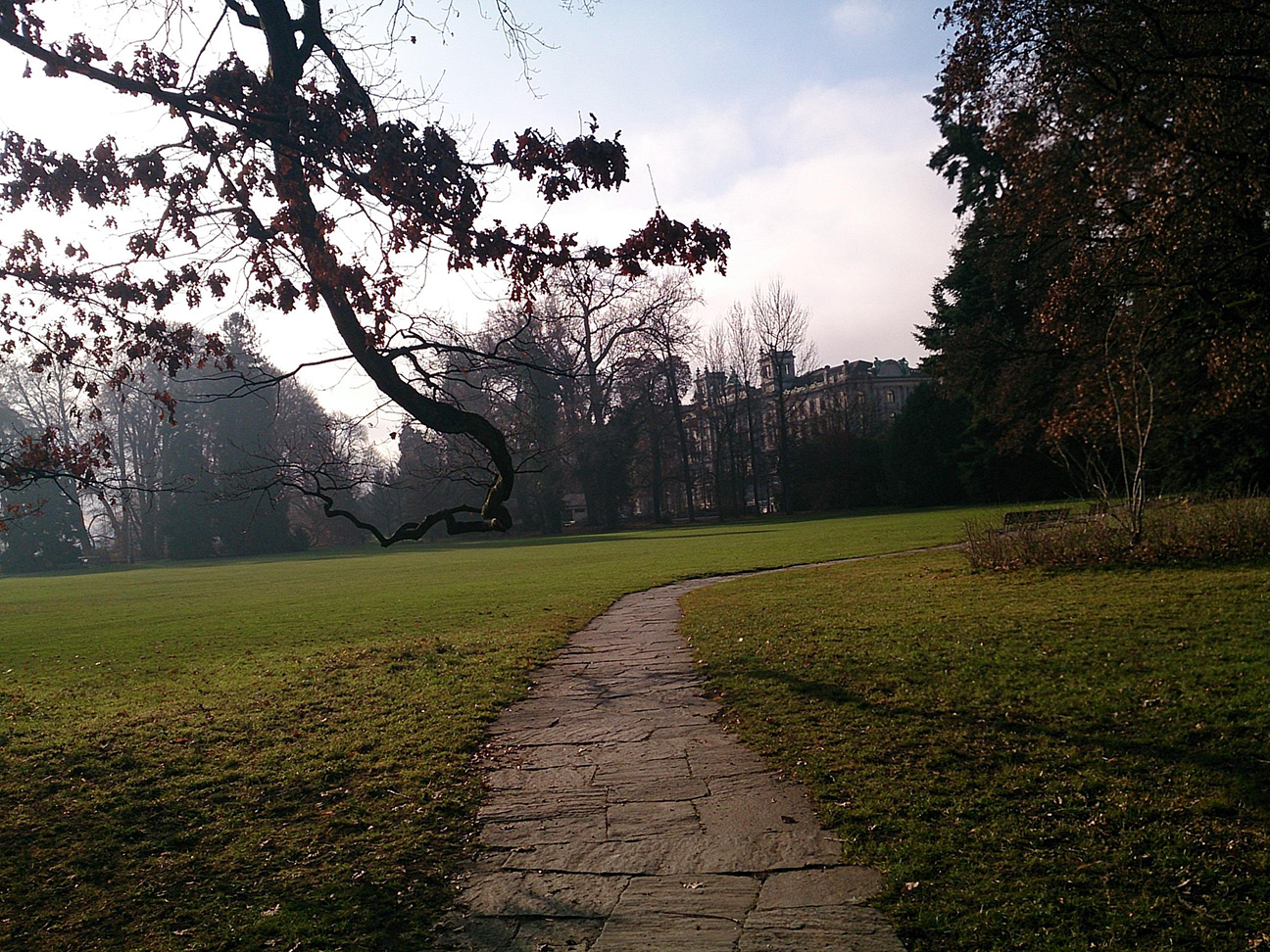}&
   \includegraphics[width=0.36\linewidth]{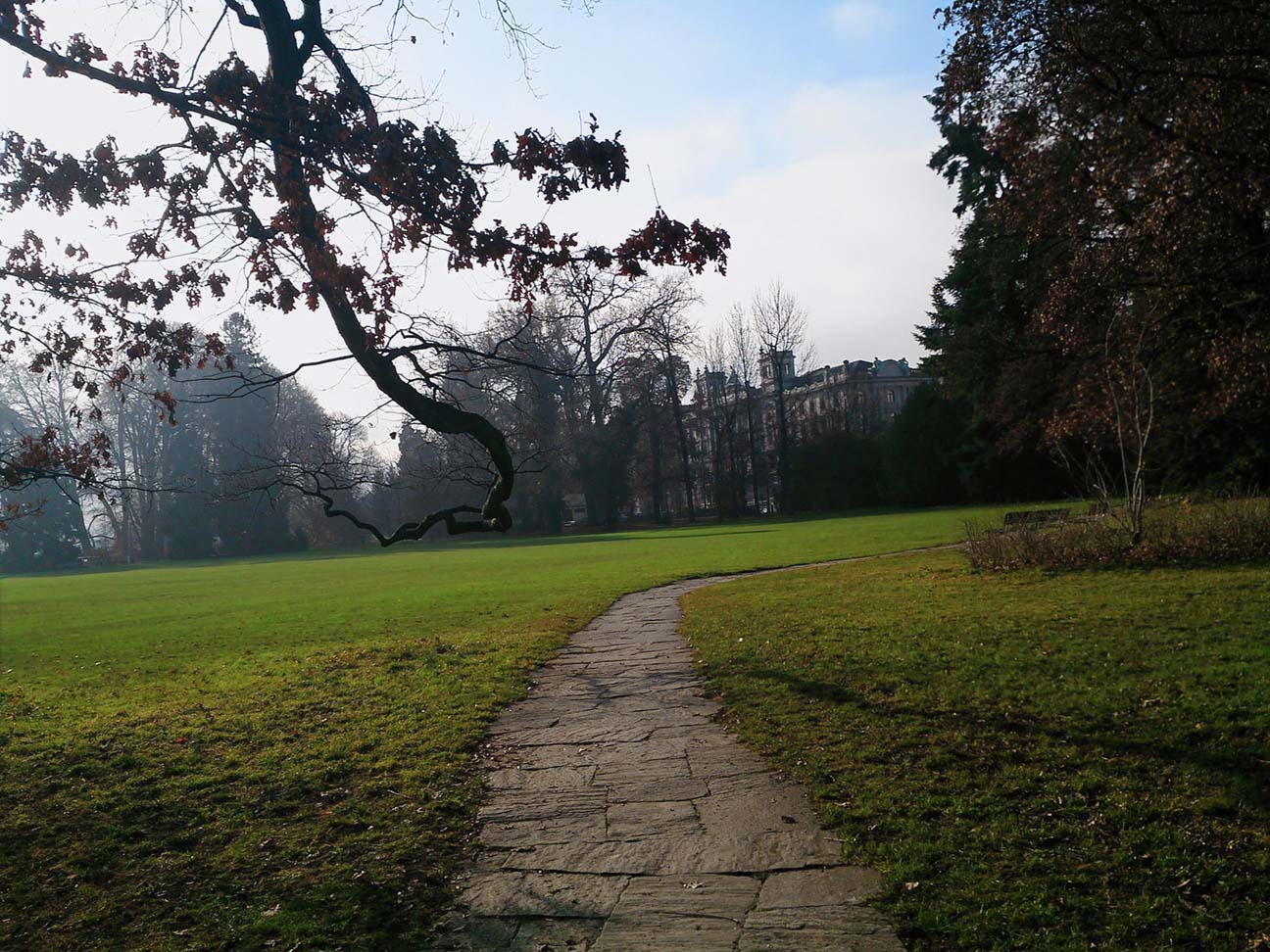}\\
   \includegraphics[width=0.36\linewidth]{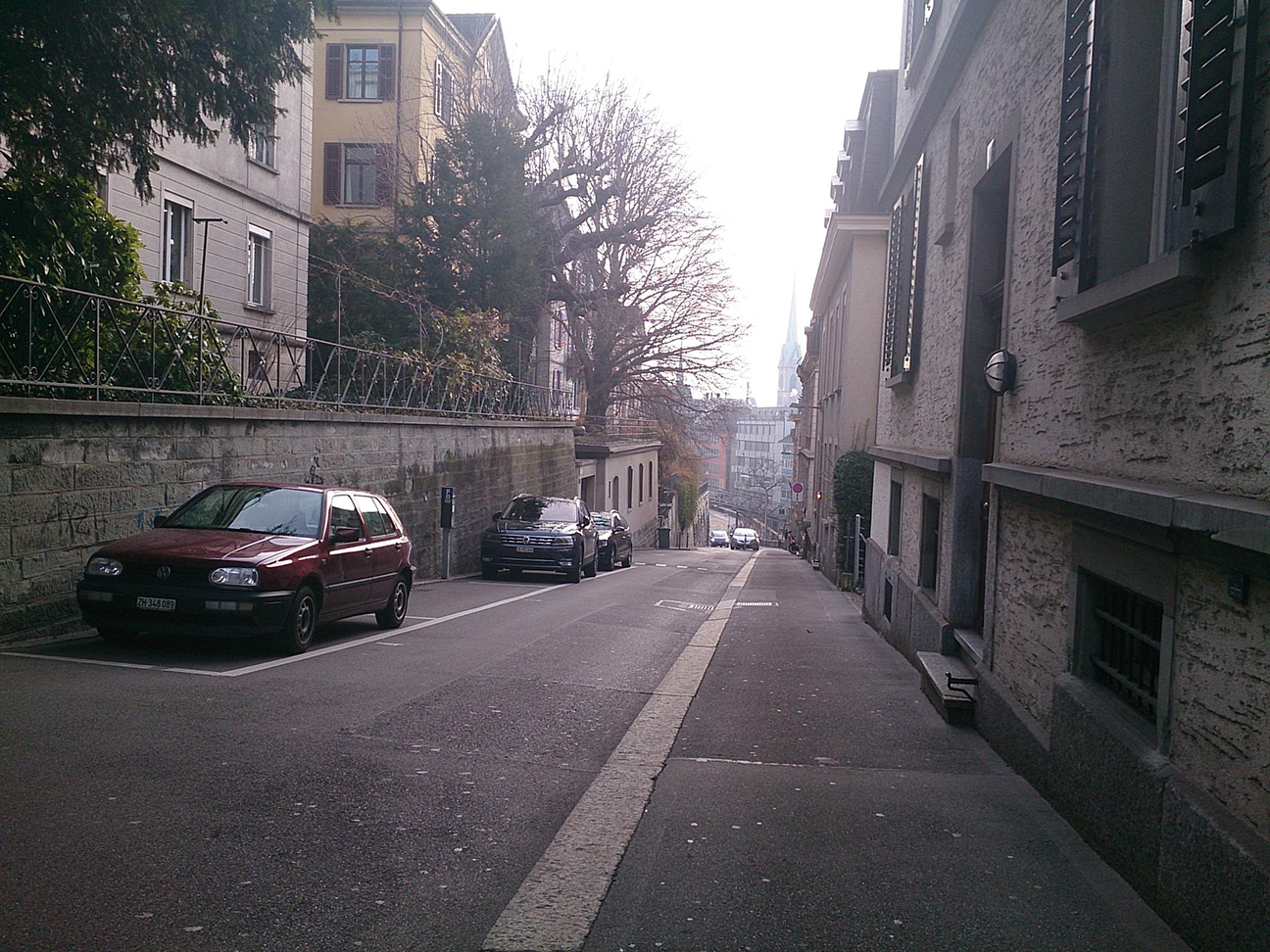}&
   \includegraphics[width=0.36\linewidth]{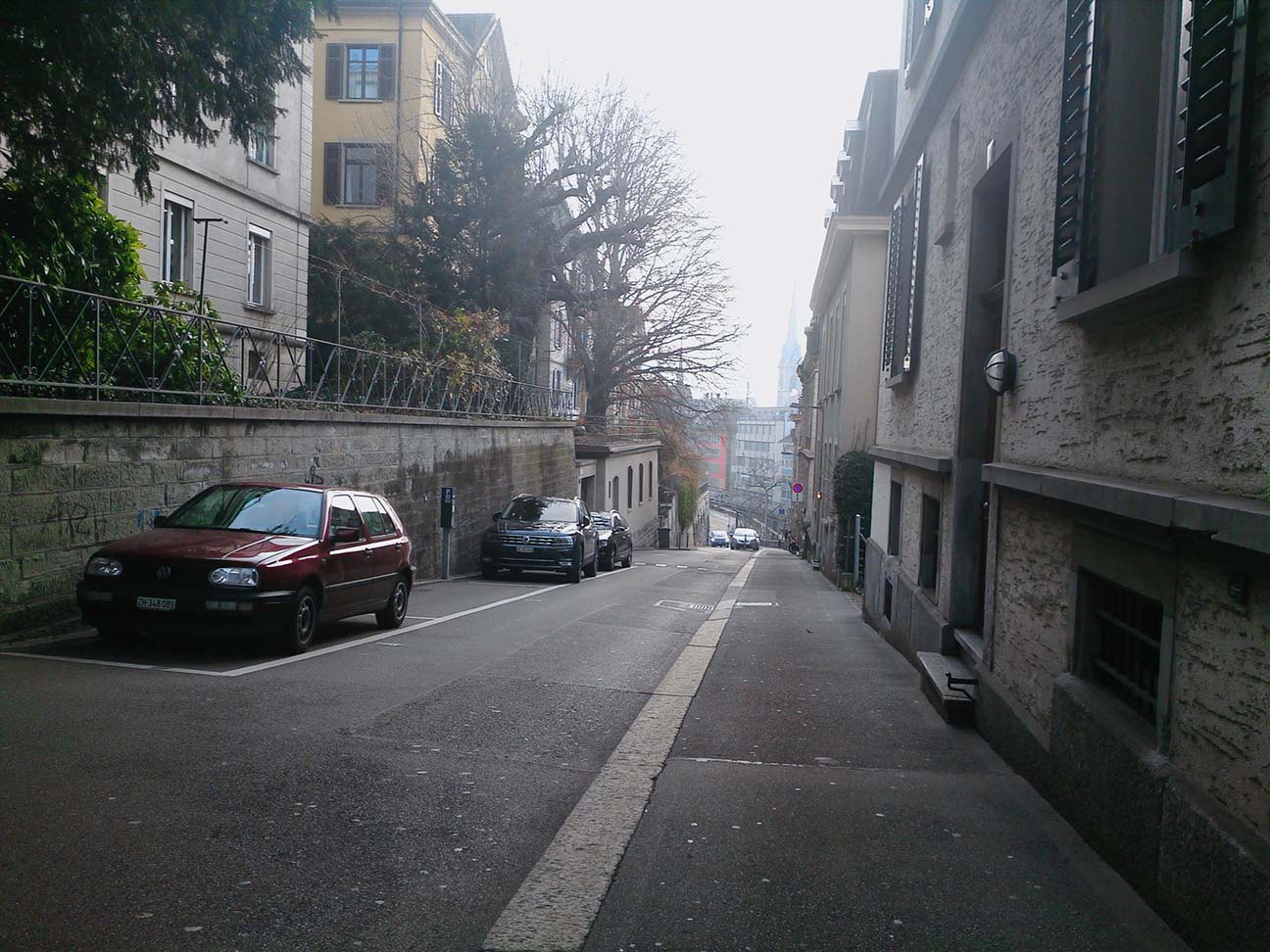}\\
\end{tabular}
\caption{Image results of our method for Sony DPED test images.}
\end{figure*}

\begin{figure*}[h]
\setlength{\tabcolsep}{1pt}
\centering
\begin{tabular}{cc}
Sony original & Enhanced with our method\\
   \includegraphics[width=0.39\linewidth]{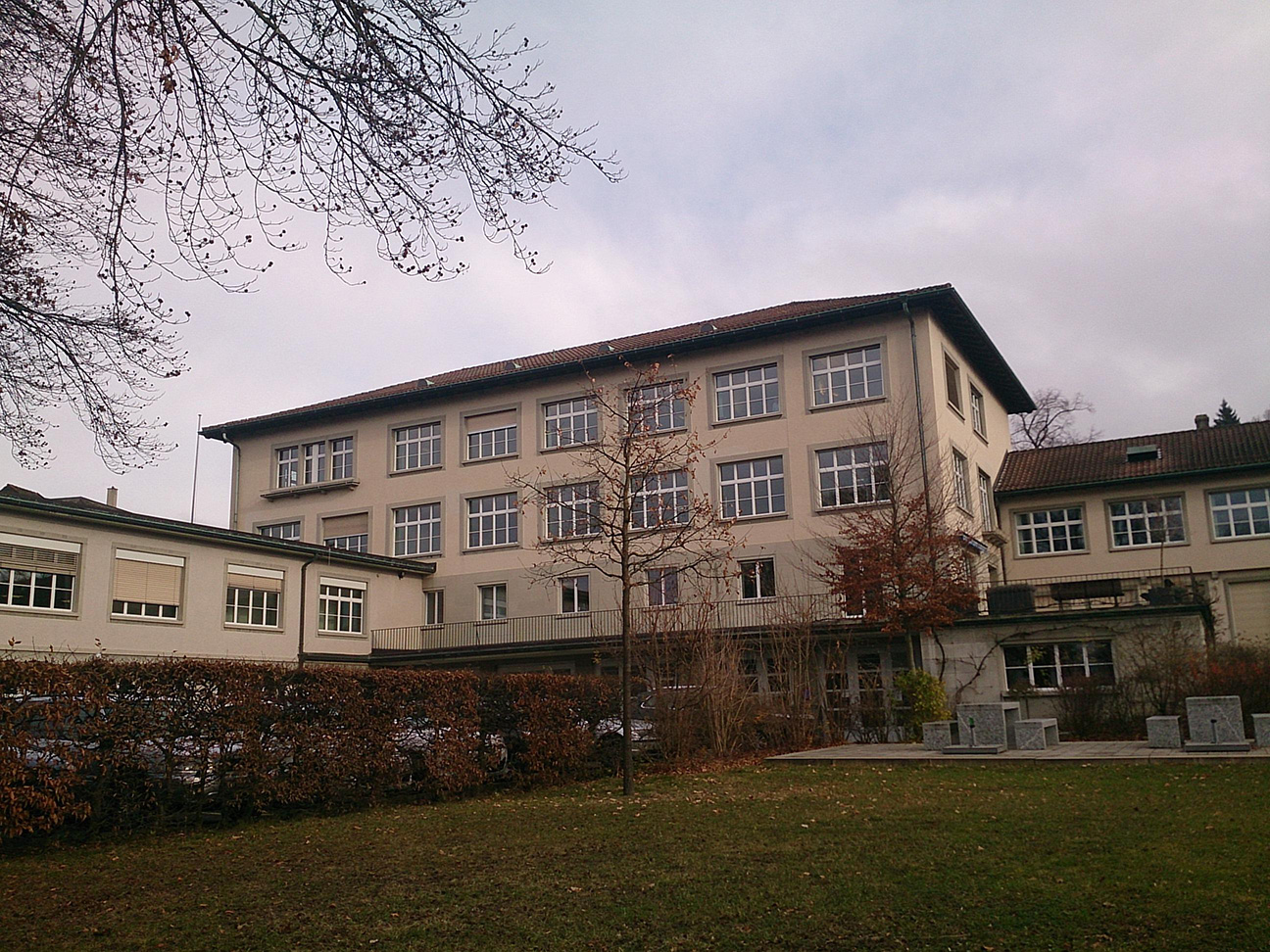}&
   \includegraphics[width=0.39\linewidth]{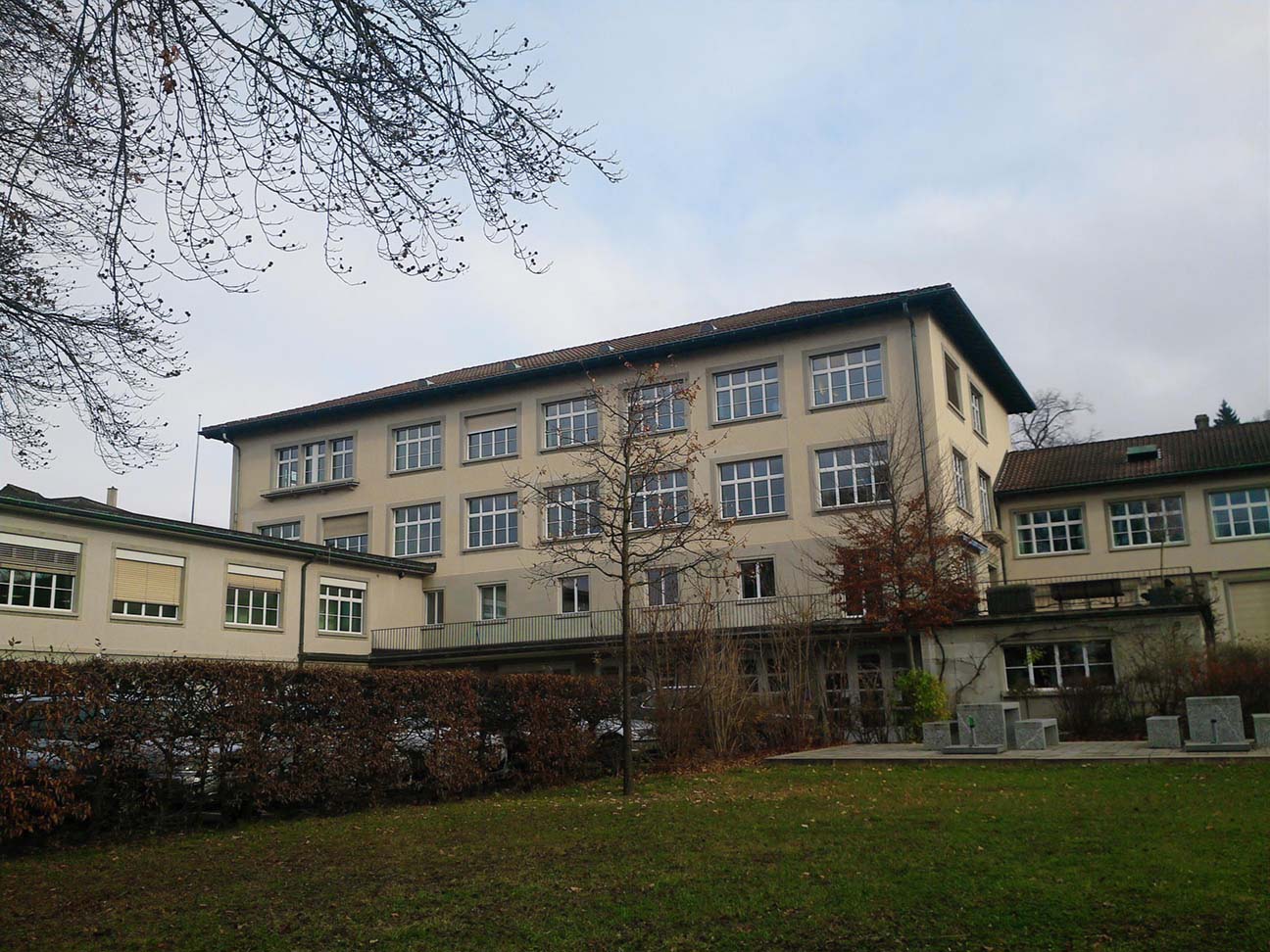}\\
   \includegraphics[width=0.39\linewidth]{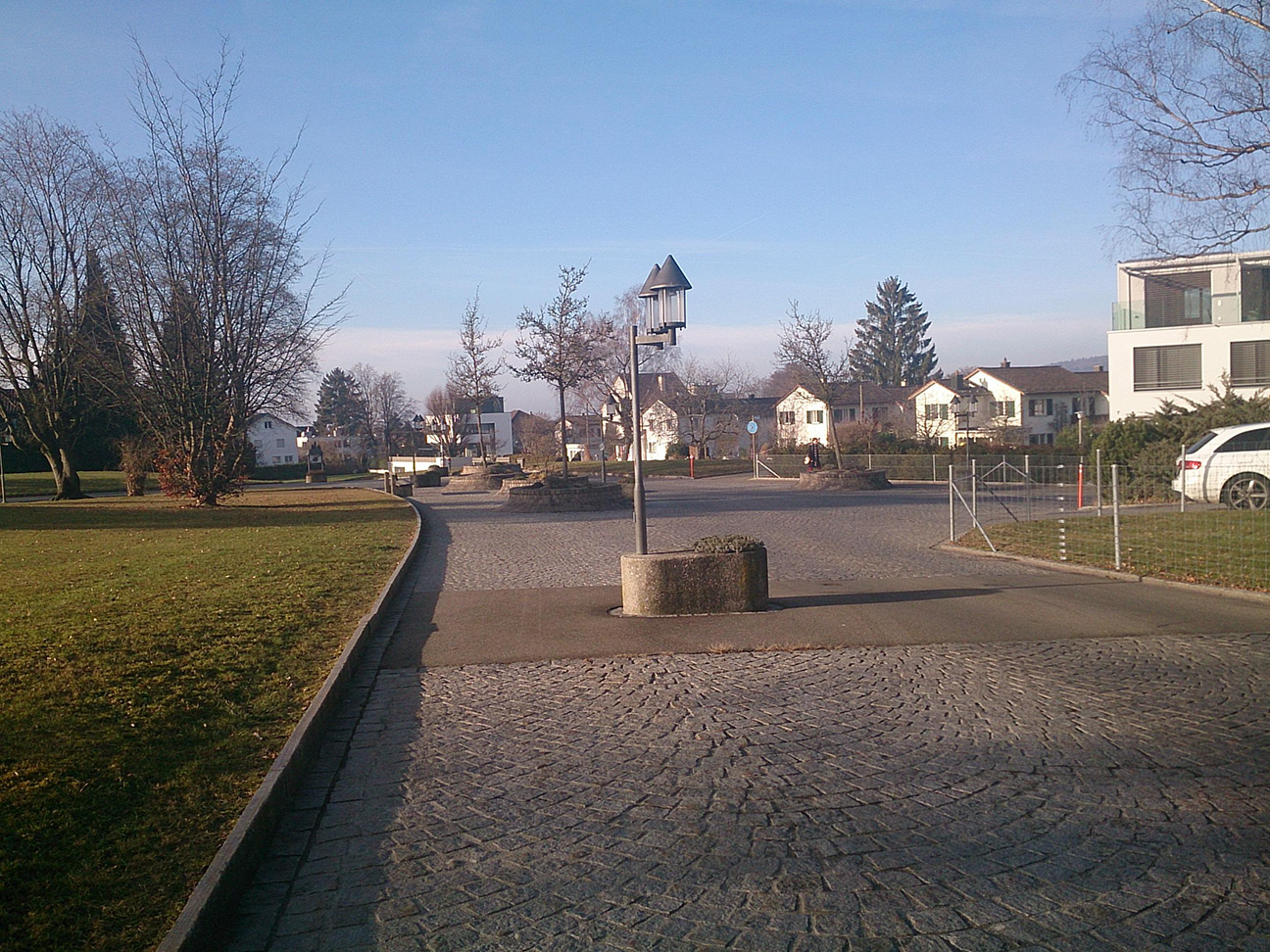}&
   \includegraphics[width=0.39\linewidth]{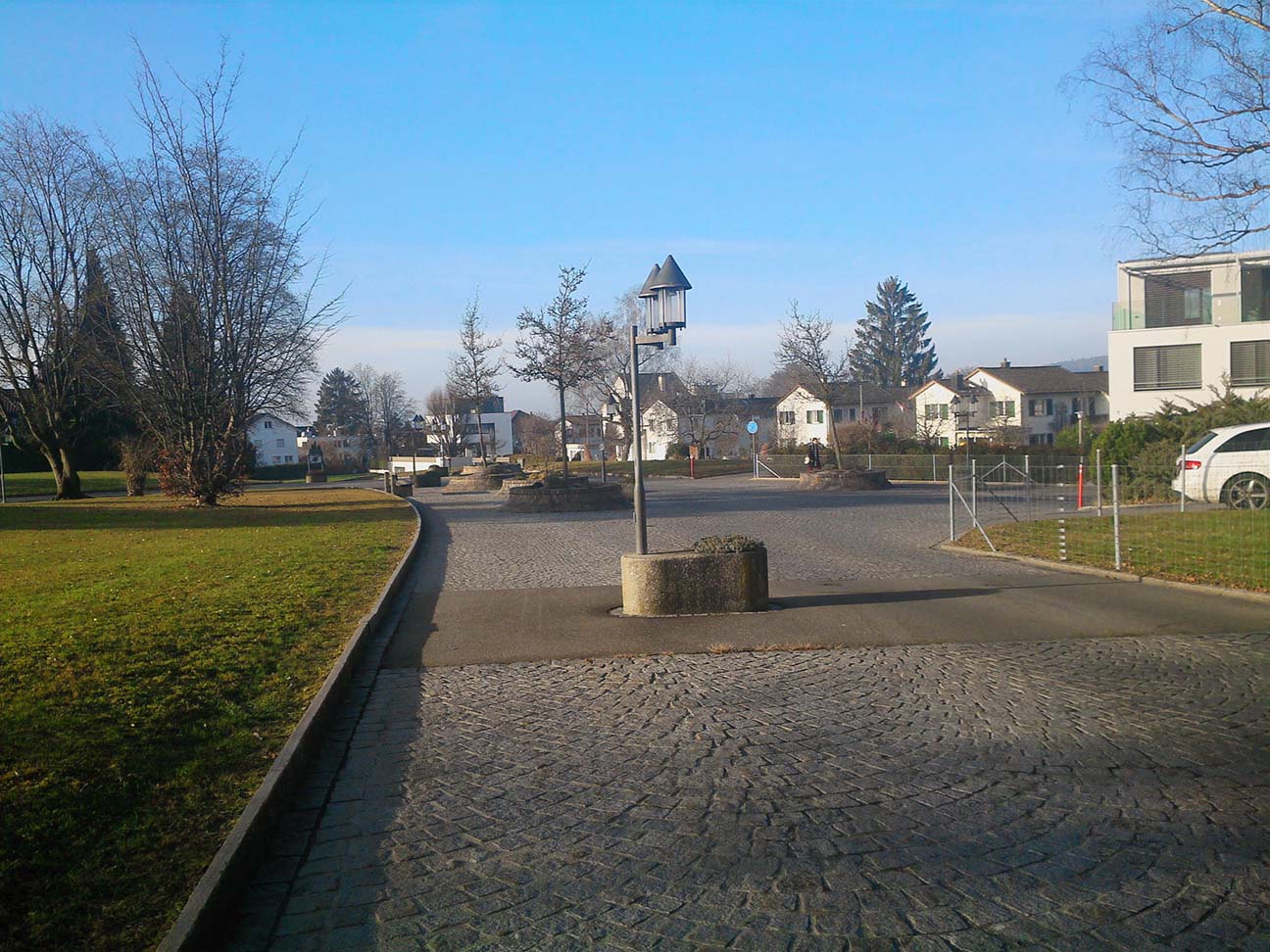}\\
   \includegraphics[width=0.39\linewidth]{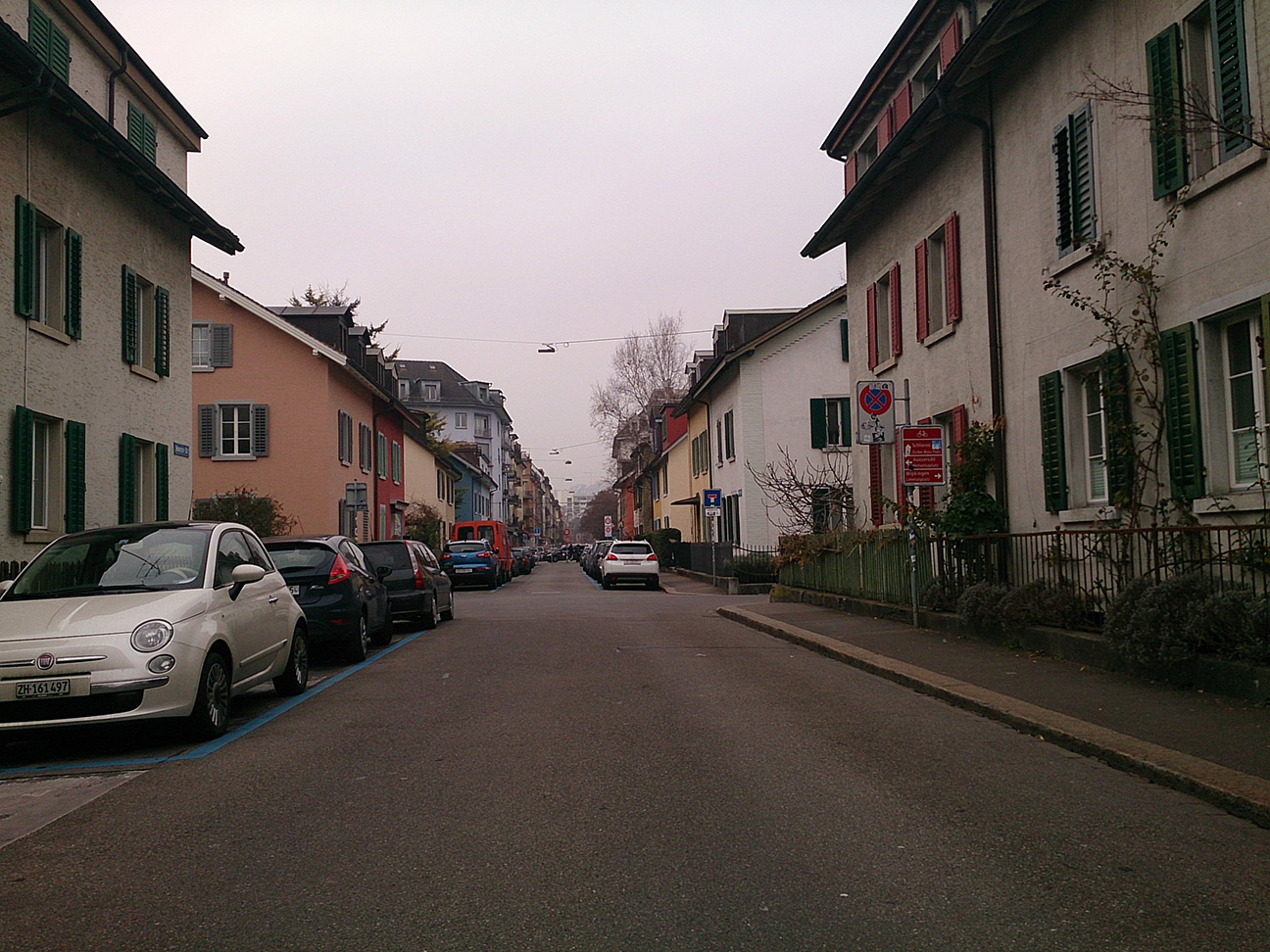}&
   \includegraphics[width=0.39\linewidth]{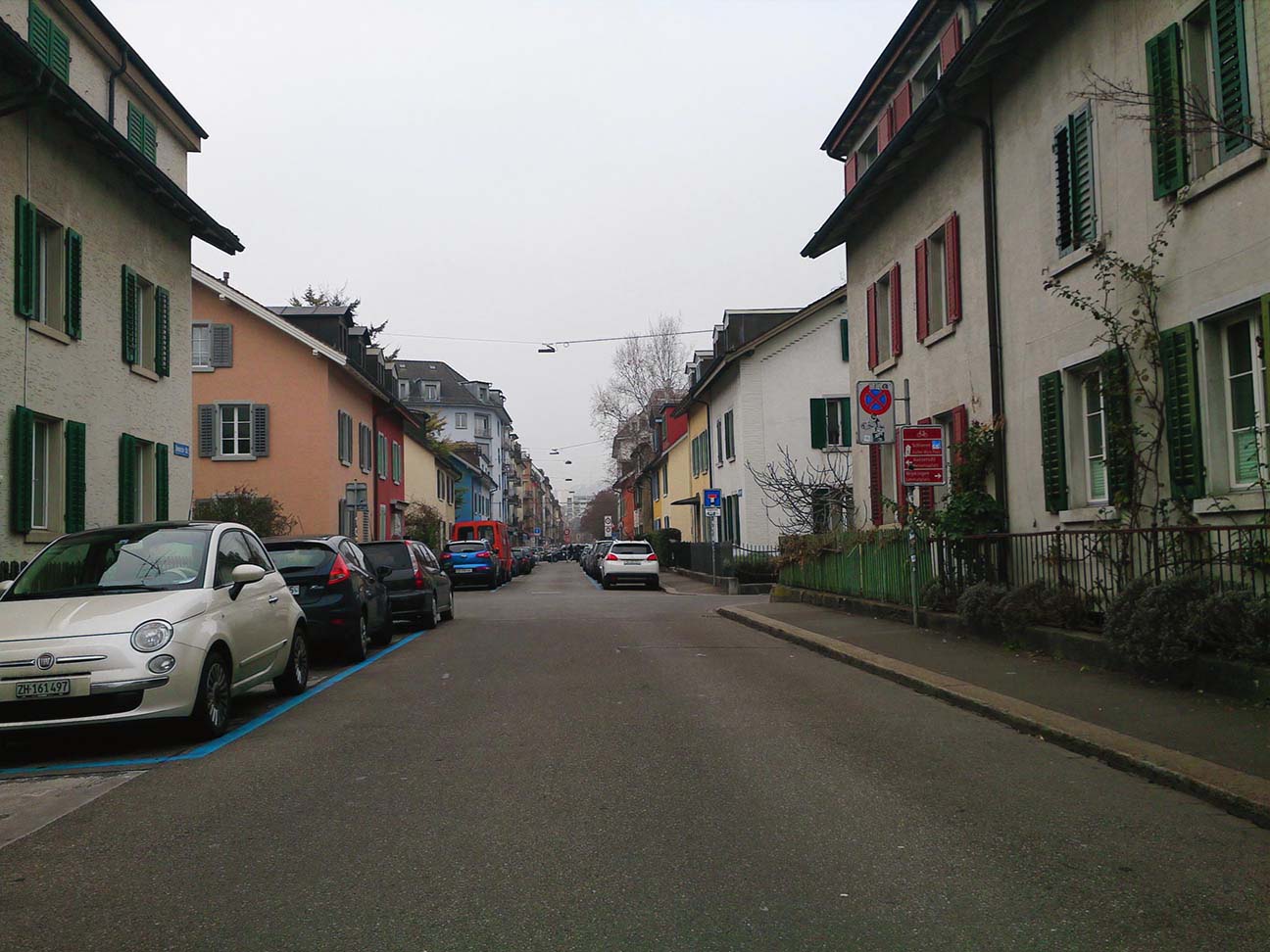}\\
   \includegraphics[width=0.39\linewidth]{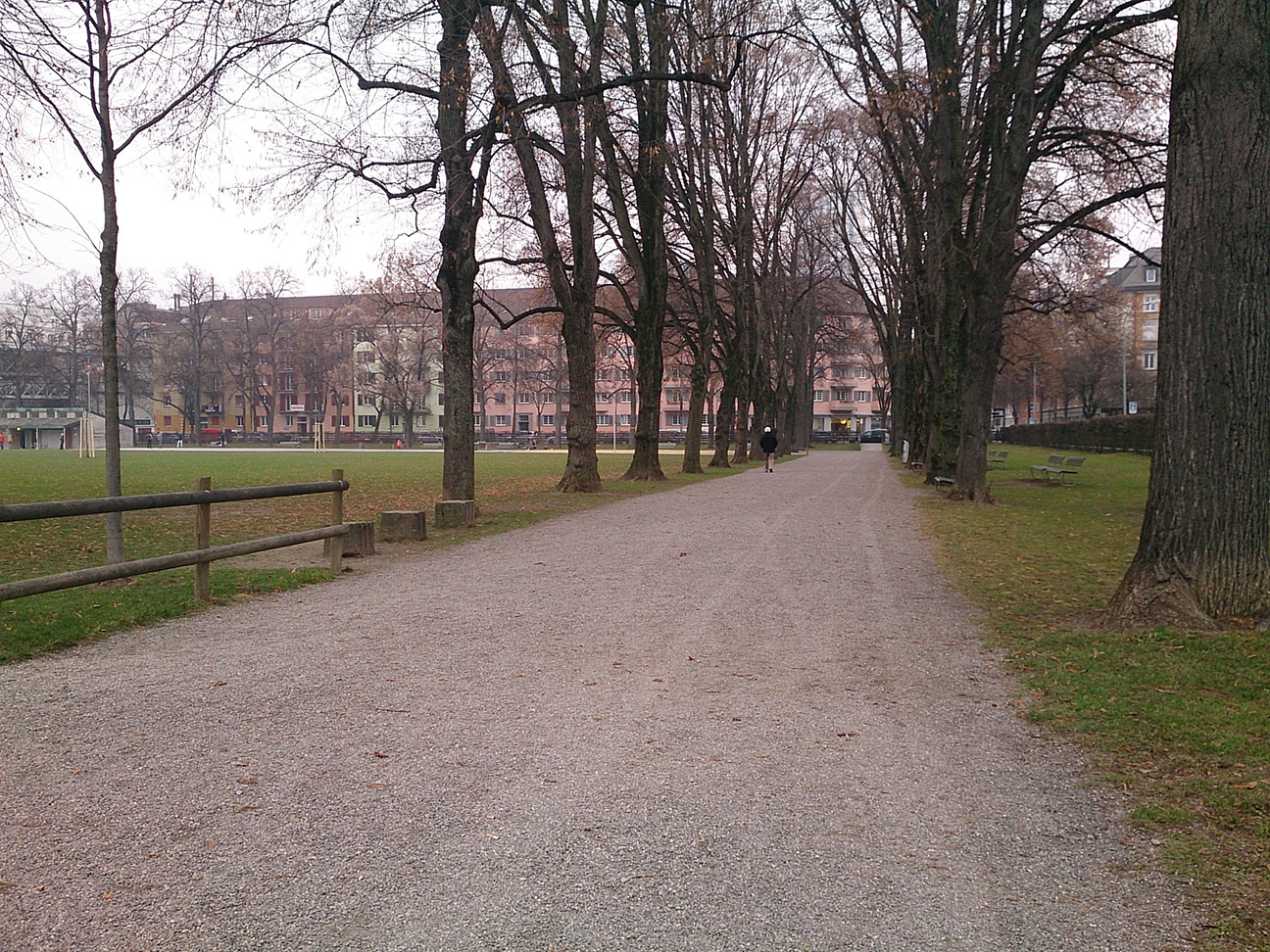}&
   \includegraphics[width=0.39\linewidth]{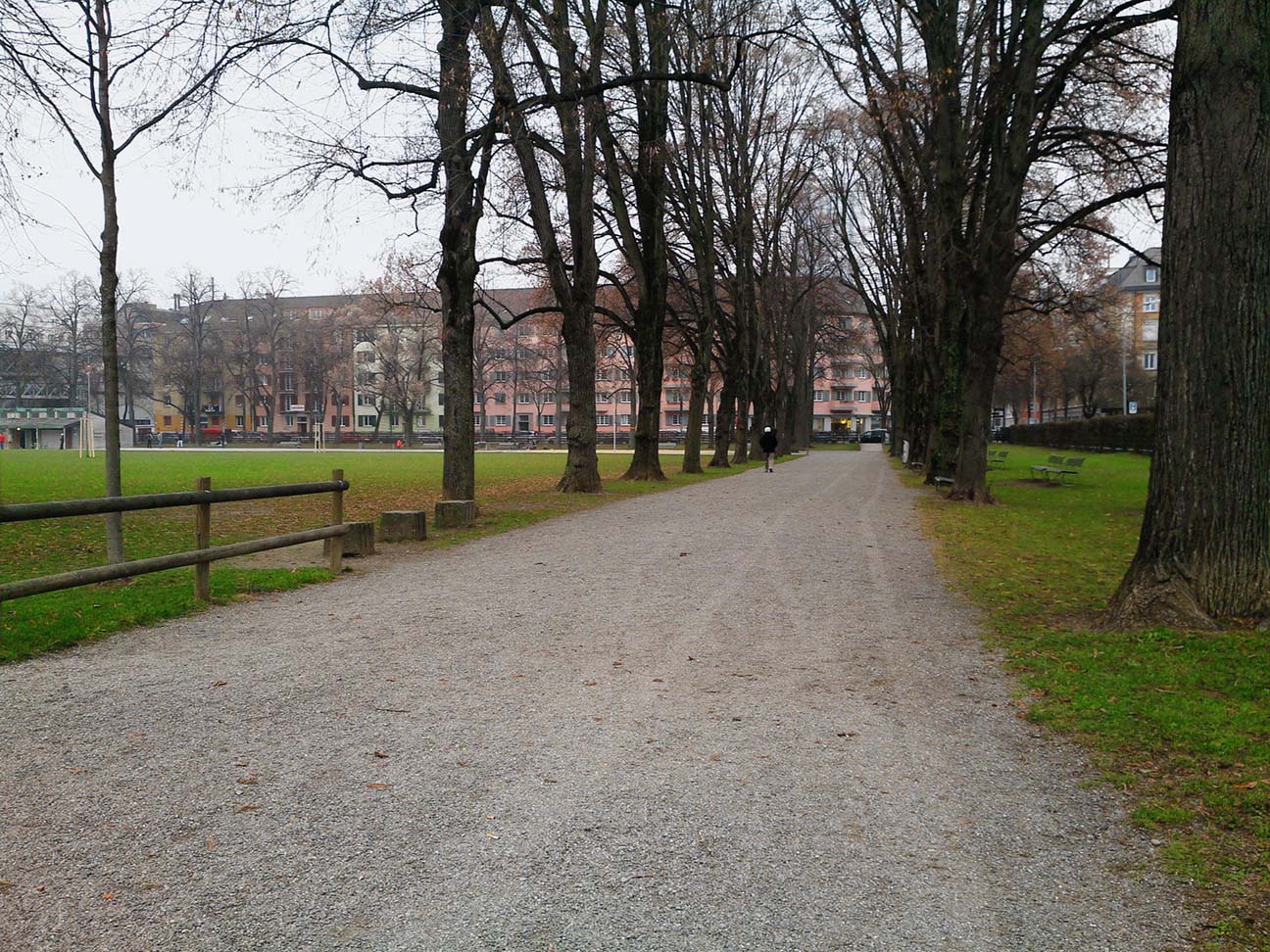}\\
\end{tabular}
\caption{Image results of our method for Sony DPED test images.}
\end{figure*}

\clearpage
\newpage


\justifying

\section{Appendix. Loss analysis}
\label{sec:loss_analysis}

\begin{figure*}[h!]
\centering
\setlength{\tabcolsep}{1pt}
\begin{tabular}{cccc}
\textbf{Color + Content + Texture} & Content + Texture & MSE + Texture & MSE\\
   \includegraphics[width=0.25\linewidth]{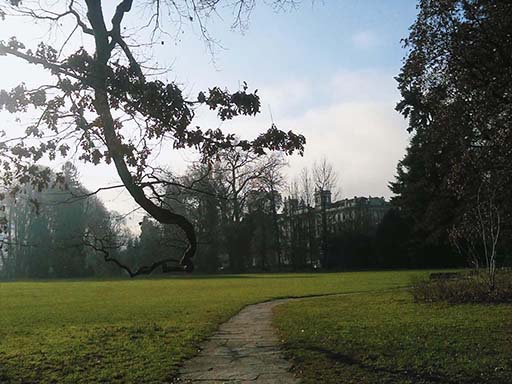}&
   \includegraphics[width=0.25\linewidth]{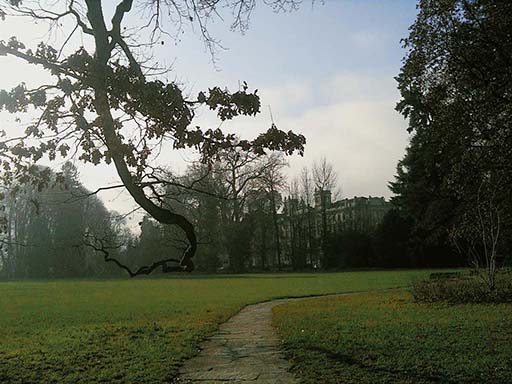}&
   \includegraphics[width=0.25\linewidth]{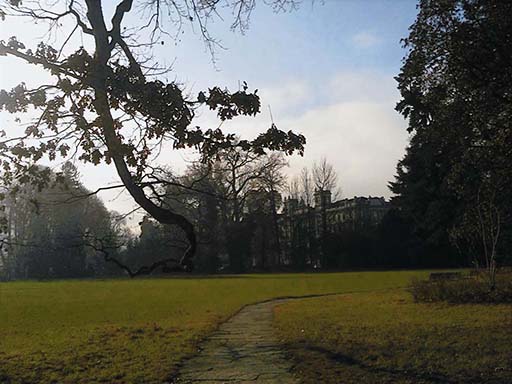}&
   \includegraphics[width=0.25\linewidth]{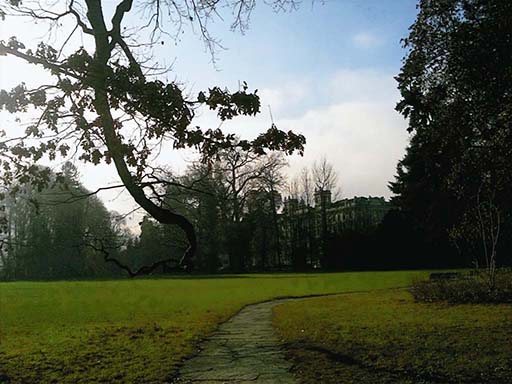}\\
   \includegraphics[width=0.25\linewidth]{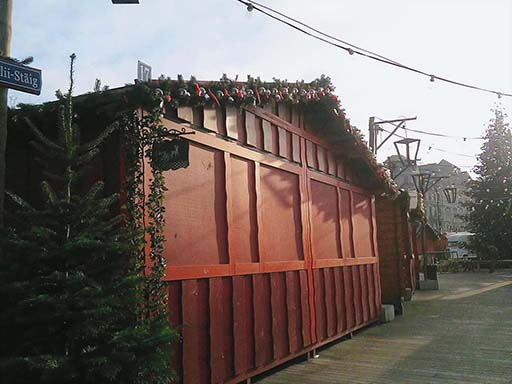}&
   \includegraphics[width=0.25\linewidth]{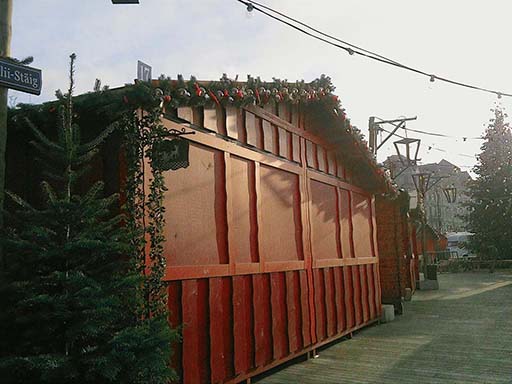}&
   \includegraphics[width=0.25\linewidth]{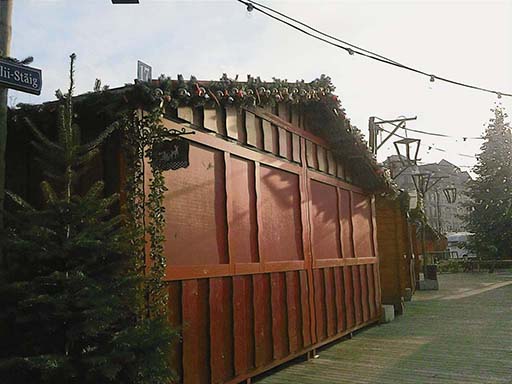}&
   \includegraphics[width=0.25\linewidth]{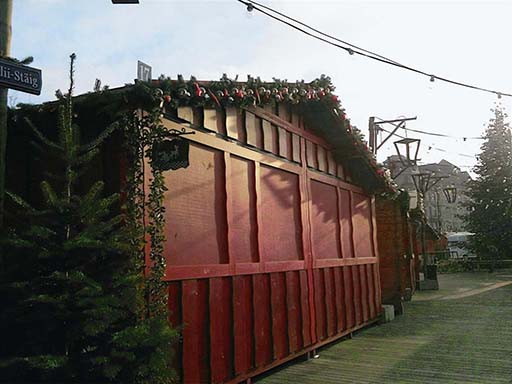}\\
   \includegraphics[width=0.25\linewidth]{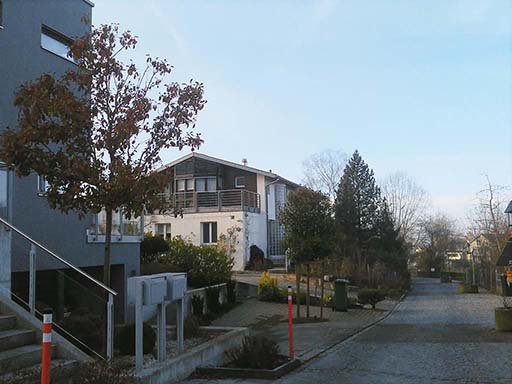}&
   \includegraphics[width=0.25\linewidth]{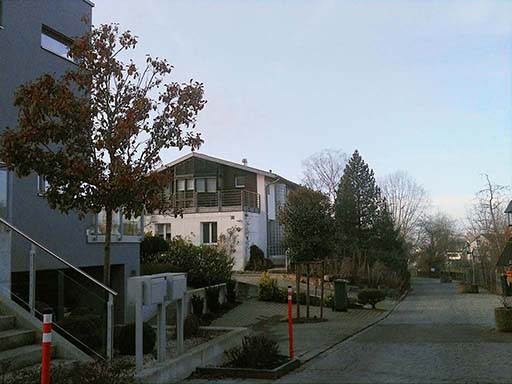}&
   \includegraphics[width=0.25\linewidth]{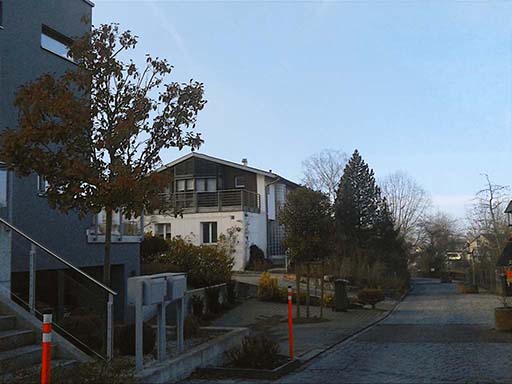}&
   \includegraphics[width=0.25\linewidth]{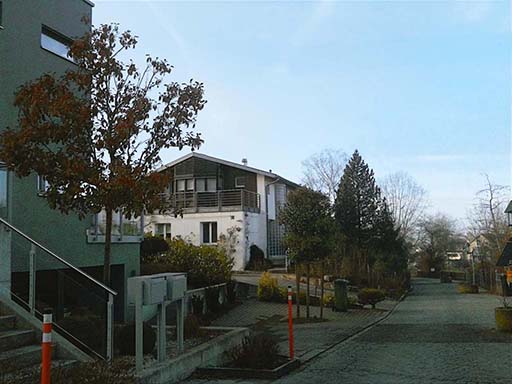}\\
\end{tabular}
\caption{Result images for iPhone camera for 4 different target loss functions.}
\label{fig:losses}
\end{figure*}

In this section, we study the contribution of different terms of the proposed perceptual loss function. For this purpose, we consider four different loss combinations: 1) the proposed one [color + content + texture], 2) [content + texture] loss, 3)~[MSE + texture] loss and 4) [MSE] loss. For each of these target loss combinations, a CNN was trained on the DPED dataset and validated on its test subset. The results of this experiment are provided in Table~\ref{tab:scores} and visual results are shown in Fig.~\ref{fig:losses}. As one can see, the adversarial network that stands behind the texture loss can cause significant color deviations, and the additional MSE term cannot effectively suppress them since it is not precise in this task (images are not perfectly aligned). Content loss shows better results in this case since it is less sensitive to image mismatches. Adding an extra color term further improves the resulting images, making the colors more saturated and closer to the target. Single MSE demonstrates high PSNR and SSIM values and natural color rendition while causing strong artifacts and slightly degrading image sharpness.
Overall our proposed [color + content + texture] loss leads to the best visual results while at the same time achieves top SSIM scores.

\begin{table*}[h!]
\caption{PSNR/SSIM scores for different loss functions.\label{tab:scores}}
\begin{center}
\begin{tabular}{|l|*{2}{c|}*{2}{c|}*{2}{c|}*{2}{c|}}
\hline
Phone &  \multicolumn{2}{|c|}{\textbf{Color + Content + Texture}} & \multicolumn{2}{|c|}{Content + Texture} & \multicolumn{2}{|c|}{MSE + Texture} &\multicolumn{2}{|c|}{MSE} \\
\hline
 & PSNR & SSIM & PSNR & SSIM & PSNR & SSIM & PSNR & SSIM \\
\hline
iPhone & 20.08 &\textbf{0.9201} & 19.05 & 0.9166 & 20.11 & 0.9125 & \textbf{20.56} & 0.9198 \\
BlackBerry & 20.07 & \textbf{0.9328} & 19.64 & 0.9312 & 20.13 & 0.9241 & \textbf{20.15} & 0.9292 \\
Sony & \textbf{21.81} & 0.9437 & 21.59 & 0.9426 & 21.72 & 0.9416 & 21.35 & \textbf{0.9453} \\
\hline
\end{tabular}
\end{center}
\end{table*}

\end{document}